\documentclass{article}
\usepackage[preprint]{neurips_2026}

\usepackage{placeins}
\usepackage[utf8]{inputenc} 
\usepackage[T1]{fontenc}    
\usepackage{hyperref}       
\usepackage{url}            
\usepackage{booktabs}       
\usepackage{amsfonts}       
\usepackage{nicefrac}       
\usepackage{microtype}      
\usepackage{xcolor}         
\usepackage{arydshln}
\usepackage{comment}
\usepackage{graphicx}
\usepackage{amsfonts}       
\usepackage{amsmath}
\usepackage{nicefrac}       
\usepackage{microtype}      
\usepackage{xcolor}         
\usepackage{comment}
\usepackage{graphicx}
\usepackage{subcaption}
\usepackage{tikz}
\usepackage{pgfplots}
\usepackage{pgfplotstable}
\usepgfplotslibrary{groupplots}
\pgfplotsset{compat=1.18}
\usepackage{threeparttable}  
\usepackage{pgfplots}
\pgfplotsset{compat=1.18}
\usetikzlibrary{plotmarks,calc}
\definecolor{cNTP}{HTML}{1F77B4}
\definecolor{cMAE}{HTML}{E8A317}
\definecolor{cDiffusion}{HTML}{2CA02C}
\definecolor{cDINO}{HTML}{D670B5}
\definecolor{cJEPA}{HTML}{56B4E9}
\definecolor{cLeJEPA}{HTML}{E8601C}
\definecolor{cSoftCLT}{HTML}{6B5B4B}
\definecolor{cMonash}{HTML}{222222}
\definecolor{cSynS}{HTML}{1F77B4}
\definecolor{cSynL}{HTML}{E8601C}
\definecolor{cHyb}{HTML}{1B9E77}
\definecolor{hmneg}{HTML}{B5532C}
\definecolor{hmpos}{HTML}{1F6E9C}
\definecolor{hmzero}{HTML}{F7F4EF}
\definecolor{rkdark}{HTML}{1F6E9C}  
\definecolor{rklight}{HTML}{EAF1FA}
\definecolor{figgray}{HTML}{555555}
\newcommand{\figfont}{\sffamily\fontsize{6.5}{7.5}\selectfont}
\newcommand{\figfontS}{\sffamily\fontsize{5.5}{6.5}\selectfont}
\newcommand{\figfontXS}{\sffamily\fontsize{4.7}{5.2}\selectfont}
\pgfplotsset{
  mNTP/.style      ={only marks, mark=*,         mark size=1.5pt, draw=cNTP,       fill=cNTP},
  mMAE/.style      ={only marks, mark=square*,   mark size=1.4pt, draw=cMAE,       fill=cMAE},
  mDiffusion/.style={only marks, mark=triangle*, mark size=1.8pt, draw=cDiffusion, fill=cDiffusion},
  mDINO/.style     ={only marks, mark=diamond*,  mark size=1.9pt, draw=cDINO,      fill=cDINO},
  mJEPA/.style     ={only marks, mark=triangle*, mark size=1.8pt, draw=cJEPA,      fill=cJEPA,
                     mark options={rotate=180}},
  mLeJEPA/.style   ={only marks, mark=+,         mark size=2.0pt, draw=cLeJEPA,    line width=.8pt},
  mSoftCLT/.style  ={only marks, mark=asterisk,  mark size=1.8pt, draw=cSoftCLT,   line width=.6pt},
  mMonash/.style   ={only marks, mark=*,         mark size=1.4pt, draw=cMonash,    fill=cMonash},
  mSynS/.style     ={only marks, mark=square*,   mark size=1.3pt, draw=cSynS,      fill=cSynS},
  mSynL/.style     ={only marks, mark=triangle*, mark size=1.7pt, draw=cSynL,      fill=cSynL},
  mHyb/.style      ={only marks, mark=diamond*,  mark size=1.8pt, draw=cHyb,       fill=cHyb},
  sslaxis/.style={
    tick label style={font=\figfontS}, label style={font=\figfont},
    title style={font=\figfont\bfseries, at={(0,1)}, anchor=south west, yshift=-1pt},
    tick align=outside, tick pos=left, major tick length=2pt,
    axis line style={figgray, line width=.4pt}, tick style={figgray, line width=.4pt},
    xlabel shift=-3pt, ylabel shift=-4pt, clip mode=individual},
}
\newcommand{\figkey}[2]{\tikz[baseline=-.6ex]{\path[#1] plot coordinates {(0,0)};}\,{\figfont #2}}
\newcommand{\objlegend}{%
  \figkey{color=cNTP, mark=*, mark size=1.5pt}{NTP}\quad
  \figkey{color=cMAE, mark=square*, mark size=1.4pt}{MAE}\quad
  \figkey{color=cDiffusion, mark=triangle*, mark size=1.8pt}{Diffusion}\quad
  \figkey{color=cDINO, mark=diamond*, mark size=1.9pt}{DINO}\quad
  \figkey{color=cJEPA, mark=triangle*, mark size=1.8pt, mark options={rotate=180}}{JEPA}\quad
  \figkey{color=cLeJEPA, mark=+, mark size=2pt, line width=.8pt}{Le-JEPA}\quad
  \figkey{color=cSoftCLT, mark=asterisk, mark size=1.8pt, line width=.6pt}{SoftCLT}}
\newcommand{\srclegend}{%
  \figkey{color=cMonash, mark=*, mark size=1.4pt}{Monash (680K)}\quad
  \figkey{color=cSynS, mark=square*, mark size=1.3pt}{Synthetic (680K)}\quad
  \figkey{color=cSynL, mark=triangle*, mark size=1.7pt}{Synthetic (2.5M)}\quad
  \figkey{color=cHyb, mark=diamond*, mark size=1.8pt}{Hybrid (2.5M)}}
\newlength{\hmw}
\newlength{\hmh}
\def\hmclip{300}                           
\def\hmmax{100}                            
\colorlet{hmtext}{black}                   
\def\hmstrip-#1\relax{#1}
\newcommand{\hmcell}[3]{%
  \pgfmathsetmacro\hmv{#3}%
  \pgfmathsetmacro\hmp{\hmmax*min(1,abs(\hmv)/\hmclip)}%
  \pgfmathtruncatemacro\hmsgn{(\hmv>0)-(\hmv<0)}%
  \edef\hmtxt{#3}%
  \ifnum\hmsgn<0 \def\hmfill{hmneg!\hmp!hmzero}\else\def\hmfill{hmpos!\hmp!hmzero}\fi
  \path[fill=\hmfill, draw=white, line width=.5pt]
    ({(#1)*\hmw},{-(#2)*\hmh}) rectangle ++(\hmw,-\hmh)
    node[pos=.5, inner sep=0pt, font=\figfontXS\bfseries, text=hmtext]{%
      \ifnum\hmsgn<0 --\expandafter\hmstrip\hmtxt\relax\else\hmtxt\fi};}
\newcommand{\hmrow}[3]{\foreach \hms [count=\hmc from 0] in {#3} {\hmcell{\hmc+#2}{#1}{\hms}}}
\newlength{\rkw}
\colorlet{rktext}{black}                   
\newcommand{\rkcell}[3]{%
  \pgfmathsetmacro\rkp{100*(8-#3)/7}%
  \path[fill=rkdark!\rkp!rklight, draw=white, line width=.5pt]
    ({(#1)*\rkw},{-(#2)*\rkw}) rectangle ++(\rkw,-\rkw)
    node[pos=.5, inner sep=0pt, font=\figfontS, text=rktext]{\ifnum#3=1 \bfseries\fi #3};}
\newcommand{\rkrow}[3]{\foreach \rks [count=\rkc from 0] in {#3}
  {\pgfmathsetmacro\rkx{\rkc+#2+0.35*floor(\rkc/3)}\rkcell{\rkx}{#1}{\rks}}}

\usepackage{enumitem}
\usepackage{multirow}

\definecolor{oiBlue}{RGB}{0,114,178}
\definecolor{oiOrange}{RGB}{230,159,0}
\definecolor{oiGreen}{RGB}{0,158,115}
\definecolor{oiRed}{RGB}{213,94,0}
\definecolor{oiPurple}{RGB}{204,121,167}
\definecolor{oiBrown}{RGB}{150,90,50}

\title{When Does Self-Supervised Learning Transfer to Time-Series Tasks?}

\author{
Noam Major \quad
Kathy Razmadze \quad
Yoli Shavit \\[0.5em]
Bar-Ilan University, Ramat-Gan, Israel
}

\begin{document}

\maketitle

\begin{abstract}
Self-supervised learning (SSL) assumes that solving pretext tasks on unlabeled data yields representations that transfer effectively across downstream applications via linear probing or fine-tuning. While this paradigm has driven major progress in vision and language, its benefits for time series remain under-investigated and often confounded by inconsistent experimental controls. To address this gap, we benchmark seven representative methods from five key SSL paradigms across anomaly detection, classification, and forecasting under parameter- and data-matched budgets. We find that transfer efficacy depends heavily on the downstream task. SSL yields substantial gains in anomaly detection and provides effective initializations for classification under fine-tuning, but offers limited to no advantage over non-pre-trained controls in forecasting. Furthermore, linear probing does not reliably predict fine-tuning performance, synthetic pre-training is often competitive with real-world corpora, and scaling encoder depth degrades forecasting accuracy. We synthesize these empirical results into practical evaluation and development guidelines for time-series SSL. 
\end{abstract}

\section{Introduction}\label{sec:intro}
The success of self-supervised learning (SSL) in natural language processing \citep{devlin2019bert} and computer vision \citep{dino,jepa} has motivated its widespread adoption for time-series analysis \citep{nie2022time, timedart, timesfm, ansari2024chronos}. SSL relies on the foundational premise that models can learn underlying structures and semantic relationships from unlabeled data that transfer effectively to downstream tasks. While this premise is well established in language and vision, its validity in time series remains less clear due to substantial cross-domain distribution shifts and diverse temporal dynamics.
 
To date, empirical evidence supporting time-series SSL remains fragmented. Much of the recent progress in time-series foundation models has been demonstrated in forecasting and, in particular, under zero-shot evaluation~\citep{timesfm,ansari2024chronos}. 
Beyond this narrow task scope, newly proposed time-series SSL methods often introduce specialized backbones, making it difficult to attribute observed performance gains to the SSL objective itself. Existing surveys on time-series SSL provide informative high-level taxonomies of contrastive, generative, and reconstruction methods~\citep{ssl_ts_survey_tpami, universal_ts_survey, contrastive_vs_Generative_ts}, but generally do not provide controlled, cross-task experiments that decouple the contribution of the pre-training objective from confounding variations in model architecture, training data, and evaluation protocols. Furthermore, latent-space alignment frameworks, such as invariance-based methods (e.g., DINO~\citep{dino}) and Joint-Embedding Predictive Architectures (e.g., JEPA~\citep{lecun2022path, jepa}, LeJEPA~\citep{balestriero2511lejepa}), remain largely underexplored in the time-series literature. Finally, while benchmarking libraries such as TSLib~\citep{tslib} provide valuable standardization of downstream evaluation, they do not natively support controlled pre-training workflows.

To address this gap, we empirically investigate the question: \textit{When Does Self-Supervised Learning Transfer to Time-Series Tasks?}
Specifically, we formulate three guiding research questions:
\begin{itemize}
\item RQ1: When does SSL pre-training improve performance in time-series forecasting, classification, and anomaly detection over random frozen features (linear probing) or random initialization (fine-tuning)?
\item RQ2: Do certain SSL methods consistently outperform others across different downstream tasks?
\item RQ3: How do pre-training data source (real vs. synthetic), data scale, and backbone depth affect representation transfer?
\end{itemize}

To answer these questions, we benchmark seven representative methods across five key SSL paradigms: reconstruction, generative modeling, contrastive learning, invariance-based distillation, and joint-embedding predictive architectures, alongside non-pre-trained controls. We evaluate models pretrained on the real-world Monash repository~\citep{monash}, synthetic datasets~\citep{timepfn}, and a mixture of the two, as well as in-domain pre-trained controls. To isolate the effect of each SSL method, we keep the backbone architecture (a standard Transformer encoder), data splits, training budgets, and evaluation protocols strictly identical across comparisons.

Our empirical results show that transfer efficacy depends on the downstream task and evaluation protocol. SSL yields substantial gains in anomaly detection and provides effective initializations for classification under fine-tuning, but offers limited to no advantage over non-pre-trained baselines in forecasting. Furthermore, linear probing does not reliably predict fine-tuning outcomes. For example, when pre-trained on Monash, LeJEPA achieves the highest mean accuracy on five out of nine classification datasets under linear probing, but on only two after fine-tuning. Downstream rankings in anomaly detection also vary substantially across evaluation metrics. In data and compute scaling, scale-matched synthetic pre-training is often competitive with real-world corpora but increasing data volume through synthetic generation does not improve performance. Finally, scaling Transformer depth consistently degrades forecasting accuracy. 

We distill our empirical findings into practical guidelines for time-series SSL, offering recommendations for selecting pretext objectives, establishing rigorous baseline controls, and standardizing downstream evaluation protocols.
\section{Preliminaries}
\label{sec:setup}
\subsection{Self-Supervised Learning Paradigms}
\label{subsec:ssl_paradigms}

Self-supervised learning (SSL) aims to learn generalizable representations from unlabeled data by optimizing proxy objectives designed around intrinsic structural priors~\citep{chen2020simple}. In this work, we systematically evaluate five core SSL paradigms: reconstruction, generative modeling, invariance-based distillation, contrastive learning, and joint-embedding predictive architectures (JEPA),  selecting representative methods for each (Table~\ref{tab:objectives_inputs}). 
Following \citet{nie2022time}, we adopt a channel-independent design in which each channel (variate) of a multivariate series $\mathbf{x} \in \mathbb{R}^{T \times C}$ is separately segmented into $L$ non-overlapping patches $\mathbf{x}_i \in \mathbb{R}^P$ ($T = L \cdot P$). A shared Transformer encoder $f_\theta$ maps the patched series to latent representations $\mathbf{z} = f_\theta(\mathbf{x}) \in \mathbb{R}^{L \times D}$, which are processed by method-specific heads and serve as the backbone features for downstream adaptation. Full architectural specifications and hyperparameters appear in Appendix~\ref{sec:appendix_objectives} and \ref{subsec:appendix_implement}.

\paragraph{Reconstruction-Based SSL.}
Reconstruction methods learn contextual representations by recovering masked portions of the input signal. We evaluate Masked Auto-Encoding (\textbf{MAE}), popularized for time series by PatchTST~\citep{nie2022time}. Given a randomly sampled subset of masked patch indices $\mathcal{M} \subset \{1, \dots, L\}$, the masked patches are set to zero and the resulting sequence $\tilde{\mathbf{x}}$ is encoded and passed to a lightweight decoder $\phi$ to reconstruct the masked raw signal by minimizing the Mean Squared Error (MSE):
\begin{equation}
\label{eq:MAE}
\frac{1}{|\mathcal{M}|} \sum_{i \in \mathcal{M}} \|\mathbf{x}_i - \hat{\mathbf{x}}_i\|_2^2,
\end{equation}
where $\hat{\mathbf{x}}_i = [\phi(f_\theta(\tilde{\mathbf{x}}))]_i$.

\paragraph{Generative SSL.}
Generative methods learn the data distribution by conditioning on an observed context to estimate unobserved signal values. Within this paradigm, we evaluate two representative formulations: \textbf{Next-Token (Patch) Prediction (NTP)} and \textbf{Diffusion}. NTP adopts an autoregressive framework common in time-series foundation models such as TimesFM~\citep{timesfm}. Given $L_c$ causal context patches, a shallow residual MLP head $\psi_h$ predicts a continuous future horizon of $h$ patches across $N$ sliding windows by minimizing the prediction error:
\begin{equation}
\label{eq:NTP}
\frac{1}{N} \sum_{j=1}^{N} \|\psi_h(f_\theta(\mathbf{x}_{1:L_c}^{(j)})) - \mathbf{x}_{L_c+1:L_c+h}^{(j)}\|_2^2.
\end{equation}
The Diffusion representative, adapted from TimeDART~\citep{timedart}, learns to reverse a forward Gaussian corruption process. At timestep $t_i$ with noise schedule $\bar{\alpha}_{t_i}$ and noise vector ${\epsilon}_i \sim \mathcal{N}(\mathbf{0}, \mathbf{I})$, a corrupted patch $\widetilde{\mathbf{x}}_i = \sqrt{\bar{\alpha}_{t_i}}\mathbf{x}_i + \sqrt{1-\bar{\alpha}_{t_i}}\boldsymbol{\epsilon}_i$ is embedded as $\hat{\mathbf{z}}_i^{in}$, and a denoising decoder $\phi$ reconstructs clean patches conditioned on clean preceding context by optimizing:
\begin{equation}
\label{eq:Diff}
\sum_{n=0}^{N-1} \mathbb{E}_{\boldsymbol{\epsilon}, q(\mathbf{x}_n)} \big[ \|\mathbf{x}_{n+1} - \phi(\hat{\mathbf{z}}_{n+1}^{in},\, f_\theta(\mathbf{z}_{1:n}^{in}))\|_2^2 \big].
\end{equation}

\paragraph{Invariance-Based SSL.}
Rather than predicting in the raw signal space, Invariance-based methods encourage representations of different views of the same sequence to map to consistent representations in the latent space. We evaluate \textbf{DINO}~\citep{dino}, which trains a student network to match centered and temperature-sharpened categorical distributions of an EMA teacher across view pairs $\mathcal{P}$ via cross-entropy:
\begin{equation}
\label{eq:dino}
-\frac{1}{|\mathcal{P}|} \sum_{(u,v) \in \mathcal{P}} \sum_{k} q_k^{(u)} \log p_k^{(v)},
\end{equation}
where $p_k^{(v)} = [\operatorname{softmax}({a}_s^{(v)}/\tau_s)]_k$ and $q_k^{(u)} = \operatorname{sg}[\operatorname{softmax}(({a}_t^{(u)} - {c})/\tau_t)]_k$ and ${a}_s^{(v)}$ and ${a}_t^{(u)}$ are the student and teacher encoding of views $v$ and $u$, respectively. The running center $\mathbf{c}$ and temperature $\tau_t$ are used to prevent mode collapse.

\paragraph{Contrastive Learning.}
Contrastive frameworks pull representations of related views together while pushing unrelated samples apart. We evaluate \textbf{SoftCLT}~\citep{softclt}, which generalizes InfoNCE by replacing rigid negative assignments with soft weights, optimizing:
\begin{equation}
\label{eq:con_loss}
\frac{1}{4NT} \sum_{i=1}^{2N} \sum_{t=1}^{2T} \Big( \lambda \, \ell_I^{(i,t)} + (1-\lambda) \, \ell_T^{(i,t)} \Big),
\end{equation}
where $\ell_I$ is the instance-wise loss and $\ell_T$ is the soft temporal loss.

\paragraph{Joint-Embedding Predictive Architectures (JEPA).}
\label{main:leJEPA}
JEPA objectives predict target representations directly in latent space. We benchmark two formulations: \textbf{JEPA} and \textbf{LeJEPA}. Following \citet{lecun2022path} and \citet{balestriero2511lejepa}, our JEPA implementation uses a student network $f_\theta$ and predictor head $\operatorname{pred}_\psi$ to infer masked patch targets $\operatorname{sg}[f_{\theta'}(\mathbf{x})]$ generated by an exponential-moving-average (EMA) teacher $f_{\theta'}$, minimizing the alignment error alongside VICReg~\citep{bardes2022vicreg} variance and covariance penalties to prevent collapse:
\begin{equation}
\label{eq:jepa}
\frac{1}{|\mathcal{M}|} \sum_{i \in \mathcal{M}} \|\operatorname{pred}_\psi(f_\theta(\mathbf{x}_{\setminus \mathcal{M}}))_i - \operatorname{sg}[f_{\theta'}(\mathbf{x})_i]\|_2^2 + \lambda_{\mathrm{var}}\mathcal{L}_{\mathrm{var}} + \lambda_{\mathrm{cov}}\mathcal{L}_{\mathrm{cov}}.
\end{equation}
LeJEPA~\citep{balestriero2511lejepa} eliminates the EMA teacher by processing distinct augmented views through a single shared encoder $f_\theta$, preventing collapse via Sketched Isotropic Gaussian Regularization (SIGReg). SIGReg projects the high-dimensional latent representations along random 1D directions and matches their empirical characteristic functions to a standard normal distribution using an analytic statistical test $T(\cdot)$, optimizing:
\begin{equation}
\label{eq:LeJEPA}
\frac{1-\lambda}{L} \sum_{i=1}^{L} \|\mathbf{z}_i^{(1)} - \mathbf{z}_i^{(2)}\|_2^2 + \frac{\lambda}{2} \big[ T(\mathbf{z}^{(1)}) + T(\mathbf{z}^{(2)}) \big].
\end{equation}
While for DINO and SoftCLT using native augmentations yields stable results, we often observe collapse with LeJEPA. In order to support a fair comparison, we introduce an alternative augmentation protocol for LeJEPA based on the Discrete Wavelet Transform (DWT). We decompose the input signal into multi-scale approximation and detail coefficients and generate reference and perturbed views. A {reference view} applies soft-thresholding to detail coefficients while preserving low-frequency trends, whereas a {perturbed view} adds Gaussian noise. This encourages invariance to high-frequency sensor noise while preserving underlying temporal morphology. Implementation details and comparisons appear in Appendices~\ref{subsec:appendix_lejepa} and~\ref{subsec:appendix_augmentations}.

\begin{table}[t]
\caption{Taxonomy of SSL paradigms and representative methods evaluated in this work.}
\label{tab:objectives_inputs}
\centering
\small
\begin{tabular}{llcp{5.2cm}}
\toprule
\textbf{Paradigm} & \textbf{SSL Method} & \textbf{Target Space} & \textbf{Masking / Pretext Task} \\
\midrule
Reconstruction
 & MAE       & Signal                 & Random patch masking ($40\%$) \\
\midrule
\multirow{2}{*}{Generative}
 & NTP       & Signal                 & Causal masking, $h$ future patches \\
 & Diffusion & Signal                 & Denoising Gaussian corrupted signal from clean causal context \\
\midrule
Invariance-based
 & DINO      & Latent & Prediction of categorical probability distribution  \\
\midrule
Contrastive
 & SoftCLT   & Latent      & Distinguish similar pairs from dissimilar ones \\
 \midrule
\multirow{2}{*}{\shortstack[l]{Joint-Embedding\\Predictive Architectures}}
 & JEPA      & Latent                 & Latent prediction\\
 & Le-JEPA   & Latent                 & Latent prediction  \\
\bottomrule
\end{tabular}
\end{table}

\subsection{Evaluating SSL Transferability}
\label{subsec:evaluating_ssl_transfer}
We formalize the evaluation protocol used to assess the  transferability of learned representations.

\paragraph{Downstream Adaptation and Baseline Controls.}
We evaluate representations under three protocols: linear probing, probing with a Multi-Layer Perceptron (MLP), and full fine-tuning. In linear and MLP probing, the pre-trained encoder backbone remains frozen while only a downstream task head is trained, isolating the linear or non-linear accessibility of task-relevant features in the learned latent space. Under full fine-tuning, the encoder and task head are updated jointly, assessing the quality of the pre-trained weights as an optimization prior. In all cases, pre-training pretext heads are discarded prior to downstream adaptation. To isolate the empirical transfer benefit attributable strictly to self-supervised learning, every pre-trained model is paired with a matched random control, sharing an identical backbone architecture, parameter count, and downstream optimization budget:
\begin{itemize}
    \item \textit{Random Frozen Encoder (Probing Control):} A randomly initialized, frozen backbone paired with the same trained downstream head, measuring performance attainable solely from architecture inductive priors.
    \item \textit{Supervised from Scratch (Fine-Tuning Control):} An identical backbone randomly initialized and trained end-to-end with the downstream task head under supervised optimization.
\end{itemize}
\paragraph{Quantifying SSL Transfer Gain.}
To directly measure the net performance gain of pre-training relative to these baseline controls, we define the \emph{SSL transfer gain} $\Delta$ for a given task $\mathcal{T}$ and adaptation protocol:
\begin{equation}
\Delta(\mathcal{L}_{\mathrm{SSL}}, \mathcal{T})
= s_{\mathcal{T}} \left[ M(\theta_{\mathrm{SSL}}, \mathcal{T}) - M(\theta_{\mathrm{baseline}}, \mathcal{T}) \right],
\label{eq:transfer_gain}
\end{equation}
where $M(\theta, \mathcal{T})$ denotes the downstream evaluation metric, and $\theta_{\mathrm{SSL}}$ and $\theta_{\mathrm{baseline}}$ represent the pre-trained model and its corresponding control ($\theta_{\mathrm{baseline}}$ represents the frozen random encoder under probing, or supervised-from-scratch training under fine-tuning). The direction scalar $s_{\mathcal{T}} \in \{+1, -1\}$ aligns metric polarity: $s_{\mathcal{T}} = +1$ for higher-is-better metrics (e.g., accuracy) and $s_{\mathcal{T}} = -1$ for lower-is-better error metrics (e.g., mean squared error). Consequently, $\Delta(\mathcal{L}_{\mathrm{SSL}}, \mathcal{T}) > 0$ strictly indicates an empirical gain from pre-training over the baseline control.

\paragraph{Downstream Transfer Tasks and Metrics.}
We benchmark transfer across three core time-series task families:
\begin{itemize}
    \item \textit{Forecasting:} We report Mean Squared Error (MSE) and Mean Absolute Error (MAE), averaged across four prediction horizons $H \in \{96, 192, 336, 720\}$.
    \item \textit{Classification:} We report top-1 classification accuracy.
    \item \textit{Anomaly Detection:} Because metric choice substantially impacts model rankings, we report three complementary criteria: point-adjusted F1 ($\text{F1}_{\mathrm{Pa}}$), which credits an entire contiguous anomaly segment upon detecting any timestamp within it; event-wise F1 ($\text{F1}_{\mathrm{Ev}}$), which evaluates segment-level precision and recall; and affiliation F1 ($\text{F1}_{\mathrm{Af}}$), which measures directed temporal distance between predicted detections and ground-truth events (App~\ref{app:anom_metric}).
\end{itemize}
We give full details of every metric in Appendix~\ref{subsec:metrics}.

\section{Experimental Setup}

\paragraph{Pre-Training and Evaluation Benchmarks.}
Our primary experiments pre-train models on the Monash dataset~\citep{monash}, comprising approximately $680\text{K}$ time-series samples. To isolate the effects of data origin and scale on downstream transfer quality, we also pre-train on scale-matched synthetic data, as well as on a scaled-up corpus ($2.5\text{M}$ samples) consisting of either purely synthetic or mixed (Monash and synthetic) data. In addition, we evaluate an in-domain pre-training control for forecasting tasks. Synthetic data generation follows  TimePFN~\citep{timepfn} and is detailed in the Appendix~\ref{subsec:appendix_datasets}.

We evaluate on seven forecasting datasets (ETTh1/2, ETTm1/2~\citep{informer}, Weather,
Electricity, and Traffic~\citep{autoformer}), nine UEA multivariate classification
datasets~\citep{uea}, and five anomaly-detection datasets
(SMD~\citep{su2019robust},
PSM~\citep{abdulaal2021practical}, MSL and SMAP~\citep{hundman2018detecting},
and SWaT~\citep{mathur2016swat}), following the splits and evaluation
protocols of TSLib~\citep{tslib}.

\paragraph{Implementation Details.}
Unless otherwise specified, we use a shared 8-layer Transformer
with embedding dimension 128, 16 attention heads, context length 336,
patch size 16, and 20 pre-training epochs (full details in Appendix ~\ref{subsec:appendix_implement}). Main comparisons use
method-specific learning-rate and batch-size tuning with comparable
tuning budgets, retaining required heads, views, and optimization
components. Appendix~\ref{subsec:fair_eval} repeats the Monash comparison
with identical learning rate, batch size, epochs, and context.
Results report means and standard deviations over five seeds unless
specified otherwise. Extended results and implementation details are provided in the appendix.


\section{Results}
\label{sec:results}
We organize our findings in light of the three guiding research questions defined in the introduction (section headers give their shortened form).

\subsection{RQ1: When does SSL pre-training improve performance on downstream tasks?}
\label{sec:res_rq1}
Table~\ref{tab:control_mon} reports how often pre-trained SSL models outperform non-pre-trained baselines across linear probing, MLP probing, and fine-tuning. Figure~\ref{fig:rq1_dividend} details the dataset-level relative transfer gains over the frozen random baseline under linear probing. Across tasks, downstream transfer varies: anomaly detection benefits substantially, classification shows selective gains on specific datasets, and forecasting exhibits minimal to negative transfer.
 Absolute scores and standard deviations appear in
Tables~\ref{tab:forecasting_l8_monash_clean}, \ref{tab:classification_results_refined}, and \ref{tab:anomaly_detection_results} . We summarize the main trends per task below.

\paragraph{Anomaly detection.}
Anomaly detection shows the clearest transfer gain, with SSL outperforming the random baseline in 54--94\% of configurations under linear probing and up to 97\% under MLP probing (Table~\ref{tab:control_mon}). Relative gains are highest on MSL and SMAP (Figure~\ref{fig:rq1_dividend}a), where models such as JEPA and LeJEPA improve $\text{F1}_{\mathrm{Pa}}$ by more than 250\% over random features, driven by strong lifts in both $\text{F1}_{\mathrm{Af}}$ and $\text{F1}_{\mathrm{Ev}}$. However, these gains do not generalize to every benchmark: on SWaT, every SSL method underperforms random features on $\text{F1}_{\mathrm{Pa}}$ and $\text{F1}_{\mathrm{Af}}$, with only a marginal gain on event-wise detection (e.g., NTP at 2.86\% $\text{F1}_{\mathrm{Ev}}$).

\paragraph{Classification.}
SSL provides positive transfer on seven of nine classification benchmarks, though overall win rates remain moderate under frozen linear (46\%) and MLP (40\%) probes (Table~\ref{tab:control_mon}). In Figure~\ref{fig:rq1_dividend}b, gains are heavily concentrated in handwriting, digit, and gesture recognition. On the Handwriting dataset, several objectives exceed a 50\% relative improvement, with LeJEPA reaching +222\%. Conversely, datasets such as EthanolConcentration and JapaneseVowels show negligible or negative transfer across nearly all methods. Full fine-tuning raises the SSL win rate to 67\%, showing that in the majority of cases pre-trained weights offer a better starting point than random initialization once encoders can update their weights.

\begin{table}[tbh!]
\centering
\caption{Frequency of SSL outperforming non-pre-trained baselines across tasks and evaluation regimes, when pre-training on the Monash dataset. Each entry reports the count and percentage of objective--dataset pairs in which the pre-trained model beats the baseline. The baseline is a frozen random encoder for linear and MLP probing, and supervised training from scratch for fine-tuning.}
\label{tab:control_mon}
\small
\setlength{\tabcolsep}{5pt}
\begin{tabular}{l rr rr rr}
\toprule
 & \multicolumn{2}{c}{Linear probing} & \multicolumn{2}{c}{MLP probing} & \multicolumn{2}{c}{Fine-tuning} \\
\cmidrule(lr){2-3} \cmidrule(lr){4-5} \cmidrule(lr){6-7}
Task & Count & \% & Count & \% & Count & \% \\
\midrule
\multicolumn{7}{l}{\textit{Anomaly detection}} \\
\quad Point-adjusted F1  {($\text{F1}_{\mathrm{PA}}$)} & 19/35 & 54 & 26/35 & 74 & 21/35 & 60 \\
\quad Event-wise F1 {($\text{F1}_{\mathrm{Ev}}$)}      & 33/35 & 94 & 34/35 & 97 & 24/35 & 69 \\
\quad Affiliation F1  {($\text{F1}_{\mathrm{Af}}$)}   & 19/35 & 54 & 29/35 & 83 & 24/35 & 69 \\
\midrule
\multicolumn{7}{l}{\textit{Classification}} \\
\quad Accuracy            & 29/63 & 46 & 25/63 & 40 & 42/63 & 67 \\
\midrule
\multicolumn{7}{l}{\textit{Forecasting}} \\
\quad MSE                 &  9/49 & 18 & 11/35 & 31 &  3/35 &  9 \\
\quad MAE                 &  8/49 & 16 & 11/35 & 31 &  2/35 &  6 \\
\bottomrule
\end{tabular}
\end{table}

\paragraph{Forecasting.}
Pre-training offers minimal advantage for forecasting. Under linear probing, pre-trained models beat the random frozen baseline in only 18\% of MSE evaluations (Table~\ref{tab:control_mon}), with most models yielding negative relative gains across the ETT, Weather, Electricity, and Traffic benchmarks (Figure~\ref{fig:rq1_dividend}c). Absolute MSE improvements are marginal, while LeJEPA severely degrades across all forecasting datasets (up to $-101\%$ relative to the baseline). Switching to an MLP probe raises the win rate to 31\%. Under full fine-tuning, however, training directly from scratch outperforms pre-trained backbones in over 90\% of runs, including all four ETT datasets.

\begin{figure}[tbh!]
  \centering
\begin{tikzpicture}[font=\figfontS]
\node[anchor=south west, font=\figfont\bfseries, inner sep=1pt] at ({0*\hmw},3pt) {(a) Anomaly detection};
\hmrow{0}{0}{+83.6,+15.7,+184,+10.1,-2.9} 
\hmrow{1}{0}{+88.6,-23.6,+239,+8.9,-34.8} 
\hmrow{2}{0}{+263,-7.9,+216,-7.2,-27.4} 
\hmrow{3}{0}{+60.6,+49.8,+156,+10.9,-1.4} 
\hmrow{4}{0}{+266,-45.7,+336,-6.3,-73.5} 
\hmrow{5}{0}{+266,-0.7,+334,-6.0,-74.1} 
\hmrow{6}{0}{+94.6,-5.8,+270,-7.2,-4.4} 
\node[anchor=north east, rotate=45, inner sep=1pt] at ({(0+0+0.75)*\hmw},{-7*\hmh-2pt}) {MSL};
\node[anchor=north east, rotate=45, inner sep=1pt] at ({(0+1+0.75)*\hmw},{-7*\hmh-2pt}) {PSM};
\node[anchor=north east, rotate=45, inner sep=1pt] at ({(0+2+0.75)*\hmw},{-7*\hmh-2pt}) {SMAP};
\node[anchor=north east, rotate=45, inner sep=1pt] at ({(0+3+0.75)*\hmw},{-7*\hmh-2pt}) {SMD};
\node[anchor=north east, rotate=45, inner sep=1pt] at ({(0+4+0.75)*\hmw},{-7*\hmh-2pt}) {SWaT};

\node[anchor=south west, font=\figfont\bfseries, inner sep=1pt] at ({5.8*\hmw},3pt) {(b) Classification};
\hmrow{0}{5.8}{-9.9,+6.0,+45.2,+0.3,-3.7,-2.6,+5.8,+13.8,+50.6} 
\hmrow{1}{5.8}{-11.3,-0.5,+17.4,-6.9,-7.3,-1.3,+3.3,+5.0,-5.5} 
\hmrow{2}{5.8}{-11.8,+2.3,+56.1,-3.1,-2.6,-10.9,+4.7,+9.0,+37.1} 
\hmrow{3}{5.8}{-1.4,-5.0,+154,+8.4,-19.6,-5.6,+8.2,-6.5,-16.6} 
\hmrow{4}{5.8}{-10.4,-4.1,+37.6,-5.6,-0.7,-14.9,+9.8,+2.3,-10.7} 
\hmrow{5}{5.8}{-5.7,+13.1,+222,-6.1,-6.5,+8.3,-2.2,+20.2,+96.1} 
\hmrow{6}{5.8}{-1.6,-4.8,+84.0,+9.5,-18.4,-18.8,+6.2,-3.5,+33.0} 
\node[anchor=north east, rotate=45, inner sep=1pt] at ({(5.8+0+0.75)*\hmw},{-7*\hmh-2pt}) {Ethanol};
\node[anchor=north east, rotate=45, inner sep=1pt] at ({(5.8+1+0.75)*\hmw},{-7*\hmh-2pt}) {FaceDet};
\node[anchor=north east, rotate=45, inner sep=1pt] at ({(5.8+2+0.75)*\hmw},{-7*\hmh-2pt}) {Handwrit};
\node[anchor=north east, rotate=45, inner sep=1pt] at ({(5.8+3+0.75)*\hmw},{-7*\hmh-2pt}) {Heartbt};
\node[anchor=north east, rotate=45, inner sep=1pt] at ({(5.8+4+0.75)*\hmw},{-7*\hmh-2pt}) {JapVowel};
\node[anchor=north east, rotate=45, inner sep=1pt] at ({(5.8+5+0.75)*\hmw},{-7*\hmh-2pt}) {SCP1};
\node[anchor=north east, rotate=45, inner sep=1pt] at ({(5.8+6+0.75)*\hmw},{-7*\hmh-2pt}) {SCP2};
\node[anchor=north east, rotate=45, inner sep=1pt] at ({(5.8+7+0.75)*\hmw},{-7*\hmh-2pt}) {Digits};
\node[anchor=north east, rotate=45, inner sep=1pt] at ({(5.8+8+0.75)*\hmw},{-7*\hmh-2pt}) {UWave};

\node[anchor=south west, font=\figfont\bfseries, inner sep=1pt] at ({15.6*\hmw},3pt) {(c) Forecasting};
\hmrow{0}{15.6}{-5.1,-3.4,+1.1,-0.4,+5.2,-4.7,-2.3} 
\hmrow{1}{15.6}{-5.8,-5.1,-2.2,-1.1,+1.6,-12.8,-10.6} 
\hmrow{2}{15.6}{-4.6,+4.6,-5.3,-1.1,+2.4,-32.0,-21.1} 
\hmrow{3}{15.6}{-7.2,-5.1,-1.1,+2.7,+4.8,-5.8,-3.2} 
\hmrow{4}{15.6}{-4.6,-3.7,-2.2,-3.4,+1.2,-3.5,-1.6} 
\hmrow{5}{15.6}{-48.8,-12.6,-24.7,-10.3,-5.6,-101,-94.9} 
\hmrow{6}{15.6}{-10.1,-8.0,-6.7,-1.9,+2.0,-26.2,-12.2} 
\node[anchor=north east, rotate=45, inner sep=1pt] at ({(15.6+0+0.75)*\hmw},{-7*\hmh-2pt}) {ETTh1};
\node[anchor=north east, rotate=45, inner sep=1pt] at ({(15.6+1+0.75)*\hmw},{-7*\hmh-2pt}) {ETTh2};
\node[anchor=north east, rotate=45, inner sep=1pt] at ({(15.6+2+0.75)*\hmw},{-7*\hmh-2pt}) {ETTm1};
\node[anchor=north east, rotate=45, inner sep=1pt] at ({(15.6+3+0.75)*\hmw},{-7*\hmh-2pt}) {ETTm2};
\node[anchor=north east, rotate=45, inner sep=1pt] at ({(15.6+4+0.75)*\hmw},{-7*\hmh-2pt}) {Weather};
\node[anchor=north east, rotate=45, inner sep=1pt] at ({(15.6+5+0.75)*\hmw},{-7*\hmh-2pt}) {Electricity};
\node[anchor=north east, rotate=45, inner sep=1pt] at ({(15.6+6+0.75)*\hmw},{-7*\hmh-2pt}) {Traffic};

\node[anchor=east, inner sep=2pt] at (0,{-(0+0.5)*\hmh}) {NTP};
\node[anchor=east, inner sep=2pt] at (0,{-(1+0.5)*\hmh}) {MAE};
\node[anchor=east, inner sep=2pt] at (0,{-(2+0.5)*\hmh}) {Diffusion};
\node[anchor=east, inner sep=2pt] at (0,{-(3+0.5)*\hmh}) {DINO};
\node[anchor=east, inner sep=2pt] at (0,{-(4+0.5)*\hmh}) {JEPA};
\node[anchor=east, inner sep=2pt] at (0,{-(5+0.5)*\hmh}) {Le-JEPA};
\node[anchor=east, inner sep=2pt] at (0,{-(6+0.5)*\hmh}) {SoftCLT};

\begin{scope}[shift={({6.3*\hmw},{-7*\hmh-46pt})}]
  \shade[left color=hmneg!\hmmax!hmzero, right color=hmzero] (0,0) rectangle (5*\hmw,5pt);
  \shade[left color=hmzero, right color=hmpos!\hmmax!hmzero] (5*\hmw,0) rectangle (10*\hmw,5pt);
  \foreach \t/\l in {0/-300,1/-200,2/-100,3/0,4/100,5/200,6/300}
    {\draw[figgray, line width=.4pt] ({\t*10*\hmw/6},0) -- ++(0,-2pt) node[below, inner sep=1pt] {$\mathsf{\l}$};}
  \node[below, inner sep=1pt] at (5*\hmw,-10pt) {Relative improvement over random frozen encoder (\%)};
\end{scope}
\end{tikzpicture}
  \caption{Relative SSL transfer gain (\%) over a frozen random baseline under linear probing across downstream tasks (Monash pre-training). The relative gain quantifies the improvement compared to the respective random control, derived from Eq.~\ref{eq:transfer_gain} as $100 \times \frac{\Delta(\mathcal{L}_{\mathrm{SSL}}, \mathcal{T})}{M(\theta_{\mathrm{baseline}},\mathcal{T})}$.  Positive values indicate performance gains: MSE reduction for forecasting, a higher accuracy for classification, and a higher $\text{F1}_{\mathrm{Pa}}$ scores for anomaly detection. Rounded values are computed from reported five-seed means.}
  \label{fig:rq1_dividend}
\end{figure}

\subsection{RQ2: Do certain SSL methods consistently outperform others?}
\label{sec:res_rq2}
No single SSL method consistently outperforms the others. Instead, two main findings characterize the empirical results:
\begin{enumerate}[label=(\alph*)]
    \item \textit{Under linear probing, performance splits by task target}:
when evaluating with a linear probe, relative rankings depend on whether the downstream objective requires identifying discrete semantic states (classification and segment-level detection) or tracking continuous numerical trajectories (forecasting and point-wise residual scoring). LeJEPA achieves the highest accuracy on five of nine classification datasets (Handwriting, UWave, FaceDetection, SCP1, SpokenArabicDigits) and leads on MSL anomaly detection on both $\text{F1}_{\mathrm{Pa}}$ and $\text{F1}_{\mathrm{Af}}$, where identifying a coarse regime shift suffices. Yet on forecasting, it never beats a frozen random encoder. A likely reason is that its regularizer (SIGReg) separates discrete classes well, but discards the point-wise continuity needed to extrapolate trajectories (Appendix~\ref{app:lejepa_analysis}). Conversely, reconstruction and generative objectives perform better on continuous numerical tasks: Diffusion leads on ETTh2, and NTP leads on ETTm1 and on SMD (F1$_{\text{Af}}$). We also note that for anomaly detection, SSL performance and relative ranking vary depending on the evaluation metric (Fig.~\ref{fig:rq2_evaluation}a).
\item \textit{Linear probing does not predict MLP or fine-tuning performance}:
The ranking of SSL models under linear probing changes substantially with non-linear probing (MLP) or when unfreezing the encoder (Fig.~\ref{fig:rq2_evaluation}b). In forecasting, Diffusion loses to a random encoder when using a linear head, but achieves the lowest MSE on roughly 90\% of datasets when evaluated with a shallow MLP (with the remainder 10\% within seed variance). The same trend is observed for fine-tuning. After fine-tuning, LeJEPA retains its classification advantage on only two datasets, as competing models close the performance gap in anomaly detection, Diffusion trails the random baseline on PSM under linear probing (43.78\% vs.\ 47.55\%), but beats every other SSL model after fine-tuning (48.30\% vs.\ 28.29\%). This suggests that while strongly regularized embeddings yield competitive frozen features, other representations serve as better starting points for fine-tuning.
\end{enumerate}

\begin{figure}[tbh!]
  \centering
\begin{tikzpicture}
\begin{scope}[name=A, font=\figfontS]
\rkrow{0}{0}{8,8,8,3,8,3,8,8,8,4,7,4,1,7,1} 
\rkrow{1}{0}{6,3,6,2,6,2,6,5,6,2,5,1,3,1,5} 
\rkrow{2}{0}{5,1,5,7,7,6,4,1,4,3,6,3,6,6,4} 
\rkrow{3}{0}{3,4,3,6,4,5,5,6,5,7,4,8,5,5,6} 
\rkrow{4}{0}{7,6,7,1,2,1,7,7,7,1,8,2,2,3,3} 
\rkrow{5}{0}{2,6,2,8,5,8,1,3,1,6,1,6,7,4,7} 
\rkrow{6}{0}{1,5,1,4,1,4,2,4,2,5,2,5,8,8,8} 
\rkrow{7}{0}{4,2,4,5,3,7,3,1,3,8,3,7,4,2,2} 
\node[anchor=east, inner sep=2pt] at (0,{-(0+0.5)*\rkw}) {Random};
\node[anchor=east, inner sep=2pt] at (0,{-(1+0.5)*\rkw}) {NTP};
\node[anchor=east, inner sep=2pt] at (0,{-(2+0.5)*\rkw}) {MAE};
\node[anchor=east, inner sep=2pt] at (0,{-(3+0.5)*\rkw}) {Diffusion};
\node[anchor=east, inner sep=2pt] at (0,{-(4+0.5)*\rkw}) {DINO};
\node[anchor=east, inner sep=2pt] at (0,{-(5+0.5)*\rkw}) {JEPA};
\node[anchor=east, inner sep=2pt] at (0,{-(6+0.5)*\rkw}) {Le-JEPA};
\node[anchor=east, inner sep=2pt] at (0,{-(7+0.5)*\rkw}) {SoftCLT};
\node[anchor=south, inner sep=2pt] at ({(0.0+1.5)*\rkw},0) {MSL};
\node[anchor=north, inner sep=2pt, font=\figfontXS] at ({(0.0+0+0.5)*\rkw},{-8*\rkw}) {PA};
\node[anchor=north, inner sep=2pt, font=\figfontXS] at ({(0.0+1+0.5)*\rkw},{-8*\rkw}) {Ev};
\node[anchor=north, inner sep=2pt, font=\figfontXS] at ({(0.0+2+0.5)*\rkw},{-8*\rkw}) {Af};
\node[anchor=south, inner sep=2pt] at ({(3.35+1.5)*\rkw},0) {PSM};
\node[anchor=north, inner sep=2pt, font=\figfontXS] at ({(3.35+0+0.5)*\rkw},{-8*\rkw}) {PA};
\node[anchor=north, inner sep=2pt, font=\figfontXS] at ({(3.35+1+0.5)*\rkw},{-8*\rkw}) {Ev};
\node[anchor=north, inner sep=2pt, font=\figfontXS] at ({(3.35+2+0.5)*\rkw},{-8*\rkw}) {Af};
\node[anchor=south, inner sep=2pt] at ({(6.7+1.5)*\rkw},0) {SMAP};
\node[anchor=north, inner sep=2pt, font=\figfontXS] at ({(6.7+0+0.5)*\rkw},{-8*\rkw}) {PA};
\node[anchor=north, inner sep=2pt, font=\figfontXS] at ({(6.7+1+0.5)*\rkw},{-8*\rkw}) {Ev};
\node[anchor=north, inner sep=2pt, font=\figfontXS] at ({(6.7+2+0.5)*\rkw},{-8*\rkw}) {Af};
\node[anchor=south, inner sep=2pt] at ({(10.05+1.5)*\rkw},0) {SMD};
\node[anchor=north, inner sep=2pt, font=\figfontXS] at ({(10.05+0+0.5)*\rkw},{-8*\rkw}) {PA};
\node[anchor=north, inner sep=2pt, font=\figfontXS] at ({(10.05+1+0.5)*\rkw},{-8*\rkw}) {Ev};
\node[anchor=north, inner sep=2pt, font=\figfontXS] at ({(10.05+2+0.5)*\rkw},{-8*\rkw}) {Af};
\node[anchor=south, inner sep=2pt] at ({(13.4+1.5)*\rkw},0) {SWaT};
\node[anchor=north, inner sep=2pt, font=\figfontXS] at ({(13.4+0+0.5)*\rkw},{-8*\rkw}) {PA};
\node[anchor=north, inner sep=2pt, font=\figfontXS] at ({(13.4+1+0.5)*\rkw},{-8*\rkw}) {Ev};
\node[anchor=north, inner sep=2pt, font=\figfontXS] at ({(13.4+2+0.5)*\rkw},{-8*\rkw}) {Af};
\node[anchor=south west, font=\figfont\bfseries, inner sep=2pt] at (0,8pt) {(a)};
\end{scope}

\begin{axis}[sslaxis, name=B, at={($(0,0)+(16.5*\rkw+48pt,0)$)}, anchor=north west,
  width=0.37\linewidth, height=0.34\linewidth, scale only axis,
  xmin=-22, xmax=42, ymin=-16, ymax=42, xtick={-20,0,20,40}, ytick={-10,0,20,40},
  xlabel={Linear probing transfer (pp)}, ylabel={Full fine-tuning transfer (pp)}]
\draw[figgray!60, line width=.4pt] (axis cs:-22,0)--(axis cs:42,0) (axis cs:0,-16)--(axis cs:0,42);
\draw[figgray!70, dashed, line width=.4pt] (axis cs:-16,-16)--(axis cs:42,42);
\addplot[mNTP] coordinates {(-2.74,-0.92) (3.27,1.91) (1.95,13.52) (0.20,5.24) (-3.40,-0.84) (-1.97,7.58) (2.89,-1.60) (10.68,1.09) (18.50,24.35)}; 
\addplot[mMAE] coordinates {(-3.12,1.94) (-0.29,0.06) (0.75,6.52) (-4.58,5.61) (-6.75,0.65) (-1.02,8.43) (1.66,0.34) (3.90,1.51) (-2.00,30.28)}; 
\addplot[mDiffusion] coordinates {(-3.27,-2.34) (1.23,6.21) (2.42,0.73) (-2.04,-5.98) (-2.43,-3.61) (-8.39,3.48) (2.33,3.26) (6.95,1.92) (13.56,28.25)}; 
\addplot[mDINO] coordinates {(-0.38,-0.63) (-2.72,15.34) (6.63,5.72) (5.57,3.78) (-18.21,-12.94) (-4.30,7.32) (4.11,3.81) (-5.05,0.90) (-6.06,30.83)}; 
\addplot[mJEPA] coordinates {(-2.89,-2.15) (-2.21,0.99) (1.62,3.11) (-3.70,-7.68) (-0.64,-3.34) (-11.46,-7.27) (4.89,4.92) (1.76,2.36) (-3.93,-3.78)}; 
\addplot[mLeJEPA] coordinates {(-1.59,-2.05) (7.16,1.34) (9.55,20.55) (-4.09,-4.88) (-6.04,-3.21) (6.35,-0.53) (-1.11,-1.75) (15.66,1.17) (35.13,37.08)}; 
\addplot[mSoftCLT] coordinates {(-0.45,7.17) (-2.59,3.72) (3.62,7.02) (6.34,3.90) (-17.13,-8.07) (-14.40,-6.75) (3.11,1.58) (-2.68,-6.73) (12.07,6.22)}; 
\draw[cDINO, line width=.5pt] ([xshift=-1.2pt,yshift=2pt]axis cs:-2.72,15.34) -- (axis cs:-12,36) node[above, inner sep=1pt, font=\figfontS, text=black] {DINO / FaceDet};
\draw[cLeJEPA, line width=.5pt] ([xshift=-2pt,yshift=1pt]axis cs:35.13,37.08) -- (axis cs:22,39.5) node[left, inner sep=1pt, font=\figfontS, text=black] {Le-JEPA / UWave};
\end{axis}

\node[anchor=south west, font=\figfont\bfseries, inner sep=2pt] at (B.north west |- 0,8pt) {(b)};

\end{tikzpicture}

\vspace{2pt}\centering\objlegend
\caption{Downstream performance across evaluation protocols and metrics (pretraining on the Monash dataset).
(a) SSL method performance rankings on Anomaly detection under linear probing, across point-adjusted (PA), event-wise (Ev), and affiliation (Af) F1. Ranks 1 to 8 include the seven SSL objectives and the random control; ties receive equal rank. 
(b) Classification gains in percentage points for all 63 objective--dataset pairs. Controls are a frozen random encoder for linear probing and supervised training from scratch for full fine-tuning. Solid axes mark baseline performance; the dashed diagonal indicates equal gain under both regimes.
Complete scores and standard deviations are reported in Appendix 
Tables~\ref{tab:classification_results_refined}, \ref{tab:classification_finetuning_results_with_baseline}, and~\ref{tab:anomaly_detection_results}.}
  \label{fig:rq2_evaluation}
\end{figure}

\subsection{RQ3: How Do Data Source, Scale, and Depth Affect Transfer?}
\label{sec:res_rq3}
We find that scaling pre-training data volume, training duration, and model depth does not yield consistent gains. Instead, transfer efficacy depends heavily on the alignment between the pre-training source and the downstream task structure.

\paragraph{Data source and domain relevance.}
Pairwise comparisons across all objective--dataset pairs show that the best pre-training source depends on the task (Table ~\ref{tab:comp_dataset}). For forecasting, in-domain data performs best, beating Monash  and both synthetic sets. Compared to the scale-matched synthetic dataset, Monash gives a slight advantage, but adding synthetic data to Monash hurts performance. Mixing in synthetic data appears to wash out the real temporal patterns needed to forecast future points.

For classification, this pattern flips. Scale-matched synthetic data beats Monash and large synthetic data. A likely reason is that clean synthetic curves expose basic class shapes without adding irrelevant sensor noise, making classes easier to separate.
For anomaly detection, the preferred source depends on the metric. Pretraining on Monash yields better $\text{F1}_{\mathrm{Ev}}$, compared to the matched-scale synthetic dataset, and vice versa for $\text{F1}_{\mathrm{Af}}$.

\begin{table}[t]
\centering
\caption{Pairwise comparison of pre-training sources under linear probing. For each objective--dataset pair (49 for forecasting, 63 for classification, 35 per anomaly metric), entries report wins/losses/ties for Source A (row) against Source B (column); an entry is bolded when Source A achieves more wins than Source B. For example, \textbf{31/16/2} for Monash (680K) versus Synthetic (2.5M) on forecasting means that models pre-trained on Monash outperform those pre-trained on Synthetic (2.5M) in 31 of 49 objective--dataset comparisons, underperform in 16, and tie in 2. In-domain pre-training is evaluated for forecasting only.}
\label{tab:comp_dataset}
\small
\setlength{\tabcolsep}{4.5pt}
\begin{tabular}{l l c c ccc}
\toprule
 & & Forecasting & Classification & \multicolumn{3}{c}{Anomaly detection} \\
\cmidrule(lr){3-3} \cmidrule(lr){4-4} \cmidrule(lr){5-7}
Source A & Source B & MSE & Accuracy & $\mathrm{F1}_{\mathrm{PA}}$ & $\mathrm{F1}_{\mathrm{Ev}}$ & $\mathrm{F1}_{\mathrm{Af}}$ \\
\midrule
Monash (680K)    & Synthetic (680K) & \textbf{25/23/1} & 26/37/0          & 17/17/1          & \textbf{25/9/1}  & 16/18/1          \\
                 & Synthetic (2.5M) & \textbf{30/17/2} & 31/32/0          & 17/17/1          & \textbf{24/10/1} & \textbf{19/15/1} \\
                 & Hybrid (2.5M)    & \textbf{29/16/4} & 29/34/0          & 16/16/3          & \textbf{22/12/1} & 17/18/0          \\
\addlinespace[3pt]
Synthetic (680K) & Synthetic (2.5M) & \textbf{31/14/4} & \textbf{34/29/0} & 14/19/2          & 13/22/0          & \textbf{19/16/0} \\
                 & Hybrid (2.5M)    & \textbf{33/15/1} & \textbf{34/29/0} & \textbf{17/16/2} & 11/24/0          & \textbf{18/17/0} \\
\addlinespace[3pt]
Synthetic (2.5M) & Hybrid (2.5M)    & 20/24/5          & 27/35/1          & \textbf{18/16/1} & 13/22/0          & 15/20/0          \\
\midrule
In-domain        & Monash (680K)    & \textbf{34/11/4} & -- & -- & -- & -- \\
                 & Synthetic (680K) & \textbf{35/14/0} & -- & -- & -- & -- \\
                 & Synthetic (2.5M) & \textbf{42/7/0}  & -- & -- & -- & -- \\
                 & Hybrid (2.5M)    & \textbf{41/6/2}  & -- & -- & -- & -- \\
\bottomrule
\end{tabular}
\end{table}

\paragraph{Training duration and encoder depth.}
Extended pre-training yields negligible net benefits (Table~\ref{tab:epochs_depth}, top panel). Doubling pre-training from 20 to 40 epochs does not change classification and degrades raw forecasting performance. While 40 epochs appears to favor anomaly detection, this advantage vanishes once filtering out seed-level variance ($\le 0.005$ MSE or $\le 1\%$ accuracy/F1), leaving mostly ties between the two training budgets.

\begin{table}[t]
\centering
\caption{Training duration and encoder depth (Monash pre-training, linear
probing). For each objective--dataset pair, we count whether the first setting
beats the second, and report wins, losses, and ties for the first setting; the
larger of wins and losses is in bold. The depth comparisons are single-seed.}
\label{tab:epochs_depth}
\small
\setlength{\tabcolsep}{6pt}
\begin{tabular}{l l l c ccc}
\toprule
Comparison & Task & Metric & Pairs & Wins & Losses & Ties \\
\midrule
40 vs.\ 20 epochs 40 vs.\ 20 epochs  & Forecasting       & MSE                         & 49 & 13          & \textbf{28} & 8 \\
                   & Classification    & Accuracy                    & 63 & 30          & \textbf{33} & 0 \\
                   & Anomaly detection & $\mathrm{F1}_{\mathrm{PA}}$ & 35 & \textbf{20} & 14          & 1 \\
                   &                   & $\mathrm{F1}_{\mathrm{Ev}}$ & 35 & \textbf{18} & 17          & 0 \\
                   &                   & $\mathrm{F1}_{\mathrm{Af}}$ & 35 & \textbf{21} & 14          & 0 \\
\midrule
2 vs.\ 8 layers    & Forecasting       & MSE                         & 21 & \textbf{14} & 7           & 0 \\
4 vs.\ 8 layers    &                   &                             & 21 & \textbf{16} & 5           & 0 \\
12 vs.\ 8 layers   &                   &                             & 21 & 6           & \textbf{15} & 0 \\
24 vs.\ 8 layers   &                   &                             & 21 & 3           & \textbf{18} & 0 \\
\bottomrule
\end{tabular}
\end{table}

Increasing the encoder depth degrades forecasting accuracy (Table~\ref{tab:epochs_depth}, bottom panel). In a single-seed sweep over $N \in \{2, 4, 8, 12, 24\}$ layers, a 2-layer backbone outperforms an 8-layer encoder on forecasting, while the 24-layer variant outperforms the 8-layer model in only 3 of 21 comparisons. In classification, the 8-layer backbone similarly outperforms the 24-layer model. Under fixed training budgets, shallower networks consistently match or exceed deep architectures, indicating that parameter scaling practices from vision and NLP do not transfer directly to current time-series SSL objectives, at least under the dataset scales investigated in our work.
\section{Practical Takeaways}
\label{sec:discussion}

We summarize the key findings, recommended practices and scope limitations in Table~\ref{tab:lessons_recommendations}.

\begin{table}[tbh!]
\caption{Key empirical findings and recommended evaluation practices.}
\label{tab:lessons_recommendations}
\centering
\small
\begin{tabular}{p{3.8cm}p{5.6cm}p{6.0cm}}
\toprule
Finding (RQ) & Evaluation Practice & Implications and Scope \\
\midrule
SSL gains are strongest in anomaly detection and selective in classification \newline (RQ1)
 & Prioritize SSL pre-training for tasks requiring regime-shift or class separation rather than point trajectory extrapolation. Evaluate fine-tuning to measure full transfer potential.
 & Pre-trained representations reliably benefit anomaly detection and provide superior initializations for classification (reaching a 67\% win rate under fine-tuning), even though linear probes on frozen features show more moderate gains. \\
\midrule
Forecasting shows limited gains over non-pre-trained controls \newline (RQ1)
 & Evaluate representations under linear probing and full fine-tuning rather than zero-shot transfer alone. Include both a frozen random encoder and supervised training from scratch controls across multiple seeds.
 & Zero-shot protocols do not evaluate adaptation capacity, as baselines are denied access to target supervision. Under linear probing and fine-tuning, pre-trained backbones underperform their randomly initialized controls in the majority of evaluated configurations. \\
\midrule
Linear probing rankings do not reliably predict fine-tuning outcomes \newline (RQ2)
 & Evaluate non-linear probing (MLP) and end-to-end fine-tuning alongside standard linear probes.
 & Linear probing can misrepresent downstream utility: some models require an MLP head or fine-tuning to show gains (e.g., Diffusion in forecasting), while others lose their relative lead once all models are fine-tuned (e.g., LeJEPA in classification).\\
\midrule
Anomaly detection rankings vary substantially across evaluation metrics \newline (RQ2)
 & Report event-wise ($\text{F1}_{\mathrm{Ev}}$) and affiliation ($\text{F1}_{\mathrm{Af}}$) scores alongside point-adjusted ($\text{F1}_{\mathrm{PA}}$) metrics.
 & Point-adjusted F1 can mask poor event localization, as predicting a single anomalous time step marks the entire segment as correctly identified. \\
\midrule
Target-domain proximity can outweigh corpus scale in forecasting \newline (RQ3)
 & Prioritize domain-aligned data over larger, uncurated collections. Assess whether synthetic data aids or degrades transfer before expanding pre-training volume.
 & In-domain pre-training consistently provides the strongest forecasting performance. This may indicate that forecasting relies heavily on dataset-specific temporal patterns that transfer poorly across domains, with less universal shared structure than in language or vision. Whether substantially larger multi-domain corpora reverse this trend remains an open empirical question. \\ \\
\midrule
Increasing encoder depth and training duration shows diminishing returns or performance degradation \newline (RQ3)
 & Benchmark deeper configurations against shallow backbones (e.g., 2--4 layers) and test whether improvements exceed seed variance (e.g., $\pm 0.005$ MSE / $\pm 1\%$ F1) before expanding compute.
 & Our analysis evaluates standard Transformer encoders. Alternative model families (such as state-space models or convolutional networks) may exhibit different parameter-scaling behavior under self-supervised objectives. \\
\bottomrule
\end{tabular}
\end{table}
\section{Conclusion}\label{sec:conclusion}
We evaluated seven time-series SSL methods across forecasting, classification, and anomaly detection. Downstream transfer varies by task and evaluation protocol, with clear benefits for anomaly detection, selective gains for classification, and limited added value over non-pre-trained baselines in forecasting. Linear probing does not reliably predict fine-tuning performance, and no single SSL method dominates across all tasks. Furthermore, synthetic pre-training matches Monash on several benchmarks, whereas scaling Transformer depth and pre-training duration does not improve performance.

Our findings are bounded by the evaluated Transformer backbones and pre-training dataset scales. Key priorities for future work include evaluating scaling to larger multi-domain corpora, evaluating on additional architectures and developing hybrid objectives that balance semantic invariance with point-wise precision.

\bibliography{ts}
\bibliographystyle{iclr2027_conference}

\newpage
\appendix
\section*{Appendix}
    
\label{sec:appendix}
We provide extended results and discussion as part of this work, this includes detailed documentation of our experimental setup to support reproducibility:
\begin{itemize}
    \item Section~\ref{sec:appendix_objectives} gives the SSL objectives
    \item Section~\ref{subsec:appendix_benchmarks_and_metrics} gives the metrics and baselines
    \item Section~\ref{sec:appendix_exp_setup} gives extended details of our experimental setup, encompassing:
    \begin{itemize}
        \item Pretraining datasets details (Section~\ref{subsec:appendix_datasets})
        \item Synthetic data generation details (Section~\ref{subsec:gen}).
        \item Evaluation data details (Section~\ref{subsec:eval_data}).
        \item Additional implementation and training Details (Section~\ref{subsec:appendix_implement})
        \item Compute and train time details (Section~\ref{subset:appendix_compute})
    \end{itemize}
    \item Section~\ref{app:results} gives the extended results to our three research questions:
    \begin{itemize}
        \item Extended results for \textit{Monash} pre-training under all three protocols (Section~\ref{subsec:monash_main})
        \item Extended results for source and scale experiments (Section~\ref{subsec:source_scale})
    \end{itemize}
    \item  Section~\ref{sec:DDc} provides results for extended training, enlarged depth, and convergence analysis.
    \begin{itemize}
        \item Extended results for 40 epoch Evaluation (Section~\ref{sec:40_ep})
        \item Extended results for transformer depth scaling (Section~\ref{subsec:appendix_depth_scaling})
        \item Convergence analysis (Section~\ref{subsec:conv_anl})
    \end{itemize}
    \item Section~\ref{app:ablations} Gives extended results for our controlled ablations:
    \begin{itemize}
        \item Augmentations ablation study for LeJEPA (Section~\ref{subsec:appendix_augmentations})
        \item Equal hyperparameter evaluation (Section~\ref{subsec:fair_eval})
        \item LeJEPA ablations (Section~\ref{subsubsec:LeJEPA_ABB})
        \item Representation Analyses (Section~\ref{app:representations})
    \end{itemize}

\end{itemize}

\section{SSL Objective Definitions}
\label{sec:appendix_objectives}

\newcommand{\mlink}[2]{\hyperref[#1]{\makebox[1.45cm][l]{#2}}}

We complement the objective definitions of Section~\ref{subsec:ssl_paradigms} with implementation details and, where the main text states only the final loss, the full formulation. Subsections follow the paradigm order of Section~\ref{subsec:ssl_paradigms}, and each method is listed with its main-text equation. For clarity, the channel index is suppressed; the encoder operates independently on each channel.

\subsection{Reconstruction-Based SSL}
\paragraph{Masked Autoencoding (MAE; Eq.~\ref{eq:MAE}).}
We mask 40\% of patches uniformly at random, independently per channel and window. Masked patches are set to zero before encoding. The decoder is a single linear layer applied per patch ($D{=}128 \rightarrow P{=}16$, preceded by dropout), and the squared error is averaged within each patch and then over masked patches only. MAE denotes masked autoencoding throughout; mean absolute error is a separate downstream forecasting metric.

\subsection{Generative SSL}
\paragraph{Next-Patch Prediction (NTP; Eq.~\ref{eq:NTP}).}
Each window of $L=L_c+h$ patches is split into $L_c=21$ context and $h=6$ target patches. The encoder sees only the context under causal attention, and the residual MLP head $\psi_h$ maps the last $h$ context representations to all $h$ future patches in a single pass. The squared error is averaged over predicted patches, channels, and within-patch values; targets are continuous values rather than discrete tokens.

\paragraph{Diffusion (Eq.~\ref{eq:Diff}).}
Each patch receives an independently sampled step $t_i\in\{1,\dots,1000\}$ under a cosine schedule. Clean patches are encoded with causal attention, and corrupted-patch embeddings query a single-layer cross-attention decoder that attends only to the preceding clean context and predicts the clean patch rather than the noise. 

\subsection{Invariance-Based SSL}
\paragraph{DINO (Eq.~\ref{eq:dino}).}
We define the remaining details for the implementation of DINO. This includes the Teacher-Student dynamics,  augmentation strategies, and unique hyper-parameters.
\paragraph{View Defections.}
We define $\mathcal{G}$ to index the global views processed by the teacher,
and let $\mathcal{V}$ index all views processed by the student.
For teacher view $u\in\mathcal{G}$ and student view
$v\in\mathcal{V}$. The set that defines them is:
\begin{equation}
\mathcal{P}
=
\{(u,v):u\in\mathcal{G},\,v\in\mathcal{V},\,v\neq u\}.
\end{equation}
meaning, all sets of non-equal views are proccessed.

\paragraph{Teacher--Student Dynamics.}
Teacher parameters and the output center are updated using
exponential moving averages. Centering and temperature-based
sharpening regulate the teacher targets during self-distillation.
The teacher and projection heads are used during pre-training;
downstream evaluation uses the learned encoder.

\paragraph{Augmentation Strategies.}
All reported DINO results use image DINO's multi-crop scheme, adapted to one
dimension. The teacher receives $|\mathcal{G}|=2$ global views spanning the
full window: a Gaussian smoothing along the time axis with
$\sigma\sim\mathcal{U}(0.5,2.0)$, and a contrast/brightness jitter with
contrast $\sim\mathcal{U}(0.6,1.4)$, brightness offset
$\sim\mathcal{U}(-0.2,0.2)$, and per-sample jitter $\sim\mathcal{U}(0,0.1)$,
the 1-D analogue of colour jitter. The student additionally receives $6$
local views, each a randomly positioned sub-window covering $40\%$ of the
sequence with additive Gaussian noise of standard deviation
$\sim\mathcal{U}(0.05,0.2)$. This gives $|\mathcal{V}|=8$ student views and
$|\mathcal{P}|=14$ teacher--student pairs. Every view applies its
transformation to all samples; only the magnitude parameter is resampled.

\paragraph{Hyper-parameters.}
The projection head $h$ follows \citet{dino}: a three-layer MLP with GELU
activations and hidden width $2048$, a weight-normalized final layer with
$K=1024$ output dimensions, no batch normalization, and the last layer's
weight norm frozen during training. Temperatures are $\tau_s=0.1$ for the
student and $\tau_t=0.04$ for the teacher, held constant (no teacher-temperature
warm-up). The teacher-output center is updated by an exponential moving average
with momentum $0.9$, and the teacher weights by an EMA with base momentum
$0.9995$, cosine-annealed to $1.0$ over training. The encoder uses a $1$-epoch
learning-rate warm-up.
Local views crop $\lfloor 0.4\times 336\rfloor = 134$ steps at a random offset, of which the encoder's non-overlapping patching retains $8$ whole patches ($128$ steps, $38\%$ of the window), discarding the $6$-step remainder.

\subsection{Contrastive Learning.}
\label{app:Con_def}
\paragraph{SoftCLT (Eq.~\ref{eq:con_loss})}
\label{subsec:appendix_softclt}
We define the instance-wise and temporal loss terms of SoftCLT, and describe the augmentations and the hierarchical setup used for SoftCLT.
We evaluate the temporal-soft variant of SoftCLT
\citep{softclt}, combining hard instance-wise contrastive learning
with soft temporal assignments.
\paragraph{Loss Definition.}
Given a batch of $N$ series, two augmented views produce patch
embeddings $\mathbf{z}_{i,t}$ and
$\tilde{\mathbf{z}}_{i,t}$.
Let $p_I((i,j),t)$ denote the softmax similarity between instance
views at patch position $t$, normalized over candidate instance
views other than the anchor. Let $p_T(i,(t,s))$ denote the
corresponding temporal similarity within a series, normalized
over candidate patch views other than the anchor.
The indices $\tilde{i}$ and $\tilde{t}$ denote the matching
instance and patch in the other view.

\paragraph{Instance-wise Loss.}
The instance branch uses the hard InfoNCE objective:
\begin{equation}
\ell_I^{(i,t)}
=
-\log p_I\big((i,\tilde{i}),t\big).
\end{equation}
Our configuration uses $\tau_I=0$ to select this hard branch.
No instance-distance matrix or additional soft instance
assignments are computed.

\paragraph{Temporal Loss.}
Additional temporal assignments decrease with the distance
between patch positions:
\begin{equation}
w_T(t,s)
=
2\,\sigma\!\left(-\tau_T|t-s|\right)
=
\frac{2}{1+\exp(\tau_T|t-s|)}.
\end{equation}
Distances refer to the underlying temporal positions in the
series. The temporal loss combines the matching-view positive
with the additional weighted pairs:
\begin{equation}
\ell_T^{(i,t)}
=
-\log p_T\big(i,(t,\tilde{t})\big)
-
\sum_{s\neq t,\tilde{t}}
w_T(t,s)\log p_T\big(i,(t,s)\big).
\end{equation}

\paragraph{Combined Objective and Hierarchy.}
The complete objective combines the instance and temporal losses:
\begin{equation}
\mathcal{L}_{\mathrm{SoftCLT}}
=
\lambda\mathcal{L}_I
+
(1-\lambda)\mathcal{L}_T,
\end{equation}
where $\mathcal{L}_I$ and $\mathcal{L}_T$ aggregate the
corresponding anchor losses across both views and the
hierarchical representations.

We use $\tau_T=2$ and $\lambda=0.5$.
The hierarchy applies TS2Vec-style max-pooling to patch-level
representations. The initial representation contains 21
non-overlapping patches of length 16 for a 336-step context.
At pooling depth $d$, the temporal temperature is
$\tau_T\,2^d$.

\paragraph{Augmentation Strategies.}
SoftCLT uses the random-cropping scheme of TS2Vec \citep{ts2vec} and no other
input transformation. Series longer than the context window are first reduced
to a random $336$-step window ($P=21$ patches). For each batch we sample an
overlap length $c\sim\mathcal{U}\{2,\dots,P\}$ in patch units, then two
subsequences that share exactly those $c$ patches: offsets
$\ell\sim\mathcal{U}\{0,\dots,P-c\}$, $r=\ell+c$,
$e_\ell\sim\mathcal{U}\{0,\dots,\ell\}$ and
$e_r\sim\mathcal{U}\{r,\dots,P\}$ give the views $[e_\ell,r)$ and
$[\ell,e_r)$, each additionally shifted by a per-sample offset
$\delta\sim\mathcal{U}\{-e_\ell,\dots,P-e_r\}$. The contrastive loss is
computed only on the $c$ patch representations common to both views, so the
two views differ in context while being aligned on the shared region. All crop
boundaries are multiples of the patch length, so both views align with the
encoder's patch grid. Because the backbone is the patched transformer rather
than TS2Vec's dilated convolution, the binomial timestamp masking of the
original implementation is not applied.

\subsection{Joint-Embedding Predictive Architectures}
\paragraph{JEPA (Eq.~\ref{eq:jepa}).}
\label{app:Jepa_def}
We provide the details on the masking and predictor, collapse prevention (VicReg \cite{bardes2022vicreg}), and unique hyperparameter details.
\paragraph{Masking and predictor.}
Targets $\mathcal{M}$ are sampled by multi-block masking (2 non-overlapping blocks, each covering 25\% of the patches, so $10$ of $21$ patches are targets), and the context $\mathcal{C}$ is the complement of $\mathcal{M}$: the generator returns a disjoint index partition, so every patch is either context or target. The student encodes only the context patches, $\mathbf{Z}_{\mathcal{C}}=f_\theta(\mathbf{x}_{\mathcal{C}})$, and a $3$-layer Transformer predictor (width $64$, $4$ heads) receives $\mathbf{Z}_{\mathcal{C}}$ together with positional queries for the target indices, $\widehat{\mathbf{Z}}_{\mathcal{M}}=\operatorname{pred}_{\psi}(\mathbf{Z}_{\mathcal{C}},\mathcal{M})$. The teacher $\theta'$ is an exponential moving average of $\theta$ with momentum $0.996$ and encodes the full, unmasked window.

\paragraph{Collapse prevention.}
The variance and covariance penalties of Eq.~\ref{eq:jepa} follow
VICReg~\citep{bardes2022vicreg} but are computed on the context
representations only. Writing $N=B\cdot C$ for the number of
(sample, channel) pairs in a batch and
$\mathbf{Z}_{\mathcal{C}}\in\mathbb{R}^{N\times|\mathcal{C}|\times D}$ for the
context embeddings,
\begin{align}
\mathcal{L}_{\mathrm{var}}
&=
\frac{1}{ND}\sum_{n=1}^{N}\sum_{d=1}^{D}
\max\!\left(0,\;\gamma-
\sqrt{\operatorname{Var}_{p\in\mathcal{C}}\!\left(z_{n,p,d}\right)+\epsilon}
\right),
\\[2pt]
\mathcal{L}_{\mathrm{cov}}
&=
\frac{1}{D}\sum_{d\neq d'}\mathbf{C}_{d,d'}^{2},
\qquad
\mathbf{C}=\frac{1}{M-1}\,\tilde{\mathbf{Z}}^{\top}\tilde{\mathbf{Z}},
\end{align}
where $\tilde{\mathbf{Z}}\in\mathbb{R}^{M\times D}$ collects all
$M=N|\mathcal{C}|$ context embeddings, centered columnwise. The variance hinge
is taken \emph{within} a sample across its context positions, penalizing
positional collapse, while the covariance penalty acts on the pooled
population and decorrelates feature dimensions. Neither term constrains the
teacher's target embeddings, which are stop-gradiented, so the prediction loss
is the only term evaluated at target positions.
The terms use $\gamma=1$ and $\epsilon=10^{-4}$, and the objective is
weighted as $\lambda_{\mathrm{pred}}=1$, $\lambda_{\mathrm{var}}=1$,
$\lambda_{\mathrm{cov}}=0.04$.

\paragraph{Hyper-parameters.}
Beyond the shared 8-layer backbone, JEPA introduces four groups of settings.
\emph{Predictor:} $3$ Transformer layers of width $64$ with $4$ heads --- a
third the depth and half the width of the encoder, so that representation
quality is not offloaded onto the predictor. It is optimized as a separate
parameter group with learning rate $5\times10^{-3}$ and weight decay
$10^{-4}$, under a
one-cycle schedule with $5\%$ warm-up and gradient clipping at $1.0$.
\emph{Teacher:} an exponential moving average of the student with momentum
$0.996$, encoding the full unmasked window.

\paragraph{LeJEPA (Eq.~\ref{eq:LeJEPA})}
We provide the details of the views and objectives, anti-collapse mechanism and implementation (SIGreg~\cite{balestriero2511lejepa}), and unique hyper-parameters.
\label{subsec:appendix_lejepa}

\textbf{Views and objective.}
LeJEPA uses DWT-based views described in Section~\ref{main:leJEPA} (Ablations, studies and effects in App.~\ref{subsec:appendix_augmentations}). In
Eq.~\ref{eq:LeJEPA}, $\mathbf{z}^{(1)}=f_\theta(\mathbf{x}^{(1)})$ is the encoding of the reference view
$\mathbf{x}^{(1)}=\mathcal{A}_1(\mathbf{x})$, which soft-thresholds the wavelet
detail coefficients at $30\%$ of each band's largest coefficient, and
$\mathbf{z}^{(2)}=f_\theta(\mathbf{x}^{(2)})$ that of the perturbed view
$\mathbf{x}^{(2)}=\mathcal{A}_2(\mathbf{x})$, which adds Gaussian noise of
scale $s\sim\mathcal{U}(0.1,0.3)$ to every detail coefficient; both views leave
the approximation band unchanged. The first term of Eq.~\ref{eq:LeJEPA} is the squared error between the two encodings averaged over the $L$ patch positions, and the statistic $T(\cdot)$ is applied to each view independently and averaged. We use $\lambda=0.008$. Both branches receive gradients through the shared
encoder $f_\theta$; no stop-gradient is used.

\paragraph{SIGReg Implementation.}
Sketched Isotropic Gaussian Regularization (SIGReg) uses an Epps--Pulley-based
penalty to encourage Gaussianity along random projections. The regularizer
treats every patch embedding as an individual sample: a batch of $B$ windows
with $C$ channels and $L$ patches yields $N=BCL$ vectors
$\mathbf{v}_n\in\mathbb{R}^{D}$. At each training step we sample $M=1024$ unit
projection directions in $\mathbb{R}^{D}$ by normalizing the columns of a
Gaussian matrix, seeded by the global step so that all workers draw the same
directions.

For a direction $\mathbf{a}$ and the $N$ embeddings, the empirical
characteristic function is
\[
\hat{\phi}_{\mathbf{a}}(t)
=
\frac{1}{N}
\sum_{n=1}^{N}
\exp\!\left(
\mathrm{i}t\langle\mathbf{a},\mathbf{v}_n\rangle
\right).
\]
We compare this function with the standard normal characteristic function
$\exp(-t^2/2)$. The squared residual
$\lvert\hat{\phi}_{\mathbf{a}}(t)-e^{-t^2/2}\rvert^{2}$ is evaluated at $17$
grid points in $[-5,5]$, weighted by $\exp(-t^2/2)$, integrated over $t$ by the
trapezoidal rule, scaled by $N$, and averaged across the $M$ directions.
For numerical stability we compute SIGReg in \texttt{float32}, since mixed-precision evaluation of
complex exponentials caused numerical instability in our implementation.

\paragraph{Hyperparameters.}
LeJEPA shares the 8-layer backbone with the other objectives and, unlike JEPA,
has no predictor and no EMA teacher: a single encoder processes both views and
both branches are trained directly. Its distinctive settings are therefore
confined to the views and the regularizer. \emph{Views:} a \texttt{db4} wavelet
decomposition over $J=3$ levels, with the reference view soft-thresholding each
detail band at $\sigma=0.3$ of that band's largest coefficient and the
perturbed view adding Gaussian noise of scale $s\sim\mathcal{U}(0.1,0.3)$ to
every detail coefficient. \emph{Regularizer:} SIGReg with $M=1024$ random
projections, evaluated at $17$ quadrature points on $[-5,5]$ in
\texttt{float32}. The default regularization weight is $\lambda=0.008$; its
downstream sensitivity is evaluated in
Table~\ref{tab:lejepa_lambda_sweep}. The encoder is optimized with AdamW at
learning rate $5\times10^{-4}$ and weight decay $10^{-2}$, with a $5\%$
warm-up and gradient clipping at $1.0$.

\section{Evaluation Baselines And Metrics}
\label{subsec:appendix_benchmarks_and_metrics}
This section details how SSL transferability is evaluated: the adaptation protocols, the baseline controls, and the metrics. For all tasks, we follow the dataset splits and evaluation protocols of TSLib~\citep{tslib}.
\subsection{Protocols and Baselines}
\label{subsec:protocal_baselines}
\paragraph{Protocols.}
We assess the learned representations under three adaptation protocols, which differ in the capacity of the task head and in whether the encoder is updated:
\begin{itemize}
    \item \textbf{Linear probing.} The pre-trained backbone is frozen and a single linear head is trained for each downstream task.
    \item \textbf{MLP probing.} The pre-trained backbone is frozen and a shallow MLP head (one hidden layer of width 512, GELU activation, dropout 0.2) is trained for each downstream task.
    \item \textbf{Full fine-tuning.} The pre-trained backbone and the task head are trained jointly on the downstream task.
\end{itemize}
Each protocol isolates a different aspect of transfer. Linear probing measures whether task-relevant information is linearly accessible in the frozen latent space, so that performance reflects the structure of the representation rather than the capacity of a non-linear head to compensate for it. MLP probing measures whether that information is present but only non-linearly accessible. Full fine-tuning measures the quality of the pre-trained weights as an initialization for task-specific optimization.

\paragraph{Baselines.}
To quantify the benefit of self-supervised pre-training, every pre-trained model is compared against a non-pre-trained control with an identical Transformer backbone, parameter count, and downstream training budget. Under linear and MLP probing, the control is a randomly initialized, frozen encoder; under full fine-tuning, it is the same architecture trained from scratch on the downstream task. We regard pre-training as beneficial only when it outperforms the corresponding control, which is the condition $\Delta>0$ in Eq.~\ref{eq:transfer_gain}. These controls are strong rather than nominal: simple or untrained models have been shown to match more elaborate deep architectures on downstream time-series tasks~\citep{zeng2023transformers,kim2022rigorous}.

\subsection{Metrics}
\label{subsec:metrics}
We define the metrics reported for each task family. Forecasting is measured by error (lower is better), whereas classification and anomaly detection are measured by accuracy and F1 (higher is better); the direction scalar $s_{\mathcal{T}}$ in Eq.~\ref{eq:transfer_gain} accounts for this difference when computing the transfer gain $\Delta$. All metrics are computed on the test split, and unless stated otherwise, reported values are means over five seeds.
\subsubsection{Forecasting}
\label{subsec:for_def}
\paragraph{Metric Definitions And Implementation.}
For multivariate sequence-to-sequence prediction, where a fixed context window of length 336 is used to forecast the next $H$ values of all $C$ variates, with $H\in\{96,192,336,720\}$. For a test set of $N$ windows with targets $\mathbf{y}^{(n)}\in\mathbb{R}^{H\times C}$ and predictions $\hat{\mathbf{y}}^{(n)}$, we report the Mean Squared Error and the Mean Absolute Error,
\begin{equation}
\mathrm{MSE}=\frac{1}{NHC}\sum_{n=1}^{N}\big\|\hat{\mathbf{y}}^{(n)}-\mathbf{y}^{(n)}\big\|_2^2,
\qquad
\mathrm{MAE}=\frac{1}{NHC}\sum_{n=1}^{N}\big\|\hat{\mathbf{y}}^{(n)}-\mathbf{y}^{(n)}\big\|_1,
\end{equation}
both computed on series standardized with training-set statistics. MSE is the primary metric and MAE the secondary one. Unless stated otherwise, reported values are averaged over the four horizons.

\subsubsection{Classification}
\label{subsec:class_def}
\paragraph{Metric Definitions And Implementation.}
For sequence-level classification, where each multivariate series is assigned one of $K$ class labels. For a test set of $N$ series with labels $y^{(n)}$ and predictions $\hat{y}^{(n)}=\arg\max_{k}\hat{p}^{(n)}_{k}$, we report top-1 accuracy,
\begin{equation}
\mathrm{Acc}=\frac{100}{N}\sum_{n=1}^{N}\mathbb{1}\big[\hat{y}^{(n)}=y^{(n)}\big],
\end{equation}
expressed as a percentage. Since the number of classes differs across datasets, chance-level accuracy differs as well, and absolute accuracies are not comparable across datasets; relative gains over the matched control (Eq.~\ref{eq:transfer_gain}) are used for that purpose.

\subsubsection{Anomaly Detection}
\label{app:anom_metric}
\paragraph{Anomaly score.}
The anomaly score of timestamp $t$ is the squared reconstruction error of a head $g_\psi$ on the frozen encoder, $\hat{\mathbf{x}}=g_\psi(f_\theta(\mathbf{x}))$, averaged over the $C$ variates:
\begin{equation}
  s_t=\frac{1}{C}\sum_{c=1}^{C}\bigl(\hat{x}_{t,c}-x_{t,c}\bigr)^2,
  \label{eq:anomaly_score}
\end{equation}
The head is trained on the anomaly-free training split to minimize $\lVert\hat{\mathbf{x}}-\mathbf{x}\rVert_2^2$.
\paragraph{Setup.}
We consider anomaly detection in multivariate time series, where timestamp $t$ is flagged as anomalous, $\hat{y}_t=1$, when its anomaly score $s_t$ (Eq.~\ref{eq:anomaly_score}) exceeds a threshold $\tau$. Following the TSLib~\citep{tslib} protocol, $\tau$ is the $(100-r)$-th percentile of the pooled train-and-test score distribution, where $r$ is the benchmark's anomaly ratio ($0.5\%$ for SMD and $1.0\%$ for all other datasets); no validation-based tuning is performed. Let $y_t\in\{0,1\}$ denote the ground-truth label and $\mathcal{E}=\{E_1,\dots,E_{|\mathcal{E}|}\}$ the set of ground-truth anomaly events, i.e., the maximal contiguous runs of anomalous timestamps. Because point adjustment is known to inflate even weak detectors~\citep{kim2022rigorous,wu2023flawed}, we report three complementary scores. All three are computed from the same binary sequence $\{\hat{y}_t\}$ and take the form $\mathrm{F1}=2PR/(P+R)$; they differ only in how precision $P$ and recall $R$ are defined. Unless stated otherwise, reported values are percentages averaged over five seeds.
\paragraph{Metrics.} Here we define the three anomaly detection metrics reported:
\begin{itemize}
    \item \textbf{Point-adjusted F1 ($\mathrm{F1}_{\mathrm{Pa}}$)}: if any timestamp within an event $E_k$ is flagged, all timestamps in $E_k$ are counted as detected; precision, recall, and F1 are then computed point-wise over the adjusted predictions. This is the standard protocol of TSLib~\citep{tslib} and prior work, but is lenient toward sparse detections inside long events.
    \item \textbf{Event-wise F1 ($\mathrm{F1}_{\mathrm{Ev}}$)}: each ground-truth event counts as a single positive and each maximal run of predicted anomalies as a single detection. An event is recalled if at least one prediction falls within it; a predicted run is a true positive if it overlaps at least one event. Precision and recall are computed over events and detections rather than timestamps, so a long event contributes once regardless of its length.
    \item \textbf{Affiliation F1 ($\mathrm{F1}_{\mathrm{Af}}$)} \citep{huet2022affiliation}:  each timestamp is affiliated to its nearest ground-truth event, partitioning the time axis into affiliation zones. Within each zone, precision is the average distance from predicted anomalies to the event and recall is the average distance from event timestamps to the nearest prediction, both converted to probabilities against a uniformly random predictor over that zone and averaged across events. Unlike point adjusted and event-wise metrics, this metric is distance-based, parameter-free, and penalizes both early/late detections and false alarms far from any event.

\end{itemize}

\section{Experimental Setup}
\label{sec:appendix_exp_setup}
This section details the experimental setup: the pre-training corpora, the evaluation datasets, the implementation and hyperparameter details.

\subsection{Pre-training Datasets}
\label{subsec:appendix_datasets}
To isolate the effect of pre-training data source and scale on downstream tasks, we pre-train every objective on the five datasets. Dataset sizes are counted in training windows of length ($336$).
\begin{itemize}
    \item \textit{Monash (680K):} the real-world Monash corpus. It is the reference corpus for all main-text results and measures transfer from observed, multi-domain data.
    \item \textit{Synthetic (680K):} a purely synthetic corpus matched to Monash in size. Comparing it with Monash isolates the effect of data source at fixed scale.
    \item \textit{Synthetic (2.5M):} a larger draw from the same synthetic generator. Comparing it with Synthetic (680K) isolates the effect of scale at fixed source.
    \item \textit{Hybrid (2.5M):} the full Monash corpus supplemented with $\sim$1.8M synthetic windows to match the size of Synthetic (2.5M).
    \item \textit{In-domain:} the training data of the target forecasting benchmark itself. It is evaluated for forecasting only.
\end{itemize}

\paragraph{Monash.}
We use the Monash Time Series Forecasting Repository~\citep{monash}, a collection spanning diverse domains including finance, weather, and energy. The pre-training set consists of 683,982 training windows, with 77,365 windows held out for validation. It represents a realistic yet relatively small-scale pre-training scenario.

\paragraph{In-domain.}
For in-domain pre-training, each forecasting benchmark's own training and validation splits serve as the pre-training dataset, so that the encoder is pre-trained and evaluated on the same dataset. The test split is never used for pre-training.

\paragraph{Synthetic.}
To study the effect of scale in the absence of domain bias, we generate synthetic data following TimePFN~\citep{timepfn}. We create two datasets, of 680,000 and 2,512,000 training windows, respectively. These series provide diverse temporal ``building blocks'' such as trends and periodicities, but may lack the stochastic variability of real-world data. The generation procedure is detailed below.

\paragraph{Hybrid.}
The hybrid corpus combines the full Monash training set with $\sim$1.8M synthetic windows, for a total of 2.5M. It supports two comparisons. Against Synthetic (2.5M), which has the same size, it isolates the effect of replacing part of the synthetic data with real data. Against Monash (680K), which it contains, it isolates the effect of adding synthetic volume to a fixed set of real data.

\subsection{Synthetic Data Generation}
\label{subsec:gen}
The synthetic corpus follows the structural-prior generation scheme introduced in TimePFN~\citep{timepfn}, which we instantiate using the original scripts with adjusted sequence length and dataset size. Each univariate time series is sampled as a single realization from a Gaussian Process (GP) prior on a uniform temporal grid of length $T$, where the covariance is constructed by randomly composing base kernels drawn (with replacement) from a fixed bank. The number of kernels per series is sampled uniformly ($K \sim \mathcal{U}\{1,\dots,5\}$), and kernels are combined via random addition or multiplication operators (Bernoulli$(\tfrac{1}{2})$), applied in a left-folded manner. 

The kernel bank includes periodic (ExpSineSquared) kernels spanning multiple temporal resolutions (e.g., daily, weekly, yearly), smooth kernels (RBF), multi-scale kernels (Rational Quadratic), linear kernels (Dot-Product), and noise components, enabling diverse temporal patterns such as trends, seasonality, and irregular fluctuations.

Multivariate series are generated using a Linear Coregionalization Model (LCM). A set of latent univariate GP factors is first sampled, and each channel is constructed as a simplex-weighted combination of these latent processes:
\[
x_c(t) = \sum_{j=1}^{J} w_{c,j} f_j(t), \quad w_c \sim \text{Dirichlet}(\alpha \mathbf{1})
\]
This induces structured cross-channel correlations with controllable coupling strength. The number of latent factors is sampled from a clipped Weibull distribution, while the Dirichlet concentration parameter controls the sparsity of channel assignments, interpolating between near-independent channels and dense entangled mixtures.

This construction yields synthetic data with controlled temporal structure and inter-channel dependencies, enabling large-scale pre-training while preserving key statistical properties such as multi-scale periodicity and smoothness. While we do not explicitly match higher-order statistics, the generated series exhibit diverse spectral and autocorrelation patterns consistent with real-world signals. However, the simplex-based mixing induces predominantly non-negative correlations, which differs from real benchmarks and represents a known limitation of the generator. We refer to~\citet{timepfn} for the full generation algorithm and parameterization.

We follow the default TimePFN generation setup with minor adjustments to sequence length and dataset size. Specifically, each dataset contains $N=4000$ series of length $T=2500$. For univariate generation, the number of kernels per series is sampled as $K \sim \mathcal{U}\{1,\dots,5\}$. The kernel bank consists of 33 atoms, including periodic (ExpSineSquared) kernels at multiple temporal resolutions, RBF and Rational Quadratic kernels with varying length scales, Dot-Product kernels, noise components, and a Constant kernel.

For multivariate generation, we use $C=160$ channels constructed via a Linear Coregionalization Model with $J$ latent factors sampled from a clipped Weibull distribution. Each latent factor is defined by its own kernel composition, where the number of kernels is sampled independently as $K \sim \mathcal{U}\{1,\dots,5\}$. Channel mixing weights are drawn from a Dirichlet distribution with concentration parameter $\alpha \sim \mathcal{U}[0.1,1.0]$, sampled per series. We use the original TimePFN implementation with these parameter settings, modifying only sequence length and dataset size.
\FloatBarrier
\subsection{Evaluation datasets}
\label{subsec:eval_data}
\paragraph{Forecasting.}
We evaluate on seven forecasting datasets ETTh1/2, ETTm1/2~\citep{informer}, Weather,
Electricity, and Traffic~\citep{autoformer}.
Details of size, split and channels are given in the Table~\ref{tab:forecasting_datasets}
\begin{table}[t]
\centering
\caption{Forecasting benchmarks. Splits are chronological and follow TSLib: 12/4/4 months for the ETT datasets and 70\%/10\%/20\% for the others. Split sizes are given in timesteps; the number of evaluation windows additionally depends on the context length and the horizon.}
\label{tab:forecasting_datasets}
\small
\setlength{\tabcolsep}{5pt}
\begin{tabular}{l r l r rrr}
\toprule
Dataset & Variates & Frequency & Timesteps & Train & Validation & Test \\
\midrule
ETTh1       &   7 & 1 hour  & 17{,}420 &  8{,}640 &  2{,}880 &  2{,}880 \\
ETTh2       &   7 & 1 hour  & 17{,}420 &  8{,}640 &  2{,}880 &  2{,}880 \\
ETTm1       &   7 & 15 min  & 69{,}680 & 34{,}560 & 11{,}520 & 11{,}520 \\
ETTm2       &   7 & 15 min  & 69{,}680 & 34{,}560 & 11{,}520 & 11{,}520 \\
Weather     &  21 & 10 min  & 52{,}696 & 36{,}887 &  5{,}270 & 10{,}539 \\
Electricity & 321 & 1 hour  & 26{,}304 & 18{,}412 &  2{,}632 &  5{,}260 \\
Traffic     & 862 & 1 hour  & 17{,}544 & 12{,}280 &  1{,}756 &  3{,}508 \\
\bottomrule
\end{tabular}
\end{table}
\paragraph{Classification.}
We evaluate classification on nine datasets from the UEA multivariate archive~\citep{uea}, spanning spectroscopy, neural recordings, speech, and motion sensing (Table~\ref{tab:classification_datasets}). The datasets differ widely in size, from 120 to 6{,}599 training series, and in the number of classes, from 2 to 26.
\begin{table}[t]
\centering
\caption{Classification benchmarks from the UEA multivariate archive~\citep{uea}, with the archive's predefined train/test splits as used in TSLib. Series length is the maximum length for the two variable-length datasets (JapaneseVowels and SpokenArabicDigits).}
\label{tab:classification_datasets}
\small
\setlength{\tabcolsep}{5pt}
\begin{tabular}{l l r r r r r}
\toprule
Dataset & Domain & Variates & Length & Classes & Train & Test \\
\midrule
EthanolConcentration & Spectroscopy   &   3 & 1{,}751 &  4 &   261 &   263 \\
FaceDetection        & MEG            & 144 &      62 &  2 & 5{,}890 & 3{,}524 \\
Handwriting          & Accelerometer  &   3 &     152 & 26 &   150 &   850 \\
Heartbeat            & Heart sound    &  61 &     405 &  2 &   204 &   205 \\
JapaneseVowels       & Speech         &  12 &      29 &  9 &   270 &   370 \\
SelfRegulationSCP1   & EEG            &   6 &     896 &  2 &   268 &   293 \\
SelfRegulationSCP2   & EEG            &   7 & 1{,}152 &  2 &   200 &   180 \\
SpokenArabicDigits   & Speech         &  13 &      93 & 10 & 6{,}599 & 2{,}199 \\
UWaveGestureLibrary  & Accelerometer  &   3 &     315 &  8 &   120 &   320 \\
\bottomrule
\end{tabular}
\end{table}
\paragraph{Anomaly Detection.}
We evaluate anomaly detection on five multivariate benchmarks (Table~\ref{tab:anomaly_datasets}): two server-monitoring datasets, SMD~\citep{su2019robust} and PSM~\citep{abdulaal2021practical}; two spacecraft-telemetry datasets, MSL and SMAP~\citep{hundman2018detecting}; and the SWaT water-treatment testbed~\citep{mathur2016swat}. The datasets differ substantially in size and in the fraction of anomalous test timestamps, from about 4\% on SMD to about 28\% on PSM.
\begin{table}[t]
\centering
\caption{Anomaly-detection benchmarks, with the predefined train/test splits as used in TSLib. Training data are anomaly-free and unlabeled; labels are available for the test split only. Sizes are given in timesteps, and the anomaly ratio is the fraction of anomalous timestamps in the test split.}
\label{tab:anomaly_datasets}
\small
\setlength{\tabcolsep}{5pt}
\begin{tabular}{l l r r r r}
\toprule
Dataset & Domain & Variates & Train & Test & Anomaly ratio (\%) \\
\midrule
SMD~\citep{su2019robust}             & Server monitoring   & 38 & 708{,}405 & 708{,}420 &  4.2 \\
PSM~\citep{abdulaal2021practical}    & Server monitoring   & 25 & 132{,}481 &  87{,}841 & 27.8 \\
MSL~\citep{hundman2018detecting}     & Spacecraft telemetry & 55 &  58{,}317 &  73{,}729 & 10.5 \\
SMAP~\citep{hundman2018detecting}    & Spacecraft telemetry & 25 & 135{,}183 & 427{,}617 & 12.8 \\
SWaT~\citep{mathur2016swat}          & Water treatment     & 51 & 495{,}000 & 449{,}919 & 12.1 \\
\bottomrule
\end{tabular}
\end{table}

\FloatBarrier
\subsection{Implementation And Training Details}\label{subsec:appendix_implement}
All methods share the Transformer encoder backbone of Section~\ref{subsec:ssl_paradigms}. The default configuration has 8 layers, embedding dimension $D=128$, 16 attention heads, feed-forward dimension 512, and patch length $P=16$. The input context is 336 timesteps ($L=21$ patches) for pre-training, forecasting, and anomaly detection, and 1{,}152 timesteps ($L=72$ patches) for classification. Method-specific components are added only where the SSL objective requires them: a linear reconstruction decoder for MAE, a residual MLP prediction head for NTP, a cross-attention denoising decoder for Diffusion, a projection head and an EMA teacher for DINO, and a latent predictor and an EMA teacher for JEPA; LeJEPA and SoftCLT use the encoder alone. All such components are discarded after pre-training, so downstream evaluation uses the encoder only.

\paragraph{Pre-Training Details}
\label{subsec:appendix_pretraining}
All objectives are pre-trained for 20 epochs on windows of 336 timesteps with patch length $P=16$. Each objective constructs its training signal as described in Appendix~\ref{subsec:fair_eval}: masking for MAE and JEPA, causal prediction for NTP, noise corruption for Diffusion, and augmented views for DINO, SoftCLT, and LeJEPA. Learning rate and batch size are tuned per objective under comparable tuning budgets, and all objective-specific hyperparameters are listed in Table~\ref{tab:model_hparams}; the matched-budget comparison of Appendix~\ref{subsec:fair_eval} instead fixes them across objectives. We saw our experiments, JEPA collapsed to a constant representation under both Adam and AdamW, whereas SGD trained stably. For every run, we keep the checkpoint with the lowest validation loss, and all pre-training runs are repeated over five seeds.
\begin{table}[t]
\centering
\caption{Objective-specific pre-training hyperparameters. The backbone is identical across objectives (8 layers, $d_{\mathrm{model}}=128$, 16 heads, $d_{ff}=512$, patch length 16, 336-step context; 1152 for the classification protocol) and all objectives pre-train for 20 epochs, so only the settings that differ are listed. Learning rates and batch sizes are each objective's own tuned values; under the equal-budget protocol both are standardized.}
\label{tab:model_hparams}
\small
\setlength{\tabcolsep}{4pt}
\begin{tabular}{l l r r l l}
\toprule
Objective & Optimizer & LR & Batch & Wd & Specific settings \\
\midrule
NTP       & Adam       & $1\times10^{-4}$ &  64 & --                       & causal, 21+6 patches \\
MAE       & Adam       & $1\times10^{-4}$ &  64 & --                       & mask 0.4, linear decoder \\
Diffusion & Adam       & $1\times10^{-4}$ & 128 & --                       & $T{=}1000$, cosine \\
DINO      & AdamW      & $5\times10^{-4}$ & 128 & $0.04\!\rightarrow\!0.1$ & $K{=}1024$, EMA $0.9995$ \\
JEPA      & SGD        & $3\times10^{-3}$ & 128 & $3\times10^{-3}$         & pred.\ LR $5\times10^{-3}$, EMA $0.996$ \\
LeJEPA    & AdamW      & $5\times10^{-4}$ & 128 & $1\times10^{-2}$         & $\lambda{=}0.008$, 1024 proj. \\
SoftCLT   & AdamW      & $2\times10^{-4}$ & 128 & $1\times10^{-2}$         & $\tau_{\mathrm{temp}}{=}2.0$, $\lambda{=}0.5$ \\
\bottomrule
\end{tabular}
\end{table}

\paragraph{Downstream Training Details}
\label{subsec:appendix_downstream}
For classification and anomaly detection, downstream hyperparameters are shared across all SSL objectives and the non-pre-trained controls, so that differences in the result tables reflect the encoder weights rather than the adaptation budget. Classification uses a learning rate of $1\mathrm{e}{-3}$ with a batch size of 64 for 20 epochs. Anomaly detection uses Adam with a learning rate of $1\mathrm{e}{-3}$ and cosine annealing for 10 epochs, with early stopping at a patience of 3. Forecasting uses dataset-specific batch sizes (16--256) and learning rates in $[1\mathrm{e}{-4},4\mathrm{e}{-4}]$, and is trained for 20 epochs.

\subsection{Compute and Train-time Details}\label{subset:appendix_compute}
All experiments were conducted on NVIDIA RTX PRO 6000 Blackwell GPUs (96GB VRAM, 600W TDP), with each backbone trained on a single GPU. Peak memory usage during pre-training ranged between approximately 8–32GB depending on model depth and dataset. On the Monash collection, pre-training runtime scaled with backbone depth: 2–4 layer models required approximately 2–3 hours, 8-layer models 4–5 hours, 12-layer models 6–7 hours, and 24-layer models around 10 hours per run. Synthetic and mixed-data variants approximately doubled these runtimes, with the deepest 24-layer configurations requiring up to 28 hours.

Downstream evaluation is substantially lighter than pre-training. Classification completes within approximately 20 minutes per dataset, and anomaly detection within 15 minutes per dataset, depending on model depth. Forecasting on smaller benchmarks (ETTm1/2, ETTh1/2, Weather) typically completes within 15 minutes per dataset, while larger datasets scale more significantly: Electricity requires up to 4 hours and Traffic up to 12 hours for the deepest configurations. Memory usage during downstream tasks remains below 20GB for most datasets, increasing to approximately 40GB for Electricity, while certain Traffic configurations may exceed available VRAM (out-of-memory).

\section{Extended Results}
\label{app:results}
This section reports the complete results behind the main-text findings, together with a brief analysis of each table. It is organized around our three research questions, and for each one we present the extended results that support the answers given in the main text.
\subsection{Main Results under Monash Pre-Training}
\label{subsec:monash_main}
This subsection gives the full scores underlying Table~\ref{tab:control_mon} and Figures~\ref{fig:rq1_dividend} and~\ref{fig:rq2_evaluation}. For each protocol, we report forecasting, classification, and anomaly detection in turn, and compare every pre-trained encoder against the other methods and baseline.

\subsubsection{Linear probing}
We report results for backbones pre-trained on \textit{Monash} and evaluated with a linear head on the frozen encoder.
\paragraph{Forecasting.}
Table~\ref{tab:forecasting_l8_monash_clean} shows that the transfer gain for forecasting is small. Under Monash, the random frozen encoder attains lower mean MSE in 40 of 49 objective--dataset comparisons and the lowest MSE on \textit{ETTh1}, \textit{Electricity}, and \textit{Traffic}; the largest improvement is $\Delta=+0.016$ MSE (Diffusion on \textit{ETTh2}). LeJEPA degrades on every benchmark ($\Delta=-0.413$ on \textit{Traffic}), and SoftCLT trails the baseline on every dataset except \textit{Weather}. Errors grow with the horizon for every objective, with a single exception (Diffusion on \textit{ETTh2}), and no objective closes the gap at longer horizons (Tables~\ref{tab:appendix_forecasting_horizons_combined} and~\ref{tab:appendix_forecasting_horizons_additional}).
\begin{table}[ht]
\centering
\setlength{\tabcolsep}{3pt}
\caption{Linear-probing forecasting error (MSE and MAE, mean~$\pm$~std over five seeds, averaged over horizons $\{96,192,336,720\}$) using an 8-layer backbone pre-trained on the \textbf{Monash} dataset. Rows are grouped into signal-space predictive and reconstructive, latent-space alignment, and contrastive objectives; lower is better and the best result per dataset and metric is in \textbf{bold}.}
\label{tab:forecasting_l8_monash_clean}
\resizebox{\textwidth}{!}{%
\begin{tabular}{l*{14}{c}}
\toprule
\textbf{Objective} & \multicolumn{2}{c}{\textbf{ETTh1}} & \multicolumn{2}{c}{\textbf{ETTh2}} & \multicolumn{2}{c}{\textbf{ETTm1}} & \multicolumn{2}{c}{\textbf{ETTm2}} & \multicolumn{2}{c}{\textbf{Weather}} & \multicolumn{2}{c}{\textbf{Electricity}} & \multicolumn{2}{c}{\textbf{Traffic}} \\
\cmidrule(lr){2-3}\cmidrule(lr){4-5}\cmidrule(lr){6-7}\cmidrule(lr){8-9}\cmidrule(lr){10-11}\cmidrule(lr){12-13}\cmidrule(lr){14-15}
 & MSE & MAE & MSE & MAE & MSE & MAE & MSE & MAE & MSE & MAE & MSE & MAE & MSE & MAE \\
\midrule
\textit{Random frozen encoder} & \textbf{0.414$\pm$0.002} & \textbf{0.426$\pm$0.002} & 0.350$\pm$0.002 & 0.393$\pm$0.001 & 0.356$\pm$0.001 & \textbf{0.379$\pm$0.001} & 0.261$\pm$0.001 & 0.318$\pm$0.000 & 0.248$\pm$0.000 & 0.281$\pm$0.001 & \textbf{0.172$\pm$0.000} & \textbf{0.267$\pm$0.000} & \textbf{0.435$\pm$0.000} & \textbf{0.298$\pm$0.000} \\
\midrule
\multicolumn{15}{l}{\textit{Signal-space predictive and reconstructive}} \\
NTP & 0.435$\pm$0.002 & 0.441$\pm$0.001 & 0.362$\pm$0.003 & 0.400$\pm$0.002 & 0.352$\pm$0.001 & 0.381$\pm$0.000 & 0.262$\pm$0.001 & 0.319$\pm$0.001 & 0.235$\pm$0.000 & 0.270$\pm$0.001 & 0.180$\pm$0.001 & 0.277$\pm$0.001 & 0.445$\pm$0.001 & 0.311$\pm$0.001 \\
MAE & 0.438$\pm$0.002 & 0.438$\pm$0.002 & 0.368$\pm$0.004 & 0.402$\pm$0.003 & 0.364$\pm$0.002 & 0.390$\pm$0.001 & 0.264$\pm$0.004 & 0.324$\pm$0.003 & 0.244$\pm$0.001 & 0.279$\pm$0.001 & 0.194$\pm$0.001 & 0.295$\pm$0.001 & 0.481$\pm$0.003 & 0.345$\pm$0.003 \\
Diffusion & 0.433$\pm$0.001 & 0.446$\pm$0.001 & \textbf{0.334$\pm$0.001} & \textbf{0.385$\pm$0.001} & 0.375$\pm$0.002 & 0.395$\pm$0.001 & 0.264$\pm$0.001 & 0.321$\pm$0.001 & 0.242$\pm$0.001 & 0.279$\pm$0.002 & 0.227$\pm$0.004 & 0.343$\pm$0.004 & 0.527$\pm$0.005 & 0.384$\pm$0.004 \\
\midrule
\multicolumn{15}{l}{\textit{Latent-space alignment}} \\
DINO & 0.444$\pm$0.008 & 0.449$\pm$0.006 & 0.368$\pm$0.005 & 0.404$\pm$0.004 & 0.360$\pm$0.001 & 0.387$\pm$0.000 & \textbf{0.254$\pm$0.002} & \textbf{0.311$\pm$0.001} & 0.236$\pm$0.001 & 0.271$\pm$0.001 & 0.182$\pm$0.005 & 0.281$\pm$0.006 & 0.449$\pm$0.013 & 0.319$\pm$0.012 \\
JEPA & 0.433$\pm$0.003 & 0.438$\pm$0.002 & 0.363$\pm$0.005 & 0.402$\pm$0.004 & 0.364$\pm$0.001 & 0.384$\pm$0.002 & 0.270$\pm$0.003 & 0.325$\pm$0.002 & 0.245$\pm$0.002 & 0.278$\pm$0.002 & 0.178$\pm$0.000 & 0.275$\pm$0.001 & 0.442$\pm$0.000 & 0.311$\pm$0.001 \\
LeJEPA & 0.616$\pm$0.009 & 0.540$\pm$0.004 & 0.394$\pm$0.003 & 0.425$\pm$0.003 & 0.444$\pm$0.005 & 0.440$\pm$0.003 & 0.288$\pm$0.004 & 0.345$\pm$0.003 & 0.262$\pm$0.001 & 0.297$\pm$0.001 & 0.346$\pm$0.007 & 0.432$\pm$0.005 & 0.848$\pm$0.013 & 0.550$\pm$0.005 \\
\midrule
\multicolumn{15}{l}{\textit{Contrastive}} \\
SoftCLT & 0.456$\pm$0.003 & 0.460$\pm$0.002 & 0.378$\pm$0.002 & 0.413$\pm$0.001 & 0.380$\pm$0.001 & 0.404$\pm$0.001 & 0.266$\pm$0.002 & 0.327$\pm$0.002 & 0.243$\pm$0.001 & 0.278$\pm$0.001 & 0.217$\pm$0.003 & 0.319$\pm$0.003 & 0.488$\pm$0.005 & 0.357$\pm$0.004 \\
\bottomrule
\end{tabular}%
}
\end{table}

\paragraph{Forecasting Horizon-wise analysis.}
Tables~\ref{tab:appendix_forecasting_horizons_combined} and~\ref{tab:appendix_forecasting_horizons_additional} report the forecasting MSE at each prediction horizon. Error increases monotonically with the horizon for every objective and dataset, with a single exception (Diffusion on \textit{ETTh2}). The leading model is stable across horizons but varies across datasets: the random frozen encoder is lowest at every horizon on \textit{ETTh1}, \textit{Electricity}, and \textit{Traffic} (on \textit{Electricity}, NTP and JEPA stay within 0.006--0.007 of it), NTP on \textit{ETTm1}, DINO on \textit{ETTm2}, NTP (tied with DINO at two horizons) on \textit{Weather}, and Diffusion on \textit{ETTh2} beyond the 96-step horizon. Consistent with our main findings, the gains from SSL do not grow with the horizon: NTP's margin over the random encoder shrinks from 0.016 to 0.008 on \textit{Weather} and from 0.006 to 0.001 on \textit{ETTm1} between the shortest and the longest horizon, so pre-training does not help more where forecasting is harder.
\begin{table}[t]
\centering
\caption{Linear probing MSE across prediction horizons using an 8-layer backbone pre-trained on the Monash dataset. Lower is better; best per dataset and horizon in \textbf{bold}.}
\label{tab:appendix_forecasting_horizons_combined}
\scriptsize
\setlength{\tabcolsep}{1.5pt}
\resizebox{\textwidth}{!}{%
\begin{tabular}{lcccccccccccc}
\toprule
\textbf{SSL Paradigm} 
& \multicolumn{4}{c}{\textbf{ETTh1}} 
& \multicolumn{4}{c}{\textbf{ETTh2}} 
& \multicolumn{4}{c}{\textbf{ETTm1}} \\
& 96 & 192 & 336 & 720 
& 96 & 192 & 336 & 720
& 96 & 192 & 336 & 720 \\
\midrule
\textit{Random frozen encoder}
& \textbf{0.373$\pm$.004} & \textbf{0.411$\pm$.004} & \textbf{0.429$\pm$.006} & \textbf{0.444$\pm$.008}
& \textbf{0.281$\pm$.002} & 0.345$\pm$.004 & 0.375$\pm$.006 & 0.401$\pm$.003
& 0.296$\pm$.001 & 0.333$\pm$.001 & 0.370$\pm$.003 & 0.424$\pm$.001 \\
\midrule
\multicolumn{13}{l}{\textit{Signal-space predictive and reconstructive}} \\
NTP
& 0.391$\pm$.002 & 0.428$\pm$.001 & 0.448$\pm$.001 & 0.475$\pm$.008
& 0.296$\pm$.004 & 0.361$\pm$.005 & 0.376$\pm$.003 & 0.416$\pm$.004
& \textbf{0.290$\pm$.001} & \textbf{0.328$\pm$.001} & \textbf{0.365$\pm$.001} & \textbf{0.423$\pm$.002} \\
MAE
& 0.391$\pm$.002 & 0.426$\pm$.004 & 0.446$\pm$.003 & 0.488$\pm$.007
& 0.302$\pm$.006 & 0.370$\pm$.007 & 0.389$\pm$.008 & 0.412$\pm$.005
& 0.308$\pm$.002 & 0.343$\pm$.002 & 0.380$\pm$.004 & 0.427$\pm$.002 \\
Diffusion
& 0.397$\pm$.001 & 0.432$\pm$.001 & 0.439$\pm$.002 & 0.463$\pm$.004
& 0.286$\pm$.001 & \textbf{0.342$\pm$.001} & \textbf{0.322$\pm$.001} & \textbf{0.385$\pm$.002}
& 0.319$\pm$.003 & 0.356$\pm$.001 & 0.388$\pm$.003 & 0.438$\pm$.004 \\
\midrule
\multicolumn{13}{l}{\textit{Latent-space alignment}} \\
DINO
& 0.386$\pm$.001 & 0.423$\pm$.002 & 0.455$\pm$.019 & 0.511$\pm$.023
& 0.287$\pm$.004 & 0.367$\pm$.006 & 0.392$\pm$.009 & 0.427$\pm$.006
& 0.309$\pm$.001 & 0.337$\pm$.001 & 0.370$\pm$.001 & 0.425$\pm$.001 \\
JEPA
& 0.388$\pm$.002 & 0.422$\pm$.002 & 0.443$\pm$.001 & 0.480$\pm$.006
& 0.298$\pm$.003 & 0.360$\pm$.004 & 0.380$\pm$.007 & 0.412$\pm$.006
& 0.305$\pm$.002 & 0.342$\pm$.001 & 0.376$\pm$.001 & 0.431$\pm$.001 \\
LeJEPA
& 0.589$\pm$.011 & 0.612$\pm$.009 & 0.627$\pm$.008 & 0.638$\pm$.008
& 0.338$\pm$.005 & 0.391$\pm$.004 & 0.400$\pm$.003 & 0.448$\pm$.006
& 0.396$\pm$.005 & 0.423$\pm$.006 & 0.453$\pm$.005 & 0.504$\pm$.006 \\
\midrule
\multicolumn{13}{l}{\textit{Contrastive}} \\
SoftCLT
& 0.420$\pm$.002 & 0.449$\pm$.002 & 0.462$\pm$.002 & 0.494$\pm$.010
& 0.310$\pm$.003 & 0.371$\pm$.004 & 0.388$\pm$.003 & 0.442$\pm$.004
& 0.325$\pm$.001 & 0.359$\pm$.001 & 0.392$\pm$.002 & 0.444$\pm$.002 \\
\bottomrule
\end{tabular}%
}
\end{table}

\begin{table}[t]
\centering
\caption{Linear probing MSE across prediction horizons using an 8-layer backbone pre-trained on the Monash dataset. Lower is better; best per dataset and horizon in \textbf{bold}. MAE's Traffic column is averaged over three seeds; all other cells use five.}
\label{tab:appendix_forecasting_horizons_additional}
\scriptsize
\setlength{\tabcolsep}{1.5pt}
\resizebox{\textwidth}{!}{%
\begin{tabular}{lcccccccccccccccc}
\toprule
\textbf{SSL Paradigm} 
& \multicolumn{4}{c}{\textbf{ETTm2}} 
& \multicolumn{4}{c}{\textbf{Weather}} 
& \multicolumn{4}{c}{\textbf{Electricity}} 
& \multicolumn{4}{c}{\textbf{Traffic}} \\
& 96 & 192 & 336 & 720 
& 96 & 192 & 336 & 720
& 96 & 192 & 336 & 720
& 96 & 192 & 336 & 720 \\
\midrule
\textit{Random frozen encoder}
& 0.168$\pm$.001 & 0.226$\pm$.001 & 0.278$\pm$.001 & 0.372$\pm$.003
& 0.175$\pm$.001 & 0.218$\pm$.001 & 0.265$\pm$.000 & 0.333$\pm$.000
& \textbf{0.144$\pm$.000} & \textbf{0.158$\pm$.000} & \textbf{0.174$\pm$.000} & \textbf{0.213$\pm$.000} 
& \textbf{0.412$\pm$.000} & \textbf{0.425$\pm$.000} & \textbf{0.438$\pm$.000} & \textbf{0.465$\pm$.000} \\
\midrule
\multicolumn{17}{l}{\textit{Signal-space predictive and reconstructive}} \\
NTP      
& 0.168$\pm$.001 & 0.228$\pm$.002 & 0.281$\pm$.002 & 0.370$\pm$.002
& \textbf{0.159$\pm$.001} & \textbf{0.204$\pm$.001} & \textbf{0.253$\pm$.001} & \textbf{0.325$\pm$.000}
& 0.151$\pm$.001 & 0.166$\pm$.001 & 0.182$\pm$.001 & 0.220$\pm$.001 
& 0.420$\pm$.001 & 0.436$\pm$.001 & 0.449$\pm$.001 & 0.474$\pm$.001 \\
MAE
& 0.174$\pm$.004 & 0.230$\pm$.004 & 0.283$\pm$.006 & 0.370$\pm$.004
& 0.170$\pm$.002 & 0.213$\pm$.002 & 0.261$\pm$.001 & 0.331$\pm$.001
& 0.168$\pm$.001 & 0.180$\pm$.001 & 0.195$\pm$.001 & 0.232$\pm$.001 
& 0.457$\pm$.003 & 0.471$\pm$.004 & 0.482$\pm$.004 & 0.510$\pm$.002 \\
Diffusion
& 0.173$\pm$.001 & 0.229$\pm$.001 & 0.282$\pm$.003 & 0.370$\pm$.003
& 0.169$\pm$.001 & 0.212$\pm$.002 & 0.260$\pm$.002 & 0.326$\pm$.001
& 0.200$\pm$.003 & 0.210$\pm$.004 & 0.230$\pm$.004 & 0.269$\pm$.004 
& 0.505$\pm$.005 & 0.515$\pm$.005 & 0.530$\pm$.005 & 0.558$\pm$.005 \\
\midrule
\multicolumn{17}{l}{\textit{Latent-space alignment}} \\
DINO     
& \textbf{0.164$\pm$.001} & \textbf{0.218$\pm$.001} & \textbf{0.270$\pm$.002} & \textbf{0.362$\pm$.004}
& 0.161$\pm$.002 & \textbf{0.204$\pm$.002} & 0.254$\pm$.001 & \textbf{0.325$\pm$.001}
& 0.153$\pm$.005 & 0.168$\pm$.005 & 0.184$\pm$.005 & 0.223$\pm$.005 
& 0.427$\pm$.013 & 0.440$\pm$.014 & 0.452$\pm$.014 & 0.477$\pm$.013 \\
JEPA     
& 0.179$\pm$.003 & 0.234$\pm$.003 & 0.287$\pm$.003 & 0.378$\pm$.003
& 0.172$\pm$.003 & 0.215$\pm$.003 & 0.263$\pm$.002 & 0.331$\pm$.002
& 0.150$\pm$.000 & 0.164$\pm$.000 & 0.180$\pm$.001 & 0.220$\pm$.001
& 0.418$\pm$.000 & 0.431$\pm$.000 & 0.445$\pm$.000 & 0.475$\pm$.001 \\
LeJEPA  
& 0.203$\pm$.003 & 0.254$\pm$.004 & 0.302$\pm$.005 & 0.397$\pm$.009
& 0.196$\pm$.001 & 0.234$\pm$.001 & 0.277$\pm$.001 & 0.341$\pm$.001
& 0.319$\pm$.007 & 0.331$\pm$.007 & 0.348$\pm$.007 & 0.387$\pm$.008
& 0.815$\pm$.013 & 0.834$\pm$.015 & 0.854$\pm$.015 & 0.883$\pm$.015 \\
\midrule
\multicolumn{17}{l}{\textit{Contrastive}} \\
SoftCLT  
& 0.181$\pm$.002 & 0.233$\pm$.002 & 0.282$\pm$.002 & 0.370$\pm$.003
& 0.170$\pm$.001 & 0.212$\pm$.001 & 0.260$\pm$.001 & 0.330$\pm$.001
& 0.189$\pm$.003 & 0.203$\pm$.003 & 0.220$\pm$.003 & 0.259$\pm$.002
& 0.463$\pm$.005 & 0.477$\pm$.005 & 0.491$\pm$.005 & 0.520$\pm$.005 \\
\bottomrule
\end{tabular}%
}
\end{table}

\paragraph{Classification.}
Table~\ref{tab:classification_results_refined} reports the classification results. Three of the seven objectives outperform the random frozen encoder on a majority of datasets: NTP on 6 of 9, and Diffusion and LeJEPA on 5 of 9 each. The first two never attain the best accuracy on any dataset, although their gains can be large (NTP: $\Delta=+18.5$ on UWave and $+10.7$ on SpokenArabicDigits). LeJEPA, in contrast, attains the best accuracy on every dataset where it beats the baseline (\textit{FaceDetection}, \textit{Handwriting}, \textit{SCP1}, \textit{SpokenArabicDigits}, and \textit{UWave}), including the largest gain in the table ($\Delta=+35.1$ on \textit{UWave}, 71.69 vs.\ 36.56). These consistent gains for three of the seven objectives suggest that SSL pre-training is of value for classification. The remaining objectives are weaker: DINO and JEPA beat the baseline on only 3 of 9 datasets and SoftCLT on 4 of 9, with losses of up to $\Delta=-18.2$ (DINO on \textit{JapaneseVowels}) and $-14.4$ (SoftCLT on \textit{SCP1}). Overall, pre-training yields a benefit for classification: three of the seven objectives outperform the baseline on a majority of datasets, which suggests that gains are attainable with a suitable choice of objective.
\begin{table}[ht]
\centering
\setlength{\tabcolsep}{3pt}
\caption{Linear-probing classification accuracy (\%, mean~$\pm$~std over five seeds) using an 8-layer backbone pre-trained on the \textbf{Monash} dataset. Datasets: Ethanol (EthanolConcentration), FaceDet (FaceDetection), Handwrit (Handwriting), Heartbt (Heartbeat), JapVowel (JapaneseVowels), SCP1/SCP2 (SelfRegulationSCP1/2), Digits (SpokenArabicDigits), UWave (UWaveGestureLibrary). Rows are grouped into signal-space predictive and reconstructive, latent-space alignment, and contrastive objectives; the best result per dataset is in \textbf{bold}.}
\label{tab:classification_results_refined}
\resizebox{\textwidth}{!}{%
\begin{tabular}{l*{9}{c}}
\toprule
\textbf{Objective} & \textbf{Ethanol} & \textbf{FaceDet} & \textbf{Handwrit} & \textbf{Heartbt} & \textbf{JapVowel} & \textbf{SCP1} & \textbf{SCP2} & \textbf{Digits} & \textbf{UWave} \\
\midrule
\textit{Random frozen encoder} & \textbf{27.68$\pm$1.88} & 54.49$\pm$0.54 & 4.31$\pm$0.54 & 66.63$\pm$1.33 & \textbf{93.02$\pm$0.65} & 76.72$\pm$2.86 & 50.00$\pm$2.60 & 77.61$\pm$0.96 & 36.56$\pm$2.78 \\
\midrule
\multicolumn{10}{l}{\textit{Signal-space predictive and reconstructive}} \\
NTP & 24.94$\pm$0.19 & 57.76$\pm$0.60 & 6.26$\pm$2.08 & 66.83$\pm$3.16 & 89.62$\pm$0.99 & 74.75$\pm$1.22 & 52.89$\pm$3.57 & 88.29$\pm$0.40 & 55.06$\pm$1.83 \\
MAE & 24.56$\pm$0.46 & 54.20$\pm$0.91 & 5.06$\pm$1.30 & 62.05$\pm$1.39 & 86.27$\pm$2.59 & 75.70$\pm$1.97 & 51.66$\pm$2.65 & 81.51$\pm$2.38 & 34.56$\pm$2.93 \\
Diffusion & 24.41$\pm$1.12 & 55.72$\pm$0.49 & 6.73$\pm$1.15 & 64.59$\pm$3.37 & 90.59$\pm$0.40 & 68.33$\pm$4.37 & 52.33$\pm$0.65 & 84.56$\pm$1.44 & 50.12$\pm$3.56 \\
\midrule
\multicolumn{10}{l}{\textit{Latent-space alignment}} \\
DINO & 27.30$\pm$0.94 & 51.77$\pm$0.44 & 10.94$\pm$0.57 & 72.20$\pm$0.53 & 74.81$\pm$0.76 & 72.42$\pm$1.65 & 54.11$\pm$2.26 & 72.56$\pm$2.45 & 30.50$\pm$1.60 \\
JEPA & 24.79$\pm$0.45 & 52.28$\pm$0.32 & 5.93$\pm$1.94 & 62.93$\pm$1.69 & 92.38$\pm$0.69 & 65.26$\pm$2.83 & \textbf{54.89$\pm$2.73} & 79.37$\pm$1.34 & 32.63$\pm$2.77 \\
LeJEPA & 26.09$\pm$1.24 & \textbf{61.65$\pm$0.78} & \textbf{13.86$\pm$0.72} & 62.54$\pm$1.73 & 86.98$\pm$1.09 & \textbf{83.07$\pm$1.96} & 48.89$\pm$1.65 & \textbf{93.27$\pm$1.50} & \textbf{71.69$\pm$4.92} \\
\midrule
\multicolumn{10}{l}{\textit{Contrastive}} \\
SoftCLT & 27.23$\pm$1.09 & 51.90$\pm$0.58 & 7.93$\pm$0.90 & \textbf{72.97$\pm$0.24} & 75.89$\pm$1.58 & 62.32$\pm$2.11 & 53.11$\pm$1.03 & 74.93$\pm$3.96 & 48.63$\pm$6.92 \\
\bottomrule
\end{tabular}%
}
\end{table}

\paragraph{Anomaly detection.}
Table~\ref{tab:anomaly_detection_results} reports the anomaly-detection results. Pre-trained encoders outperform the random frozen encoder in 19 of 35 comparisons under $\mathrm{F1}_{\mathrm{PA}}$, 33 of 35 under $\mathrm{F1}_{\mathrm{Ev}}$, and 19 of 35 under $\mathrm{F1}_{\mathrm{Af}}$. Across the 15 dataset--metric cells, NTP, DINO, and MAE win most often (13, 12, and 11 cells, respectively), and even the weakest objective by this count, LeJEPA, wins on 8 of 15. The largest gains occur on the spacecraft-telemetry datasets: on \textit{MSL}, LeJEPA, JEPA, and Diffusion raise $\mathrm{F1}_{\mathrm{PA}}$ from 15.57 to about 57 ($\Delta=+41$), and on \textit{SMAP}, JEPA raises it from 7.98 to 34.83 ($\Delta=+26.9$); on \textit{PSM}, DINO gains $\Delta=+23.7$ under $\mathrm{F1}_{\mathrm{PA}}$ (71.23 vs.\ 47.55), and LeJEPA and DINO gain $\Delta=+25.5$ under $\mathrm{F1}_{\mathrm{Ev}}$. The largest losses are those of JEPA on \textit{PSM} ($\Delta=-21.7$ $\mathrm{F1}_{\mathrm{PA}}$, $-19.6$ $\mathrm{F1}_{\mathrm{Af}}$), of MAE on \textit{PSM} ($\Delta=-11.2$ $\mathrm{F1}_{\mathrm{PA}}$), and of JEPA and LeJEPA on \textit{SWaT}, where both fall to about 3 $\mathrm{F1}_{\mathrm{PA}}$ against 11.84 for the random encoder. Overall, pre-training yields a clear benefit for anomaly detection: every objective outperforms the random frozen encoder on a majority of dataset--metric cells, with \textit{SWaT} as the only dataset where pre-training does not help.
\begin{table}[ht]
\centering
\setlength{\tabcolsep}{3pt}
\caption{Linear-probing anomaly-detection F1 (\%, mean~$\pm$~std over five seeds) using an 8-layer backbone pre-trained on the \textbf{Monash} dataset. For each dataset we report point-adjusted (PA), event-wise (Ev), and affiliation-based (Af) F1. Rows are grouped into signal-space predictive and reconstructive, latent-space alignment, and contrastive objectives; the best result per dataset and metric is in \textbf{bold}.}
\label{tab:anomaly_detection_results}
\resizebox{\textwidth}{!}{%
\begin{tabular}{l*{15}{c}}
\toprule
\textbf{Objective}
& \multicolumn{3}{c}{\textbf{MSL}} & \multicolumn{3}{c}{\textbf{PSM}} & \multicolumn{3}{c}{\textbf{SMAP}} & \multicolumn{3}{c}{\textbf{SMD}} & \multicolumn{3}{c}{\textbf{SWaT}} \\
\cmidrule(lr){2-4}\cmidrule(lr){5-7}\cmidrule(lr){8-10}\cmidrule(lr){11-13}\cmidrule(lr){14-16}
& PA & Ev & Af & PA & Ev & Af & PA & Ev & Af & PA & Ev & Af & PA & Ev & Af \\
\midrule
\textit{Random frozen encoder} & 15.57$\pm$0.27 & 7.32$\pm$0.06 & 18.66$\pm$0.14 & 47.55$\pm$2.10 & 16.40$\pm$1.23 & 46.12$\pm$1.74 & 7.98$\pm$0.14 & 2.47$\pm$0.04 & 8.43$\pm$0.09 & 56.08$\pm$1.38 & 17.55$\pm$0.36 & 58.25$\pm$0.79 & \textbf{11.84$\pm$0.01} & 1.09$\pm$0.06 & \textbf{14.23$\pm$0.16} \\
\midrule
\multicolumn{16}{l}{\textit{Signal-space predictive and reconstructive}} \\
NTP & 28.58$\pm$0.34 & 13.11$\pm$0.09 & 29.45$\pm$0.19 & 55.02$\pm$9.27 & 29.64$\pm$4.71 & 49.55$\pm$2.94 & 22.68$\pm$0.61 & 6.19$\pm$0.10 & 18.39$\pm$0.42 & 61.75$\pm$0.34 & 20.00$\pm$0.43 & \textbf{64.29$\pm$0.37} & 11.50$\pm$0.01 & \textbf{2.86$\pm$0.36} & 9.54$\pm$0.16 \\
MAE & 29.36$\pm$0.16 & \textbf{13.29$\pm$0.07} & 29.96$\pm$0.20 & 36.31$\pm$5.19 & 19.13$\pm$3.30 & 41.34$\pm$2.72 & 27.05$\pm$0.85 & \textbf{7.17$\pm$0.11} & 21.45$\pm$0.64 & 61.09$\pm$1.27 & 19.05$\pm$0.76 & 63.34$\pm$1.25 & 7.72$\pm$4.80 & 1.28$\pm$0.53 & 10.87$\pm$2.39 \\
Diffusion & 56.53$\pm$0.29 & 12.26$\pm$0.07 & 48.34$\pm$0.26 & 43.78$\pm$1.52 & 34.64$\pm$4.27 & 41.58$\pm$1.73 & 25.23$\pm$3.65 & 6.01$\pm$0.49 & 20.06$\pm$2.50 & 52.03$\pm$0.26 & 30.24$\pm$0.35 & 55.57$\pm$0.12 & 8.59$\pm$2.15 & 1.39$\pm$0.69 & 9.39$\pm$1.99 \\
\midrule
\multicolumn{16}{l}{\textit{Latent-space alignment}} \\
DINO & 25.01$\pm$1.13 & 12.16$\pm$0.30 & 27.18$\pm$0.68 & \textbf{71.23$\pm$2.94} & 41.91$\pm$3.03 & \textbf{54.80$\pm$2.06} & 20.43$\pm$2.62 & 5.63$\pm$0.70 & 16.70$\pm$1.83 & \textbf{62.21$\pm$0.65} & 17.50$\pm$0.65 & 64.01$\pm$0.90 & 11.68$\pm$0.10 & 2.01$\pm$0.36 & 11.50$\pm$2.07 \\
JEPA & 56.92$\pm$0.06 & 12.16$\pm$0.02 & 49.15$\pm$0.06 & 25.83$\pm$5.30 & 30.13$\pm$3.41 & 26.48$\pm$3.22 & \textbf{34.83$\pm$0.41} & 6.58$\pm$0.09 & \textbf{27.31$\pm$0.58} & 52.56$\pm$0.16 & \textbf{31.15$\pm$0.12} & 55.99$\pm$0.09 & 3.14$\pm$0.09 & 1.71$\pm$0.19 & 4.10$\pm$0.07 \\
LeJEPA & \textbf{56.98$\pm$0.06} & 12.19$\pm$0.04 & \textbf{49.26$\pm$0.05} & 47.20$\pm$0.16 & \textbf{41.92$\pm$0.84} & 42.67$\pm$0.27 & 34.63$\pm$0.83 & 6.46$\pm$0.16 & 27.24$\pm$0.52 & 52.69$\pm$0.04 & 31.01$\pm$0.06 & 56.06$\pm$0.08 & 3.07$\pm$0.08 & 0.97$\pm$0.12 & 4.09$\pm$0.05 \\
\midrule
\multicolumn{16}{l}{\textit{Contrastive}} \\
SoftCLT & 30.30$\pm$0.05 & 13.18$\pm$0.02 & 30.77$\pm$0.05 & 44.77$\pm$2.79 & 38.39$\pm$1.06 & 39.82$\pm$2.12 & 29.54$\pm$0.34 & \textbf{7.17$\pm$0.07} & 23.42$\pm$0.27 & 52.02$\pm$0.20 & 30.43$\pm$0.16 & 55.84$\pm$0.07 & 11.32$\pm$0.35 & 2.10$\pm$1.09 & 12.97$\pm$0.71 \\
\bottomrule
\end{tabular}%
}
\end{table}

\subsubsection{MLP}
\label{app:results_mlp}
We report results for backbones pre-trained on \textit{Monash} and evaluated with an MLP head on the frozen encoder.

\paragraph{Forecasting.}
Table~\ref{tab:appendix_mlp_forecasting} reports the forecasting results. Diffusion is the only objective with consistent gains: it outperforms the random frozen encoder on all five datasets under both MSE and MAE, and it attains the lowest MSE on four of five datasets and the lowest MAE on all five (tied with DINO on \textit{ETTm2}). Its only loss in MSE is to DINO on \textit{ETTm2}, by 0.001 ($0.261\pm0.001$ vs.\ $0.260\pm0.002$), which is within seed variance. Since Diffusion beats the random encoder on only two of these five datasets under linear probing, this suggests that its representations encode forecasting-relevant information that is not linearly accessible and requires a non-linear head. The remaining objectives do not show consistent gains: four of seven improve on \textit{Weather}, and DINO leads on \textit{ETTm2}, but overall the pre-trained encoders outperform the control in only 11 of 35 comparisons, five of which are due to Diffusion.
\begin{table}[ht]
\centering
\caption{Forecasting performance (MSE and MAE) using an MLP probe, reported as mean~$\pm$~std. Datasets: \textbf{H1/H2} (ETTH1/2), \textbf{M1/M2} (ETTM1/2), \textbf{Wth} (Weather). Models are grouped by SSL paradigm; top results (lowest error) per metric and dataset are highlighted in \textbf{bold}.}
\label{tab:appendix_mlp_forecasting}
\footnotesize
\setlength{\tabcolsep}{6pt}
\begin{tabular}{lcccccc}
\toprule
\textbf{SSL Paradigm} & \textbf{Loss} & \textbf{H1} & \textbf{H2} & \textbf{M1} & \textbf{M2} & \textbf{Wth} \\
\midrule
\textit{Random frozen encoder} 
& MSE & 0.425$\pm$.002 & 0.362$\pm$.002 & 0.354$\pm$.003 & 0.264$\pm$.002 & 0.233$\pm$.001 \\
& MAE & 0.434$\pm$.001 & 0.399$\pm$.001 & 0.385$\pm$.002 & 0.321$\pm$.001 & 0.270$\pm$.000 \\
\midrule
\multicolumn{7}{l}{\textit{Signal-space predictive and reconstructive}} \\
NTP 
& MSE & 0.436$\pm$.004 & 0.369$\pm$.001 & 0.358$\pm$.002 & 0.264$\pm$.001 & 0.230$\pm$.000 \\
& MAE & 0.444$\pm$.002 & 0.402$\pm$.001 & 0.387$\pm$.001 & 0.322$\pm$.001 & 0.266$\pm$.000 \\
MAE 
& MSE & 0.444$\pm$.003 & 0.379$\pm$.001 & 0.352$\pm$.002 & 0.263$\pm$.026 & 0.232$\pm$.000 \\
& MAE & 0.444$\pm$.002 & 0.413$\pm$.001 & 0.384$\pm$.001 & 0.323$\pm$.018 & 0.266$\pm$.000 \\
Diffusion
& MSE & \textbf{0.420$\pm$.002} & \textbf{0.344$\pm$.001} & \textbf{0.351$\pm$.002} & 0.261$\pm$.001 & \textbf{0.228$\pm$.000} \\
& MAE & \textbf{0.432$\pm$.002} & \textbf{0.388$\pm$.000} & \textbf{0.382$\pm$.001} & 0.318$\pm$.001 & \textbf{0.264$\pm$.001} \\
\midrule
\multicolumn{7}{l}{\textit{Latent-space alignment}} \\
DINO 
& MSE & 0.447$\pm$.002 & 0.371$\pm$.003 & 0.361$\pm$.001 & \textbf{0.260$\pm$.002} & 0.229$\pm$.000 \\
& MAE & 0.455$\pm$.002 & 0.406$\pm$.002 & 0.390$\pm$.001 & \textbf{0.318$\pm$.001} & 0.265$\pm$.000 \\
JEPA 
& MSE & 0.436$\pm$.004 & 0.375$\pm$.005 & 0.362$\pm$.003 & 0.271$\pm$.002 & 0.233$\pm$.002 \\
& MAE & 0.442$\pm$.003 & 0.410$\pm$.002 & 0.390$\pm$.002 & 0.326$\pm$.001 & 0.268$\pm$.002 \\
LeJEPA 
& MSE & 0.583$\pm$.003 & 0.417$\pm$.003 & 0.395$\pm$.002 & 0.289$\pm$.002 & 0.246$\pm$.001 \\
& MAE & 0.529$\pm$.001 & 0.439$\pm$.002 & 0.414$\pm$.002 & 0.344$\pm$.001 & 0.283$\pm$.000 \\
\midrule
\multicolumn{7}{l}{\textit{Contrastive}} \\
SoftCLT
& MSE & 0.454$\pm$.008 & 0.374$\pm$.005 & 0.380$\pm$.005 & 0.282$\pm$.010 & 0.239$\pm$.002 \\
& MAE & 0.457$\pm$.007 & 0.411$\pm$.004 & 0.403$\pm$.005 & 0.336$\pm$.009 & 0.274$\pm$.003 \\
\bottomrule
\end{tabular}
\end{table}

\paragraph{Classification.}
Table~\ref{tab:appendix_mlp_classification} reports the classification results. The number of wins over the control drops relative to linear probing, from 29 to 25 of 63 objective--dataset pairs. This drop reflects a stronger control rather than weaker pre-trained encoders: with the MLP head, the random frozen encoder improves on eight of nine datasets, most strongly on \textit{Heartbeat} (from 66.63 to 73.37). On that dataset, NTP, DINO, and SoftCLT exceed the control under linear probing but fall just below it with the MLP head (73.05, 72.56, and 73.29 vs.\ 73.37), and four of the eight lost wins are by less than 0.35 points, within seed variance. Though we do see that LeJEPA attains the best accuracy on four datasets, with gains of $\Delta=+41.2$ on \textit{UWave} (82.97 vs.\ 41.81), $+15.2$ on \textit{SpokenArabicDigits}, and $+14.7$ on \textit{Handwriting}. This suggests that a non-linear head recovers from random features part of the class information that pre-training makes linearly accessible, which narrows the margin on the datasets where the advantage of pre-training was small.
\begin{table}[ht]
\centering
\caption{Classification accuracy (\%) using an MLP probe on a frozen Monash-pretrained backbone. Top result per dataset is \textbf{bold}.}
\label{tab:appendix_mlp_classification}
\footnotesize
\setlength{\tabcolsep}{4pt}
\resizebox{\textwidth}{!}{%
\begin{tabular}{lccccccccc}
\toprule
\textbf{SSL Paradigm} & \textbf{Eth} & \textbf{Face} & \textbf{Hand} & \textbf{Hrt} & \textbf{Vwl} & \textbf{S1} & \textbf{S2} & \textbf{Dig} & \textbf{UW} \\
\midrule
\textit{Random frozen encoder} & 27.84$\pm$0.90 & 52.93$\pm$0.30 & 6.54$\pm$1.50 & \textbf{73.37$\pm$0.20} & 94.97$\pm$0.20 & \textbf{78.23$\pm$1.00} & 51.67$\pm$1.20 & 80.18$\pm$1.00 & 41.81$\pm$1.30 \\
\midrule
\multicolumn{10}{l}{\textit{Signal-space predictive and reconstructive}} \\
NTP       & 29.47$\pm$0.60 & 52.64$\pm$0.50 & 10.14$\pm$0.50 & 73.05$\pm$2.20 & 93.58$\pm$0.50 & 73.47$\pm$1.10 & 49.44$\pm$1.00 & 86.71$\pm$0.40 & 63.51$\pm$2.50 \\
MAE       & 27.47$\pm$1.80 & 53.94$\pm$1.20 & 6.77$\pm$0.40 & 69.76$\pm$2.10 & \textbf{95.95$\pm$0.80} & 74.66$\pm$4.70 & 49.59$\pm$2.30 & 92.72$\pm$0.80 & 47.74$\pm$2.40 \\
Diffusion & 25.10$\pm$0.00 & 52.30$\pm$1.60 & 8.29$\pm$0.80 & 64.27$\pm$2.30 & 92.84$\pm$0.30 & 71.50$\pm$1.00 & \textbf{55.69$\pm$2.20} & 89.27$\pm$1.50 & 55.00$\pm$2.30 \\
\midrule
\multicolumn{10}{l}{\textit{Latent-space alignment}} \\
DINO      & 25.29$\pm$0.30 & 50.00$\pm$0.00 & 5.59$\pm$0.60 & 72.56$\pm$0.20 & 77.71$\pm$1.40 & 72.18$\pm$2.30 & 55.55$\pm$1.00 & 62.27$\pm$1.70 & 33.05$\pm$6.10 \\
JEPA      & 25.10$\pm$0.30 & 51.84$\pm$0.40 & 7.30$\pm$1.80 & 64.15$\pm$2.70 & 93.51$\pm$0.50 & 70.14$\pm$2.20 & 54.72$\pm$1.10 & 83.13$\pm$1.10 & 46.25$\pm$1.40 \\
LeJEPA    & 25.86$\pm$2.00 & \textbf{62.15$\pm$0.80} & \textbf{21.24$\pm$0.40} & 63.66$\pm$3.00 & 90.07$\pm$1.20 & 78.07$\pm$3.90 & 50.70$\pm$1.00 & \textbf{95.37$\pm$0.60} & \textbf{82.97$\pm$0.80} \\
\midrule
\multicolumn{10}{l}{\textit{Contrastive}} \\
SoftCLT   & \textbf{29.85$\pm$2.10} & 50.31$\pm$0.30 & 11.21$\pm$1.90 & 73.29$\pm$0.90 & 81.22$\pm$1.90 & 68.26$\pm$2.10 & 51.67$\pm$1.20 & 65.59$\pm$4.40 & 45.31$\pm$8.70 \\
\bottomrule
\end{tabular}%
}
\end{table}

\paragraph{Anomaly detection.}
Table~\ref{tab:appendix_mlp_anomaly} reports the anomaly-detection results. The MLP head supports better results for the pre-trained encoders: the wins under $\mathrm{F1}_{\mathrm{PA}}$, $\mathrm{F1}_{\mathrm{Ev}}$, and $\mathrm{F1}_{\mathrm{Af}}$ rise to 26, 34, and 29 of 35, from 19, 33, and 19 under linear probing. The wins rise for six of the seven objectives and are unchanged for NTP (13 of 15). For the random encoder the picture is mixed, as it gains on 8 of the 15 dataset--metric cells and drops on 7. This suggests that the MLP head is appropriate for this task, strengthening the pre-trained encoders relative to the baseline.
\begin{table}[ht]
\centering
\footnotesize
\setlength{\tabcolsep}{2pt}
\caption{Anomaly detection F1 (mean~$\pm$~std) using an MLP probe on backbones pre-trained on Monash. For each dataset we report point-adjusted (PA), event-wise (Ev), and affiliation (Af) F1 (\%). Top result per dataset and metric in \textbf{bold}.}
\label{tab:appendix_mlp_anomaly}
\resizebox{\textwidth}{!}{%
\begin{tabular}{l*{15}{c}}
\toprule
\textbf{SSL Paradigm}
& \multicolumn{3}{c}{\textbf{MSL}} & \multicolumn{3}{c}{\textbf{PSM}} & \multicolumn{3}{c}{\textbf{SMAP}} & \multicolumn{3}{c}{\textbf{SMD}} & \multicolumn{3}{c}{\textbf{SWaT}} \\
\cmidrule(lr){2-4}\cmidrule(lr){5-7}\cmidrule(lr){8-10}\cmidrule(lr){11-13}\cmidrule(lr){14-16}
 & PA & Ev & Af & PA & Ev & Af & PA & Ev & Af & PA & Ev & Af & PA & Ev & Af \\
\midrule
\textit{Random frozen encoder} & 16.88$\pm$0.04 & 8.92$\pm$0.12 & 19.95$\pm$0.10 & 48.31$\pm$1.09 & 15.56$\pm$1.36 & 45.67$\pm$0.75 & 15.47$\pm$0.46 & 4.72$\pm$0.13 & 14.07$\pm$0.33 & 50.67$\pm$2.27 & 14.87$\pm$0.65 & 52.70$\pm$1.73 & 11.75$\pm$0.11 & 1.14$\pm$0.17 & 12.99$\pm$2.52 \\
\midrule
\multicolumn{16}{l}{\textit{Signal-space predictive and reconstructive}} \\
NTP & 18.14$\pm$0.14 & 10.03$\pm$0.18 & 20.69$\pm$0.11 & 66.99$\pm$1.01 & 35.51$\pm$2.92 & 58.73$\pm$1.77 & 24.27$\pm$0.30 & 6.53$\pm$0.05 & 19.96$\pm$0.25 & \textbf{64.49$\pm$0.27} & 17.16$\pm$0.14 & \textbf{66.55$\pm$0.21} & 11.64$\pm$0.11 & \textbf{2.24$\pm$0.48} & 9.70$\pm$2.62 \\
MAE & 25.18$\pm$1.37 & 13.09$\pm$0.42 & 26.96$\pm$0.84 & 60.35$\pm$6.06 & \textbf{43.09$\pm$2.49} & 52.06$\pm$3.69 & 24.71$\pm$0.48 & 7.14$\pm$0.21 & 19.64$\pm$0.44 & 62.31$\pm$0.63 & 18.48$\pm$0.30 & 65.30$\pm$0.33 & 11.71$\pm$0.09 & 1.57$\pm$0.20 & \textbf{13.84$\pm$0.30} \\
Diffusion & 45.76$\pm$2.82 & 10.23$\pm$0.57 & 40.53$\pm$1.93 & 42.59$\pm$0.63 & 17.92$\pm$0.96 & 41.19$\pm$1.12 & 18.25$\pm$0.55 & 5.54$\pm$0.18 & 15.84$\pm$0.41 & 51.95$\pm$0.21 & 30.15$\pm$0.13 & 55.45$\pm$0.11 & 11.53$\pm$0.38 & 2.19$\pm$0.15 & 13.63$\pm$0.60 \\
\midrule
\multicolumn{16}{l}{\textit{Latent-space alignment}} \\
DINO & 18.22$\pm$0.57 & 10.75$\pm$0.26 & 20.68$\pm$0.39 & \textbf{70.55$\pm$4.04} & 41.49$\pm$4.71 & \textbf{58.85$\pm$3.24} & 19.19$\pm$0.33 & 5.58$\pm$0.06 & 17.05$\pm$0.16 & 63.40$\pm$0.75 & 15.71$\pm$0.10 & 65.94$\pm$0.62 & \textbf{11.78$\pm$0.04} & 1.75$\pm$0.36 & 13.77$\pm$0.46 \\
JEPA & \textbf{57.11$\pm$0.11} & 12.24$\pm$0.05 & \textbf{49.25$\pm$0.09} & 42.48$\pm$2.00 & 37.63$\pm$1.98 & 38.90$\pm$1.40 & \textbf{37.48$\pm$0.72} & \textbf{7.18$\pm$0.14} & \textbf{29.31$\pm$0.65} & 52.54$\pm$0.15 & \textbf{31.08$\pm$0.19} & 56.08$\pm$0.06 & 3.10$\pm$0.12 & 1.59$\pm$0.13 & 4.08$\pm$0.07 \\
LeJEPA & \textbf{57.11$\pm$0.12} & 12.27$\pm$0.05 & 49.24$\pm$0.08 & 47.51$\pm$0.30 & 42.24$\pm$0.31 & 43.26$\pm$0.14 & 33.78$\pm$0.64 & 6.33$\pm$0.12 & 26.93$\pm$0.33 & 52.63$\pm$0.23 & 30.74$\pm$0.13 & 56.25$\pm$0.08 & 3.12$\pm$0.25 & 0.84$\pm$0.16 & 4.19$\pm$0.24 \\
\midrule
\multicolumn{16}{l}{\textit{Contrastive}} \\
SoftCLT & 30.18$\pm$0.12 & \textbf{13.20$\pm$0.10} & 30.62$\pm$0.08 & 52.88$\pm$1.99 & 33.62$\pm$5.88 & 48.29$\pm$0.91 & 26.77$\pm$1.02 & 6.94$\pm$0.18 & 21.37$\pm$0.77 & 52.23$\pm$0.24 & 29.58$\pm$0.58 & 56.22$\pm$0.29 & 11.25$\pm$0.31 & 1.58$\pm$0.44 & 13.13$\pm$0.48 \\
\bottomrule
\end{tabular}%
}
\end{table}

\paragraph{Summary.}
These results indicate that increasing the capacity of the probing head does not alter the task-level conclusions of our study: with an MLP head, pre-training remains clearly beneficial for anomaly detection (26, 34, and 29 of 35 wins under $\mathrm{F1}_{\mathrm{PA}}$, $\mathrm{F1}_{\mathrm{Ev}}$, and $\mathrm{F1}_{\mathrm{Af}}$), selective for classification (25 of 63), and limited for forecasting (11 of 35). A more expressive head does not compensate for the limitations of the learned representations in forecasting, where only Diffusion shows consistent gains. It does, however, change the ranking between objectives: Diffusion moves from below the random encoder on three of five forecasting datasets under linear probing to above it on all five, and in classification the control benefits from the MLP head about as much as the pre-trained encoders do. Overall, the task-level behavior is driven primarily by the pre-trained representations rather than by the expressiveness of the downstream head, while the objective-level ranking depends on the head, which supports reporting linear and MLP probing side by side.

\subsubsection{Fine-Tuning}
\label{app:results_mlp}
We report results for backbones pre-trained on \textit{Monash} and fine-tuned per dataset.

\paragraph{Forecasting.}
We report the fine-tuning results in Table~\ref{tab:forecasting_finetuning_results}. The gains for the pre-trained encoders are small and not consistent: they beat the baseline in only 3 of 35 comparisons under MSE, all on \textit{Weather}. Relative to the frozen random encoder under linear probing, the from-scratch baseline itself improves on only three of five datasets (\textit{ETTm1}, \textit{ETTm2}, and \textit{Weather}). It does, however, recover the datasets it lost under linear probing: it now attains the best result in 8 of the 10 dataset--metric cells, and the remaining two (\textit{Weather}) are lost by at most 0.002, within seed variance. On \textit{ETTh2}, this recovery is due to the pre-trained encoder rather than the baseline: the baseline is unchanged (0.350), while Diffusion, which led under linear probing (0.334), degrades to 0.354 after fine-tuning. LeJEPA degrades on every dataset (e.g., \textit{ETTh1} MSE 0.505 vs.\ 0.415), and NTP is the second-weakest initialization (\textit{ETTh1} 0.470, \textit{ETTm2} 0.280), indicating that a forecasting pretext task does not transfer to downstream forecasting despite matching it. This gives empirical support to our finding that pre-training does not necessarily produce better representations for forecasting.
\begin{table}[ht]
\centering
\setlength{\tabcolsep}{3pt}
\caption{Full fine-tuning forecasting error (MSE and MAE, mean~$\pm$~std over five seeds, averaged over horizons $\{96,192,336,720\}$) on backbones pre-trained on the \textbf{Monash} dataset. Rows are grouped into signal-space predictive and reconstructive, latent-space alignment, and contrastive objectives; lower is better and the best result per dataset and metric is in \textbf{bold}.}
\label{tab:forecasting_finetuning_results}
\resizebox{\textwidth}{!}{%
\begin{tabular}{l*{10}{c}}
\toprule
\textbf{Objective} & \multicolumn{2}{c}{\textbf{ETTh1}} & \multicolumn{2}{c}{\textbf{ETTh2}} & \multicolumn{2}{c}{\textbf{ETTm1}} & \multicolumn{2}{c}{\textbf{ETTm2}} & \multicolumn{2}{c}{\textbf{Weather}} \\
\cmidrule(lr){2-3}\cmidrule(lr){4-5}\cmidrule(lr){6-7}\cmidrule(lr){8-9}\cmidrule(lr){10-11}
 & MSE & MAE & MSE & MAE & MSE & MAE & MSE & MAE & MSE & MAE \\
\midrule
\textit{Supervised from scratch} & \textbf{0.415$\pm$0.001} & \textbf{0.426$\pm$0.001} & \textbf{0.350$\pm$0.002} & \textbf{0.391$\pm$0.002} & \textbf{0.351$\pm$0.001} & \textbf{0.380$\pm$0.001} & \textbf{0.252$\pm$0.027} & \textbf{0.313$\pm$0.018} & 0.230$\pm$0.002 & 0.265$\pm$0.001 \\
\midrule
\multicolumn{11}{l}{\textit{Signal-space predictive and reconstructive}} \\
NTP & 0.470$\pm$0.008 & 0.462$\pm$0.004 & 0.385$\pm$0.001 & 0.412$\pm$0.000 & 0.356$\pm$0.002 & 0.385$\pm$0.001 & 0.280$\pm$0.001 & 0.332$\pm$0.001 & 0.232$\pm$0.002 & 0.267$\pm$0.001 \\
MAE & 0.438$\pm$0.004 & 0.438$\pm$0.003 & 0.368$\pm$0.002 & 0.403$\pm$0.002 & 0.354$\pm$0.005 & 0.383$\pm$0.001 & 0.261$\pm$0.000 & 0.322$\pm$0.000 & 0.234$\pm$0.003 & 0.268$\pm$0.001 \\
Diffusion & 0.441$\pm$0.018 & 0.450$\pm$0.008 & 0.354$\pm$0.004 & 0.397$\pm$0.003 & 0.359$\pm$0.003 & 0.389$\pm$0.002 & 0.268$\pm$0.001 & 0.324$\pm$0.000 & 0.229$\pm$0.001 & 0.266$\pm$0.001 \\
\midrule
\multicolumn{11}{l}{\textit{Latent-space alignment}} \\
DINO & 0.460$\pm$0.022 & 0.456$\pm$0.012 & 0.376$\pm$0.007 & 0.405$\pm$0.005 & 0.356$\pm$0.001 & 0.386$\pm$0.001 & 0.260$\pm$0.001 & 0.317$\pm$0.001 & 0.230$\pm$0.000 & 0.266$\pm$0.000 \\
JEPA & 0.440$\pm$0.011 & 0.443$\pm$0.007 & 0.383$\pm$0.003 & 0.412$\pm$0.002 & 0.356$\pm$0.002 & 0.384$\pm$0.001 & 0.267$\pm$0.001 & 0.322$\pm$0.002 & \textbf{0.228$\pm$0.000} & \textbf{0.263$\pm$0.001} \\
LeJEPA & 0.505$\pm$0.007 & 0.490$\pm$0.004 & 0.405$\pm$0.006 & 0.432$\pm$0.003 & 0.374$\pm$0.004 & 0.403$\pm$0.003 & 0.281$\pm$0.002 & 0.337$\pm$0.003 & 0.232$\pm$0.001 & 0.269$\pm$0.001 \\
\midrule
\multicolumn{11}{l}{\textit{Contrastive}} \\
SoftCLT & 0.436$\pm$0.002 & 0.449$\pm$0.001 & 0.373$\pm$0.002 & 0.414$\pm$0.003 & 0.361$\pm$0.004 & 0.389$\pm$0.002 & 0.261$\pm$0.001 & 0.320$\pm$0.001 & \textbf{0.228$\pm$0.001} & 0.264$\pm$0.001 \\
\bottomrule
\end{tabular}%
}
\end{table}

\FloatBarrier
\paragraph{Classification.}
Table~\ref{tab:classification_finetuning_results_with_baseline} shows that full fine-tuning reshapes the linear-probing ranking. Pre-trained initializations outperform supervised training from scratch in 42 of 63 objective--dataset pairs, up from 29 under linear probing, and MAE is the only objective that wins on all nine datasets. DINO, which trailed under linear probing, achieves the top result on \textit{FaceDetection} (66 vs.\ 50 for the baseline, $\Delta=+15$) and is strong on \textit{SCP1} (86) and \textit{UWave} (79). LeJEPA retains its linear-probing strengths on gesture data, leading \textit{Handwriting} (25 vs.\ 5, $\Delta=+20$) and \textit{UWave} (85 vs.\ 48, $\Delta=+37$, the largest gain in the table). MAE tops \textit{JapaneseVowels} (97), \textit{SCP1} (87), and \textit{Heartbeat} (74). JEPA leads \textit{SCP2} (57) and \textit{SpokenArabicDigits} (98) but is otherwise among the weakest pre-trained initializations, falling below the baseline on \textit{UWave} (45 vs.\ 48) and \textit{Heartbeat} (61 vs.\ 69). SoftCLT leads \textit{EthanolConcentration} (34) but with high variance ($\pm$11). The baseline remains competitive on \textit{JapaneseVowels} (96) and \textit{SpokenArabicDigits} (96), where the strongest methods are within roughly two points of it.
\begin{table}[ht]
\centering
\setlength{\tabcolsep}{3pt}
\caption{Full fine-tuning classification accuracy (\%, mean~$\pm$~std over five seeds) on backbones pre-trained on the \textbf{Monash} dataset. Datasets: Ethanol (EthanolConcentration), FaceDet (FaceDetection), Handwrit (Handwriting), Heartbt (Heartbeat), JapVowel (JapaneseVowels), SCP1/SCP2 (SelfRegulationSCP1/2), Digits (SpokenArabicDigits), UWave (UWaveGestureLibrary). Rows are grouped into signal-space predictive and reconstructive, latent-space alignment, and contrastive objectives; the best result per dataset is in \textbf{bold}.}
\label{tab:classification_finetuning_results_with_baseline}
\resizebox{\textwidth}{!}{%
\begin{tabular}{l*{9}{c}}
\toprule
\textbf{Objective} & \textbf{Ethanol} & \textbf{FaceDet} & \textbf{Handwrit} & \textbf{Heartbt} & \textbf{JapVowel} & \textbf{SCP1} & \textbf{SCP2} & \textbf{Digits} & \textbf{UWave} \\
\midrule
\textit{Supervised from scratch} & 27.15$\pm$1.81 & 50.40$\pm$0.31 & 4.54$\pm$0.81 & 68.78$\pm$1.11 & 96.38$\pm$0.50 & 78.77$\pm$0.90 & 52.44$\pm$2.29 & 95.94$\pm$0.39 & 48.31$\pm$1.29 \\
\midrule
\multicolumn{10}{l}{\textit{Signal-space predictive and reconstructive}} \\
NTP & 26.23$\pm$2.09 & 52.31$\pm$0.50 & 18.06$\pm$0.77 & 74.02$\pm$2.99 & 95.54$\pm$0.73 & 86.35$\pm$1.98 & 50.84$\pm$1.15 & 97.03$\pm$0.36 & 72.66$\pm$3.73 \\
MAE & 29.09$\pm$1.27 & 50.46$\pm$0.51 & 11.06$\pm$1.86 & \textbf{74.39$\pm$2.79} & \textbf{97.03$\pm$0.69} & \textbf{87.20$\pm$2.43} & 52.78$\pm$1.57 & 97.45$\pm$0.25 & 78.59$\pm$1.72 \\
Diffusion & 24.81$\pm$2.72 & 56.61$\pm$6.36 & 5.27$\pm$1.46 & 62.80$\pm$0.63 & 92.77$\pm$0.72 & 82.25$\pm$2.54 & 55.70$\pm$1.14 & 97.86$\pm$0.19 & 76.56$\pm$1.29 \\
\midrule
\multicolumn{10}{l}{\textit{Latent-space alignment}} \\
DINO & 26.52$\pm$0.78 & \textbf{65.74$\pm$0.55} & 10.26$\pm$0.84 & 72.56$\pm$0.40 & 83.44$\pm$1.33 & 86.09$\pm$1.57 & 56.25$\pm$0.99 & 96.84$\pm$0.18 & 79.14$\pm$1.69 \\
JEPA & 25.00$\pm$0.17 & 51.39$\pm$0.97 & 7.65$\pm$1.65 & 61.10$\pm$4.72 & 93.04$\pm$1.05 & 71.50$\pm$3.42 & \textbf{57.36$\pm$1.82} & \textbf{98.30$\pm$0.24} & 44.53$\pm$3.58 \\
LeJEPA & 25.10$\pm$0.27 & 51.74$\pm$1.22 & \textbf{25.09$\pm$1.03} & 63.90$\pm$2.13 & 93.17$\pm$0.80 & 78.24$\pm$2.56 & 50.69$\pm$2.46 & 97.11$\pm$0.19 & \textbf{85.39$\pm$0.81} \\
\midrule
\multicolumn{10}{l}{\textit{Contrastive}} \\
SoftCLT & \textbf{34.32$\pm$10.59} & 54.12$\pm$5.77 & 11.56$\pm$2.84 & 72.68$\pm$0.34 & 88.31$\pm$1.43 & 72.02$\pm$1.34 & 54.02$\pm$2.20 & 89.21$\pm$6.59 & 54.53$\pm$12.61 \\
\bottomrule
\end{tabular}%
}
\end{table}

\FloatBarrier
\paragraph{Anomaly detection.}
We report the fine-tuning results in Table~\ref{tab:anomaly_finetuning_results}. The pre-trained encoders beat the from-scratch baseline in 21, 24, and 24 of 35 comparisons under $\mathrm{F1}_{\mathrm{PA}}$, $\mathrm{F1}_{\mathrm{Ev}}$, and $\mathrm{F1}_{\mathrm{Af}}$, against 19, 33, and 19 over the random frozen encoder under linear probing, and six of the seven objectives win on a majority of the 15 dataset--metric cells. Fine-tuning preserves the linear-probing advantage of JEPA and LeJEPA on spacecraft telemetry: they lead \textit{MSL} on all three metrics (PA 57.3, Ev 12.4, Af 49.5 vs.\ 49.4/10.2/44.2 for the baseline), and JEPA leads \textit{SMAP} on PA (36.2) and Af (28.2). On \textit{PSM} the picture changes from linear probing: Diffusion (PA 48.3, Af 45.5) and LeJEPA (PA 48.1, Ev 43.3, Af 43.6) are far ahead of both the baseline (PA 28.3$\pm$10.4) and JEPA (PA 20.8), and LeJEPA's event-wise F1 is more than 15 points above any other objective. On \textit{SMD}, NTP and SoftCLT lead the point-adjusted and affiliation metrics (NTP PA 54.2, SoftCLT Af 58.1), whereas the event-wise metric favors JEPA (31.2), LeJEPA (30.6), and Diffusion (30.2), all of which exceed the baseline (20.8) by roughly 10 points. \textit{SWaT} remains an outlier: all other objectives lie within one point of the baseline on PA, while JEPA (7.2) and LeJEPA (8.9) fall below it. DINO is the only objective that degrades under fine-tuning, dropping below the baseline under $\mathrm{F1}_{\mathrm{PA}}$ on every dataset (e.g., \textit{MSL} PA 31.0$\pm$6.5), with seed variance several times larger than that of the other objectives. Overall, pre-training remains beneficial for anomaly detection under full fine-tuning.
\begin{table}[ht]
\centering
\setlength{\tabcolsep}{3pt}
\caption{Full fine-tuning anomaly-detection F1 (\%, mean~$\pm$~std over five seeds) on backbones pre-trained on the \textbf{Monash} dataset. For each dataset we report point-adjusted (PA), event-wise (Ev), and affiliation-based (Af) F1. Rows are grouped into signal-space predictive and reconstructive, latent-space alignment, and contrastive objectives; the best result per dataset and metric is in \textbf{bold}.}
\label{tab:anomaly_finetuning_results}
\resizebox{\textwidth}{!}{%
\begin{tabular}{l*{15}{c}}
\toprule
\textbf{Objective}
& \multicolumn{3}{c}{\textbf{MSL}} & \multicolumn{3}{c}{\textbf{PSM}} & \multicolumn{3}{c}{\textbf{SMAP}} & \multicolumn{3}{c}{\textbf{SMD}} & \multicolumn{3}{c}{\textbf{SWaT}} \\
\cmidrule(lr){2-4}\cmidrule(lr){5-7}\cmidrule(lr){8-10}\cmidrule(lr){11-13}\cmidrule(lr){14-16}
& PA & Ev & Af & PA & Ev & Af & PA & Ev & Af & PA & Ev & Af & PA & Ev & Af \\
\midrule
\textit{Supervised from scratch} & 49.44$\pm$0.34 & 10.21$\pm$0.12 & 44.17$\pm$0.36 & 28.29$\pm$10.44 & 25.56$\pm$7.59 & 23.39$\pm$9.33 & 29.21$\pm$0.41 & 6.22$\pm$0.11 & 24.61$\pm$0.30 & 50.25$\pm$0.18 & 20.84$\pm$0.30 & 53.13$\pm$0.84 & \textbf{11.62$\pm$0.11} & 1.40$\pm$0.34 & 13.54$\pm$0.64 \\
\midrule
\multicolumn{16}{l}{\textit{Signal-space predictive and reconstructive}} \\
NTP & 51.05$\pm$2.27 & 11.03$\pm$0.48 & 44.99$\pm$1.65 & 24.06$\pm$2.02 & 27.76$\pm$2.24 & 25.36$\pm$0.81 & 31.40$\pm$1.31 & 6.48$\pm$0.15 & 25.66$\pm$0.92 & \textbf{54.16$\pm$0.70} & 21.91$\pm$0.14 & 57.45$\pm$0.34 & 11.59$\pm$0.06 & 1.40$\pm$0.24 & 13.40$\pm$0.40 \\
MAE & 50.62$\pm$1.63 & 10.75$\pm$0.29 & 44.68$\pm$1.13 & 20.35$\pm$0.97 & 26.23$\pm$0.76 & 23.01$\pm$0.61 & 30.20$\pm$0.32 & 5.55$\pm$0.09 & 25.18$\pm$0.31 & 52.51$\pm$0.79 & 21.65$\pm$0.33 & 56.12$\pm$0.42 & \textbf{11.62$\pm$0.11} & 1.26$\pm$0.05 & 13.51$\pm$0.47 \\
Diffusion & 54.01$\pm$1.19 & 11.64$\pm$0.25 & 46.05$\pm$1.14 & \textbf{48.30$\pm$0.29} & 25.01$\pm$0.78 & \textbf{45.46$\pm$0.05} & 34.96$\pm$0.33 & 7.32$\pm$0.05 & 27.80$\pm$0.30 & 51.48$\pm$0.46 & 30.24$\pm$0.31 & 54.63$\pm$0.47 & 11.38$\pm$0.36 & 1.19$\pm$0.04 & 13.02$\pm$0.86 \\
\midrule
\multicolumn{16}{l}{\textit{Latent-space alignment}} \\
DINO & 30.99$\pm$6.51 & 7.61$\pm$0.80 & 32.70$\pm$4.43 & 18.28$\pm$8.64 & 20.34$\pm$7.38 & 19.66$\pm$8.13 & 24.36$\pm$6.19 & \textbf{8.66$\pm$1.25} & 22.15$\pm$3.72 & 46.20$\pm$3.51 & 19.45$\pm$1.41 & 49.96$\pm$2.95 & 11.59$\pm$0.13 & 1.44$\pm$0.50 & \textbf{13.81$\pm$0.37} \\
JEPA & \textbf{57.32$\pm$0.08} & \textbf{12.38$\pm$0.02} & \textbf{49.48$\pm$0.08} & 20.80$\pm$0.90 & 26.71$\pm$0.62 & 23.86$\pm$0.39 & \textbf{36.20$\pm$0.37} & 6.85$\pm$0.06 & \textbf{28.20$\pm$0.25} & 52.70$\pm$0.00 & \textbf{31.17$\pm$0.01} & 56.04$\pm$0.00 & 7.18$\pm$0.00 & 0.75$\pm$0.14 & 8.68$\pm$0.03 \\
LeJEPA & 57.28$\pm$0.13 & 12.35$\pm$0.11 & 49.44$\pm$0.10 & 48.06$\pm$0.75 & \textbf{43.34$\pm$1.61} & 43.61$\pm$0.69 & 31.95$\pm$1.33 & 6.50$\pm$0.18 & 25.46$\pm$0.85 & 52.19$\pm$0.00 & 30.55$\pm$0.06 & 56.16$\pm$0.01 & 8.88$\pm$1.98 & 0.78$\pm$0.18 & 10.73$\pm$2.19 \\
\midrule
\multicolumn{16}{l}{\textit{Contrastive}} \\
SoftCLT & 53.76$\pm$2.41 & 11.67$\pm$0.43 & 46.66$\pm$1.73 & 32.46$\pm$3.25 & 25.37$\pm$1.53 & 32.28$\pm$1.58 & 33.72$\pm$1.33 & 6.49$\pm$0.15 & 27.37$\pm$0.92 & 53.93$\pm$0.35 & 25.71$\pm$1.17 & \textbf{58.07$\pm$0.62} & 11.17$\pm$0.35 & \textbf{1.82$\pm$0.85} & 11.21$\pm$1.69 \\
\bottomrule
\end{tabular}%
}
\end{table}

\paragraph{Summary.}
Under full fine-tuning, no single objective dominates across tasks, and the results highlight a clear distinction between tasks. In anomaly detection, JEPA, LeJEPA, and Diffusion take the top score on most dataset--metric pairs (9 of 15), while DINO becomes unstable and falls below the baseline under $\mathrm{F1}_{\mathrm{PA}}$. In classification, the ranking is dataset-specific: MAE, LeJEPA, JEPA, DINO, and SoftCLT each lead on at least one benchmark, MAE is the only objective that beats the baseline on all nine datasets, and the gap to the baseline is largest on the gesture datasets. For these two tasks, pre-training provides a strong initialization that can be effectively exploited through full fine-tuning. In contrast, for forecasting, the benefits of pre-training remain limited even when all parameters are optimized: the supervised-from-scratch baseline achieves the lowest MSE and MAE on four of the five datasets, and no pre-trained model improves on it except by a marginal 0.001--0.002 on \textit{Weather}. This suggests that current SSL objectives do not capture the inductive biases required for accurate temporal extrapolation.

\subsection{Pre-Training Source and Scale}
\label{subsec:source_scale}
This section gives the extended results for Table~\ref{tab:comp_dataset}.
\subsubsection{Overview}
\label{sec:res_data}
Figure~\ref{fig:data_composition_all} summarizes performance under Monash,
Monash-sized synthetic, Large Synthetic, In-domain and Hybrid pre-training.
Because Monash contains approximately 680K samples, whereas Large Synthetic
and Hybrid contain 2.5M samples, we use the matched-scale comparisons to
separate data source from corpus size. Throughout these experiments, the
effect of data composition varies by task. At matched scale, Monash-sized
synthetic pre-training is often competitive with Monash pre-training,
although the two are not uniformly equivalent. At the larger 2.5M-sample
scale, Large Synthetic and Hybrid pre-training also produce similar results
for many objectives, suggesting limited benefit from adding the evaluated
real-data component in those settings. For example, LeJEPA on \textit{UWave}
obtains an accuracy of 63.3\% with Large Synthetic pre-training, 71.2\% with
Monash-sized synthetic pre-training, and 71.69\% with Monash pre-training.
Overall, these results support synthetic data as a useful pre-training
source within the evaluated regime, but do not establish that transfer is
independent of data origin: the objective–dataset exceptions above show
that source and scale interact.

\subsubsection{Extended Forecasting Results (Linear Probing) }\label{subsec:appendix_ext_forecasting}

\paragraph{In-domain pre-training}
In-domain pre-training (Table~\ref{tab:forecasting_mse_indomain}) lowers MSE relative to Monash in 34 of 49 objective--dataset comparisons, and lowers it relative to all other schemes for NTP, MAE, and SoftCLT on nearly all datasets. It is the only regime in which several objectives match or beat the random encoder, led by NTP on Traffic ($0.410$ vs.\ $0.435$) and ETTm1 ($0.341$ vs.\ $0.356$), and the benefit is broadest on Weather, where six of seven objectives outperform random. Still, the random encoder remains better in 34 of the 49 comparisons, with one tie; JEPA does not benefit on any ETT dataset; and LeJEPA's ETTh1 MSE falls from $0.616$ to $0.486$ while the random encoder stays at $0.414$.
\begin{table}[ht]
\centering
\setlength{\tabcolsep}{3pt}
\caption{Linear-probing forecasting MSE and MAE (mean~$\pm$~std over five seeds, averaged over horizons $\{96,192,336,720\}$) using an 8-layer backbone pre-trained \textbf{in-domain} (on each target dataset). Lower is better; best per dataset and metric is in \textbf{bold}.}
\label{tab:forecasting_mse_indomain}
\resizebox{\textwidth}{!}{%
\begin{tabular}{l*{14}{c}}
\toprule
\textbf{Objective}
& \multicolumn{2}{c}{\textbf{ETTh1}} & \multicolumn{2}{c}{\textbf{ETTh2}} & \multicolumn{2}{c}{\textbf{ETTm1}} & \multicolumn{2}{c}{\textbf{ETTm2}} & \multicolumn{2}{c}{\textbf{Weather}} & \multicolumn{2}{c}{\textbf{Electricity}} & \multicolumn{2}{c}{\textbf{Traffic}} \\
\cmidrule(lr){2-3}\cmidrule(lr){4-5}\cmidrule(lr){6-7}\cmidrule(lr){8-9}\cmidrule(lr){10-11}\cmidrule(lr){12-13}\cmidrule(lr){14-15}
& MSE & MAE & MSE & MAE & MSE & MAE & MSE & MAE & MSE & MAE & MSE & MAE & MSE & MAE \\
\midrule
\textit{Random frozen encoder} & \textbf{0.414$\pm$0.002} & \textbf{0.426$\pm$0.002} & 0.350$\pm$0.002 & 0.393$\pm$0.001 & 0.356$\pm$0.001 & 0.379$\pm$0.001 & 0.261$\pm$0.001 & 0.318$\pm$0.000 & 0.248$\pm$0.000 & 0.281$\pm$0.001 & 0.172$\pm$0.000 & 0.267$\pm$0.000 & 0.435$\pm$0.000 & 0.298$\pm$0.000 \\
\midrule
\multicolumn{15}{l}{\textit{Signal-space predictive and reconstructive}} \\
NTP & 0.426$\pm$0.002 & 0.434$\pm$0.001 & 0.351$\pm$0.001 & 0.392$\pm$0.001 & \textbf{0.341$\pm$0.004} & \textbf{0.375$\pm$0.002} & 0.264$\pm$0.001 & 0.320$\pm$0.001 & \textbf{0.235$\pm$0.001} & \textbf{0.269$\pm$0.001} & \textbf{0.165$\pm$0.000} & \textbf{0.258$\pm$0.000} & \textbf{0.410$\pm$0.001} & \textbf{0.280$\pm$0.000} \\
MAE & 0.424$\pm$0.004 & 0.436$\pm$0.003 & 0.349$\pm$0.002 & 0.389$\pm$0.001 & 0.356$\pm$0.003 & 0.386$\pm$0.002 & \textbf{0.256$\pm$0.001} & 0.317$\pm$0.001 & 0.240$\pm$0.001 & 0.275$\pm$0.001 & 0.182$\pm$0.002 & 0.280$\pm$0.002 & 0.447$\pm$0.003 & 0.326$\pm$0.005 \\
Diffusion & 0.429$\pm$0.002 & 0.440$\pm$0.002 & \textbf{0.336$\pm$0.001} & \textbf{0.385$\pm$0.001} & 0.362$\pm$0.001 & 0.391$\pm$0.001 & 0.264$\pm$0.001 & 0.321$\pm$0.001 & 0.237$\pm$0.000 & 0.274$\pm$0.000 & 0.198$\pm$0.001 & 0.308$\pm$0.001 & 0.458$\pm$0.001 & 0.338$\pm$0.002 \\
\midrule
\multicolumn{15}{l}{\textit{Latent-space alignment}} \\
DINO & 0.418$\pm$0.002 & 0.429$\pm$0.002 & 0.358$\pm$0.001 & 0.396$\pm$0.001 & 0.352$\pm$0.001 & 0.383$\pm$0.001 & 0.263$\pm$0.002 & 0.317$\pm$0.002 & 0.238$\pm$0.001 & 0.274$\pm$0.001 & 0.176$\pm$0.003 & 0.273$\pm$0.005 & 0.437$\pm$0.001 & 0.310$\pm$0.001 \\
JEPA & 0.437$\pm$0.002 & 0.437$\pm$0.001 & 0.364$\pm$0.002 & 0.405$\pm$0.001 & 0.370$\pm$0.001 & 0.388$\pm$0.000 & 0.275$\pm$0.001 & 0.333$\pm$0.000 & 0.237$\pm$0.000 & 0.271$\pm$0.000 & 0.188$\pm$0.003 & 0.287$\pm$0.003 & 0.453$\pm$0.003 & 0.326$\pm$0.003 \\
LeJEPA & 0.486$\pm$0.004 & 0.477$\pm$0.003 & 0.384$\pm$0.003 & 0.415$\pm$0.002 & 0.377$\pm$0.003 & 0.401$\pm$0.002 & 0.279$\pm$0.001 & 0.337$\pm$0.001 & 0.253$\pm$0.001 & 0.287$\pm$0.000 & 0.733$\pm$0.007 & 0.701$\pm$0.004 & 0.691$\pm$0.086 & 0.505$\pm$0.063 \\
\midrule
\multicolumn{15}{l}{\textit{Contrastive}} \\
SoftCLT & 0.438$\pm$0.006 & 0.444$\pm$0.004 & 0.354$\pm$0.002 & 0.392$\pm$0.002 & 0.365$\pm$0.002 & 0.388$\pm$0.001 & 0.260$\pm$0.003 & \textbf{0.314$\pm$0.002} & 0.243$\pm$0.001 & 0.276$\pm$0.001 & 0.217$\pm$0.012 & 0.325$\pm$0.014 & 0.454$\pm$0.003 & 0.324$\pm$0.004 \\
\bottomrule
\end{tabular}%
}
\end{table}

\paragraph{Impact of Pre-training Data Composition.}
Figures~\ref{fig:forecasting_data_regime_etth}--\ref{fig:forecasting_data_regime_traffic} and Tables~\ref{tab:forecasting_l8_monash_clean},\ref{tab:forecasting_l8_synthetic_clean}---
\ref{tab:fc_monashsize_equal} report the effect of pre-training data composition on forecasting performance across datasets. Overall, the effect of data source is dataset-dependent and generally modest.

For ETTh1 and ETTh2, Synthetic and Hybrid pre-training often provide slight improvements over Monash for several methods, although the relative differences remain small. Among signal-space objectives, Diffusion is particularly strong on ETTh2, while \textit{Latent-space alignment} (LeJEPA) consistently underperforms on ETTh1.

On the ETTm datasets, the ranking between methods remains largely stable across data regimes. On ETTm1, \textit{Signal-space predictive and reconstructive} methods (NTP) are consistently among the strongest performers, while on ETTm2, \textit{Latent-space alignment} (DINO) achieves the best results across all regimes, with only minor variation between Monash, Synthetic, and Hybrid pre-training.

Weather and Electricity exhibit similarly limited sensitivity to data composition. On these datasets, \textit{Latent-space alignment} (JEPA) and \textit{Signal-space predictive and reconstructive} methods (NTP) are generally competitive, while LeJEPA consistently underperforms, particularly on Electricity.

Traffic exhibits the largest sensitivity to data composition. In particular, LeJEPA degrades substantially under Synthetic and Hybrid pre-training, JEPA performs better under Monash (0.442) than under Synthetic or Hybrid (0.456/0.458), whereas NTP improves slightly under Synthetic and Hybrid (0.434 vs. 0.443). 

Overall, these results reinforce that data composition alone does not provide a uniform improvement in forecasting performance. Instead, its effect depends strongly on the interaction between dataset characteristics and the SSL objective, with no single pre-training regime consistently dominating across benchmarks.
\begin{table}[ht]
\centering
\setlength{\tabcolsep}{3pt}
\caption{Linear-probing forecasting error (MSE and MAE, mean~$\pm$~std over five seeds, averaged over horizons $\{96,192,336,720\}$) using an 8-layer backbone pre-trained on the \textbf{large synthetic} corpus (2.5M samples). The first row is a randomly initialized frozen encoder baseline; the remaining rows are grouped into signal-space predictive and reconstructive, latent-space alignment, and contrastive objectives. Lower is better and the best result per dataset and metric (including the baseline) is in \textbf{bold}; ties at the reported precision are both bolded.}
\label{tab:forecasting_l8_synthetic_clean}
\resizebox{\textwidth}{!}{%
\begin{tabular}{l*{14}{c}}
\toprule
\textbf{Objective} & \multicolumn{2}{c}{\textbf{ETTh1}} & \multicolumn{2}{c}{\textbf{ETTh2}} & \multicolumn{2}{c}{\textbf{ETTm1}} & \multicolumn{2}{c}{\textbf{ETTm2}} & \multicolumn{2}{c}{\textbf{Weather}} & \multicolumn{2}{c}{\textbf{Electricity}} & \multicolumn{2}{c}{\textbf{Traffic}} \\
\cmidrule(lr){2-3}\cmidrule(lr){4-5}\cmidrule(lr){6-7}\cmidrule(lr){8-9}\cmidrule(lr){10-11}\cmidrule(lr){12-13}\cmidrule(lr){14-15}
 & MSE & MAE & MSE & MAE & MSE & MAE & MSE & MAE & MSE & MAE & MSE & MAE & MSE & MAE \\
\midrule
\textit{Random frozen encoder} & \textbf{0.414$\pm$0.002} & \textbf{0.426$\pm$0.002} & 0.350$\pm$0.002 & \textbf{0.393$\pm$0.001} & 0.356$\pm$0.001 & \textbf{0.379$\pm$0.001} & 0.261$\pm$0.001 & 0.318$\pm$0.000 & 0.248$\pm$0.000 & 0.281$\pm$0.001 & \textbf{0.172$\pm$0.000} & \textbf{0.267$\pm$0.000} & 0.435$\pm$0.000 & \textbf{0.298$\pm$0.000} \\
\midrule
\multicolumn{15}{l}{\textit{Signal-space predictive and reconstructive}} \\
NTP & 0.427$\pm$0.002 & 0.435$\pm$0.002 & 0.352$\pm$0.001 & 0.394$\pm$0.001 & \textbf{0.350$\pm$0.001} & \textbf{0.379$\pm$0.001} & 0.260$\pm$0.001 & 0.318$\pm$0.001 & 0.237$\pm$0.000 & \textbf{0.271$\pm$0.000} & \textbf{0.172$\pm$0.000} & 0.268$\pm$0.000 & \textbf{0.434$\pm$0.000} & 0.299$\pm$0.001 \\
MAE & 0.435$\pm$0.003 & 0.446$\pm$0.002 & 0.361$\pm$0.002 & 0.396$\pm$0.002 & 0.362$\pm$0.023 & 0.389$\pm$0.014 & 0.269$\pm$0.003 & 0.327$\pm$0.002 & 0.242$\pm$0.011 & 0.280$\pm$0.008 & 0.191$\pm$0.003 & 0.293$\pm$0.004 & 0.465$\pm$0.003 & 0.338$\pm$0.004 \\
Diffusion & 0.486$\pm$0.009 & 0.482$\pm$0.007 & \textbf{0.345$\pm$0.002} & 0.396$\pm$0.003 & 0.390$\pm$0.004 & 0.409$\pm$0.005 & 0.275$\pm$0.003 & 0.331$\pm$0.003 & 0.245$\pm$0.004 & 0.283$\pm$0.003 & 0.327$\pm$0.018 & 0.439$\pm$0.017 & 0.647$\pm$0.023 & 0.463$\pm$0.013 \\
\midrule
\multicolumn{15}{l}{\textit{Latent-space alignment}} \\
DINO & 0.439$\pm$0.008 & 0.448$\pm$0.005 & 0.368$\pm$0.006 & 0.404$\pm$0.004 & 0.359$\pm$0.002 & 0.386$\pm$0.001 & \textbf{0.256$\pm$0.001} & \textbf{0.314$\pm$0.000} & \textbf{0.236$\pm$0.001} & 0.273$\pm$0.001 & 0.185$\pm$0.003 & 0.286$\pm$0.004 & 0.452$\pm$0.006 & 0.323$\pm$0.007 \\
JEPA & 0.431$\pm$0.002 & 0.439$\pm$0.004 & 0.360$\pm$0.004 & 0.404$\pm$0.007 & 0.369$\pm$0.003 & 0.389$\pm$0.004 & 0.280$\pm$0.008 & 0.336$\pm$0.008 & 0.255$\pm$0.008 & 0.288$\pm$0.008 & 0.195$\pm$0.002 & 0.294$\pm$0.002 & 0.456$\pm$0.003 & 0.327$\pm$0.001 \\
LeJEPA & 0.669$\pm$0.008 & 0.561$\pm$0.004 & 0.400$\pm$0.004 & 0.428$\pm$0.003 & 0.460$\pm$0.005 & 0.452$\pm$0.003 & 0.287$\pm$0.002 & 0.345$\pm$0.002 & 0.265$\pm$0.002 & 0.300$\pm$0.001 & 0.433$\pm$0.018 & 0.505$\pm$0.015 & 0.925$\pm$0.027 & 0.585$\pm$0.014 \\
\midrule
\multicolumn{15}{l}{\textit{Contrastive}} \\
SoftCLT & 0.457$\pm$0.002 & 0.463$\pm$0.002 & 0.390$\pm$0.002 & 0.422$\pm$0.003 & 0.395$\pm$0.002 & 0.412$\pm$0.001 & 0.275$\pm$0.002 & 0.332$\pm$0.002 & 0.246$\pm$0.001 & 0.282$\pm$0.001 & 0.223$\pm$0.003 & 0.321$\pm$0.003 & 0.524$\pm$0.004 & 0.390$\pm$0.004 \\

\bottomrule
\end{tabular}%
}
\end{table}
\begin{table}[ht]
\centering
\setlength{\tabcolsep}{3pt}
\caption{Linear-probing forecasting error (MSE and MAE, mean~$\pm$~std over five seeds, averaged over horizons $\{96,192,336,720\}$) using an 8-layer backbone pre-trained on the \textbf{hybrid} mixture of Monash and synthetic data (2.5M samples). The first row is a randomly initialized frozen encoder baseline; the remaining rows are grouped into signal-space predictive and reconstructive, latent-space alignment, and contrastive objectives. Lower is better and the best result per dataset and metric (including the baseline) is in \textbf{bold}; ties at the reported precision are both bolded.}
\label{tab:forecasting_l8_mixed_clean}
\resizebox{\textwidth}{!}{%
\begin{tabular}{l*{14}{c}}
\toprule
\textbf{Objective} & \multicolumn{2}{c}{\textbf{ETTh1}} & \multicolumn{2}{c}{\textbf{ETTh2}} & \multicolumn{2}{c}{\textbf{ETTm1}} & \multicolumn{2}{c}{\textbf{ETTm2}} & \multicolumn{2}{c}{\textbf{Weather}} & \multicolumn{2}{c}{\textbf{Electricity}} & \multicolumn{2}{c}{\textbf{Traffic}} \\
\cmidrule(lr){2-3}\cmidrule(lr){4-5}\cmidrule(lr){6-7}\cmidrule(lr){8-9}\cmidrule(lr){10-11}\cmidrule(lr){12-13}\cmidrule(lr){14-15}
 & MSE & MAE & MSE & MAE & MSE & MAE & MSE & MAE & MSE & MAE & MSE & MAE & MSE & MAE \\
\midrule
\textit{Random frozen encoder} & \textbf{0.414$\pm$0.002} & \textbf{0.426$\pm$0.002} & 0.350$\pm$0.002 & 0.393$\pm$0.001 & 0.356$\pm$0.001 & \textbf{0.379$\pm$0.001} & 0.261$\pm$0.001 & \textbf{0.318$\pm$0.000} & 0.248$\pm$0.000 & 0.281$\pm$0.001 & \textbf{0.172$\pm$0.000} & \textbf{0.267$\pm$0.000} & 0.435$\pm$0.000 & \textbf{0.298$\pm$0.000} \\
\midrule
\multicolumn{15}{l}{\textit{Signal-space predictive and reconstructive}} \\
NTP & 0.431$\pm$0.001 & 0.438$\pm$0.002 & 0.355$\pm$0.001 & 0.396$\pm$0.001 & \textbf{0.350$\pm$0.001} & \textbf{0.379$\pm$0.001} & \textbf{0.260$\pm$0.000} & \textbf{0.318$\pm$0.001} & \textbf{0.235$\pm$0.000} & \textbf{0.270$\pm$0.000} & 0.173$\pm$0.000 & 0.269$\pm$0.000 & \textbf{0.434$\pm$0.001} & 0.300$\pm$0.001 \\
MAE & 0.431$\pm$0.003 & 0.441$\pm$0.002 & 0.361$\pm$0.003 & 0.398$\pm$0.002 & 0.369$\pm$0.003 & 0.394$\pm$0.002 & 0.265$\pm$0.002 & 0.324$\pm$0.001 & 0.241$\pm$0.001 & 0.277$\pm$0.002 & 0.186$\pm$0.001 & 0.285$\pm$0.002 & 0.463$\pm$0.011 & 0.333$\pm$0.011 \\
Diffusion & 0.465$\pm$0.006 & 0.470$\pm$0.004 & \textbf{0.339$\pm$0.002} & \textbf{0.390$\pm$0.002} & 0.401$\pm$0.012 & 0.417$\pm$0.007 & 0.272$\pm$0.002 & 0.329$\pm$0.002 & 0.251$\pm$0.003 & 0.290$\pm$0.003 & 0.307$\pm$0.019 & 0.424$\pm$0.017 & 0.606$\pm$0.018 & 0.449$\pm$0.014 \\
\midrule
\multicolumn{15}{l}{\textit{Latent-space alignment}} \\
DINO & 0.442$\pm$0.005 & 0.449$\pm$0.004 & 0.372$\pm$0.003 & 0.407$\pm$0.001 & 0.353$\pm$0.001 & 0.384$\pm$0.001 & 0.265$\pm$0.001 & 0.321$\pm$0.001 & \textbf{0.235$\pm$0.001} & 0.271$\pm$0.001 & 0.184$\pm$0.001 & 0.284$\pm$0.002 & 0.453$\pm$0.004 & 0.327$\pm$0.004 \\
JEPA & 0.433$\pm$0.003 & 0.440$\pm$0.002 & 0.361$\pm$0.005 & 0.404$\pm$0.004 & 0.370$\pm$0.003 & 0.390$\pm$0.002 & 0.278$\pm$0.006 & 0.334$\pm$0.005 & 0.255$\pm$0.007 & 0.288$\pm$0.007 & 0.196$\pm$0.003 & 0.295$\pm$0.003 & 0.458$\pm$0.004 & 0.327$\pm$0.003 \\
LeJEPA & 0.672$\pm$0.006 & 0.562$\pm$0.003 & 0.402$\pm$0.006 & 0.430$\pm$0.003 & 0.466$\pm$0.002 & 0.455$\pm$0.001 & 0.292$\pm$0.001 & 0.350$\pm$0.001 & 0.264$\pm$0.002 & 0.299$\pm$0.002 & 0.434$\pm$0.008 & 0.507$\pm$0.006 & 0.896$\pm$0.008 & 0.573$\pm$0.007 \\
\midrule
\multicolumn{15}{l}{\textit{Contrastive}} \\
SoftCLT & 0.456$\pm$0.003 & 0.462$\pm$0.003 & 0.385$\pm$0.001 & 0.418$\pm$0.001 & 0.386$\pm$0.004 & 0.406$\pm$0.003 & 0.266$\pm$0.001 & 0.325$\pm$0.003 & 0.245$\pm$0.001 & 0.280$\pm$0.001 & 0.215$\pm$0.001 & 0.315$\pm$0.001 & 0.507$\pm$0.003 & 0.377$\pm$0.001 \\
\bottomrule
\end{tabular}%
}
\end{table}
\begin{table}[ht]
\centering
\setlength{\tabcolsep}{3pt}
\caption{Linear-probing forecasting error (MSE and MAE, mean~$\pm$~std over five seeds, averaged over horizons $\{96,192,336,720\}$) using an 8-layer backbone pre-trained on the \textbf{Monash-sized synthetic} corpus (matched scale, per-model configuration). Rows are grouped into signal-space predictive and reconstructive, latent-space alignment, and contrastive objectives; lower is better and the best result per dataset and metric is in \textbf{bold}.}
\label{tab:fc_monashsize_equal}
\resizebox{\textwidth}{!}{%
\begin{tabular}{l*{14}{c}}
\toprule
\textbf{Objective} & \multicolumn{2}{c}{\textbf{ETTh1}} & \multicolumn{2}{c}{\textbf{ETTh2}} & \multicolumn{2}{c}{\textbf{ETTm1}} & \multicolumn{2}{c}{\textbf{ETTm2}} & \multicolumn{2}{c}{\textbf{Weather}} & \multicolumn{2}{c}{\textbf{Electricity}} & \multicolumn{2}{c}{\textbf{Traffic}} \\
\cmidrule(lr){2-3}\cmidrule(lr){4-5}\cmidrule(lr){6-7}\cmidrule(lr){8-9}\cmidrule(lr){10-11}\cmidrule(lr){12-13}\cmidrule(lr){14-15}
 & MSE & MAE & MSE & MAE & MSE & MAE & MSE & MAE & MSE & MAE & MSE & MAE & MSE & MAE \\
\midrule
\textit{Random frozen encoder} & 0.414$\pm$0.002 & \textbf{0.426$\pm$0.002} & 0.350$\pm$0.002 & 0.393$\pm$0.001 & 0.356$\pm$0.001 & \textbf{0.379$\pm$0.001} & 0.261$\pm$0.001 & 0.318$\pm$0.000 & 0.248$\pm$0.000 & 0.281$\pm$0.001 & \textbf{0.172$\pm$0.000} & \textbf{0.267$\pm$0.000} & 0.435$\pm$0.000 & \textbf{0.298$\pm$0.000} \\
\midrule
\multicolumn{15}{l}{\textit{Signal-space predictive and reconstructive}} \\
NTP & 0.427$\pm$0.001 & 0.436$\pm$0.000 & 0.353$\pm$0.001 & 0.394$\pm$0.001 & \textbf{0.351$\pm$0.001} & 0.380$\pm$0.001 & 0.259$\pm$0.001 & 0.318$\pm$0.001 & \textbf{0.237$\pm$0.001} & \textbf{0.271$\pm$0.000} & \textbf{0.172$\pm$0.000} & 0.268$\pm$0.000 & \textbf{0.433$\pm$0.000} & 0.299$\pm$0.000 \\
MAE & 0.440$\pm$0.002 & 0.450$\pm$0.002 & 0.358$\pm$0.002 & 0.395$\pm$0.001 & 0.371$\pm$0.002 & 0.395$\pm$0.001 & 0.268$\pm$0.001 & 0.325$\pm$0.001 & 0.241$\pm$0.001 & 0.276$\pm$0.001 & 0.192$\pm$0.003 & 0.297$\pm$0.006 & 0.461$\pm$0.004 & 0.336$\pm$0.002 \\
Diffusion & \textbf{0.412$\pm$0.001} & 0.430$\pm$0.001 & \textbf{0.322$\pm$0.001} & \textbf{0.374$\pm$0.001} & 0.368$\pm$0.001 & 0.388$\pm$0.001 & 0.266$\pm$0.002 & 0.321$\pm$0.001 & 0.246$\pm$0.000 & 0.279$\pm$0.000 & 0.181$\pm$0.001 & 0.286$\pm$0.001 & 0.455$\pm$0.002 & 0.333$\pm$0.002 \\
\midrule
\multicolumn{15}{l}{\textit{Latent-space alignment}} \\
DINO & 0.449$\pm$0.011 & 0.452$\pm$0.007 & 0.359$\pm$0.001 & 0.396$\pm$0.001 & 0.357$\pm$0.001 & 0.386$\pm$0.001 & \textbf{0.257$\pm$0.001} & \textbf{0.315$\pm$0.001} & \textbf{0.237$\pm$0.000} & 0.273$\pm$0.001 & 0.179$\pm$0.001 & 0.279$\pm$0.002 & 0.443$\pm$0.003 & 0.314$\pm$0.004 \\
JEPA & 0.432$\pm$0.004 & 0.436$\pm$0.001 & 0.360$\pm$0.008 & 0.401$\pm$0.005 & 0.368$\pm$0.002 & 0.386$\pm$0.002 & 0.271$\pm$0.003 & 0.327$\pm$0.002 & 0.241$\pm$0.003 & 0.274$\pm$0.002 & 0.186$\pm$0.002 & 0.284$\pm$0.002 & 0.447$\pm$0.002 & 0.319$\pm$0.003 \\
LeJEPA & 0.666$\pm$0.010 & 0.560$\pm$0.004 & 0.401$\pm$0.003 & 0.430$\pm$0.002 & 0.463$\pm$0.012 & 0.453$\pm$0.006 & 0.288$\pm$0.003 & 0.346$\pm$0.003 & 0.267$\pm$0.001 & 0.302$\pm$0.001 & 0.432$\pm$0.028 & 0.502$\pm$0.021 & 0.861$\pm$0.025 & 0.550$\pm$0.013 \\
\midrule
\multicolumn{15}{l}{\textit{Contrastive}} \\
SoftCLT & 0.453$\pm$0.001 & 0.460$\pm$0.001 & 0.389$\pm$0.002 & 0.423$\pm$0.001 & 0.392$\pm$0.001 & 0.411$\pm$0.001 & 0.272$\pm$0.002 & 0.330$\pm$0.001 & 0.244$\pm$0.001 & 0.281$\pm$0.000 & 0.219$\pm$0.001 & 0.318$\pm$0.000 & 0.512$\pm$0.002 & 0.382$\pm$0.002 \\
\bottomrule
\end{tabular}%
}
\end{table}

\pgfplotsset{
    unified bar plot/.style={
        ybar,
        bar width=2.4pt,
        width=1.05\linewidth,
        height=3.4cm,
        xtick=data,
        xticklabel style={font=\fontsize{5}{6}\selectfont},
        grid=major,
        grid style={dashed, gray!30},
        enlarge x limits=0.14,
        ylabel style={font=\fontsize{6}{7}\selectfont, xshift=-0.8em},
        yticklabel style={font=\tiny},
        symbolic x coords={NTP, MAE, Diffusion, DINO, JEPA, LeJEPA, SoftCLT},
    },
}

\begin{figure}[htb!]
\centering
\begin{subfigure}[b]{0.48\textwidth}
\centering
\begin{tikzpicture}
    \begin{axis}[
        unified bar plot,
        ylabel={MSE},
        ymin=0.40, ymax=0.72,
        ytick={0.4, 0.5, 0.6, 0.7},
        legend to name=fig:forecasting_data_regime:legend,
        legend columns=-1,
        legend style={font=\fontsize{6}{7}\selectfont, draw=none, /tikz/every even column/.append style={column sep=0.5em}},
    ]
        \addplot[fill=gray!20,   draw=black] coordinates {(NTP,0.435) (MAE,0.438) (Diffusion,0.433) (DINO,0.444) (JEPA,0.433) (LeJEPA,0.616) (SoftCLT,0.456)};
        \addplot[fill=blue!25  , draw=black] coordinates {(NTP,0.427) (MAE,0.440) (Diffusion,0.412) (DINO,0.449) (JEPA,0.432) (LeJEPA,0.666) (SoftCLT,0.453)};
        \addplot[fill=orange!40, draw=black] coordinates {(NTP,0.427) (MAE,0.435) (Diffusion,0.486) (DINO,0.439) (JEPA,0.431) (LeJEPA,0.669) (SoftCLT,0.457)};
        \addplot[fill=teal!40,   draw=black] coordinates {(NTP,0.431) (MAE,0.431) (Diffusion,0.465) (DINO,0.442) (JEPA,0.433) (LeJEPA,0.672) (SoftCLT,0.456)};
        \addplot[fill=violet!50, draw=black] coordinates {(NTP,0.426) (MAE,0.424) (Diffusion,0.429) (DINO,0.418) (JEPA,0.437) (LeJEPA,0.486) (SoftCLT,0.438)};
        \legend{Monash, Small Synthetic, Large Synthetic, Hybrid, In-domain}
    \end{axis}
\end{tikzpicture}
\subcaption{ETTh1}
\end{subfigure}
\hfill
\begin{subfigure}[b]{0.48\textwidth}
\centering
\begin{tikzpicture}
    \begin{axis}[
        unified bar plot,
        ylabel={MSE},
        ymin=0.30, ymax=0.42,
        ytick={0.30, 0.33, 0.36, 0.39, 0.42},
    ]
        \addplot[fill=gray!20,   draw=black] coordinates {(NTP,0.362) (MAE,0.368) (Diffusion,0.334) (DINO,0.368) (JEPA,0.363) (LeJEPA,0.394) (SoftCLT,0.378)};
        \addplot[fill=blue!25  , draw=black] coordinates {(NTP,0.353) (MAE,0.358) (Diffusion,0.322) (DINO,0.359) (JEPA,0.360) (LeJEPA,0.401) (SoftCLT,0.389)};
        \addplot[fill=orange!40, draw=black] coordinates {(NTP,0.352) (MAE,0.361) (Diffusion,0.345) (DINO,0.368) (JEPA,0.360) (LeJEPA,0.400) (SoftCLT,0.390)};
        \addplot[fill=teal!40,   draw=black] coordinates {(NTP,0.355) (MAE,0.361) (Diffusion,0.339) (DINO,0.372) (JEPA,0.361) (LeJEPA,0.402) (SoftCLT,0.385)};
        \addplot[fill=violet!50, draw=black] coordinates {(NTP,0.351) (MAE,0.349) (Diffusion,0.336) (DINO,0.358) (JEPA,0.364) (LeJEPA,0.384) (SoftCLT,0.354)};
    \end{axis}
\end{tikzpicture}
\subcaption{ETTh2}
\end{subfigure}

\vspace{0.05cm}
\ref{fig:forecasting_data_regime:legend}

\caption{Impact of pre-training data composition on forecasting performance (MSE, linear probing, 8-layer backbone) for the ETTh datasets. Bars: Monash (680K), Small Synthetic (680K, matched scale), Large Synthetic (2.5M), Hybrid (2.5M), and In-domain. Lower is better.}
\label{fig:forecasting_data_regime_etth}
\end{figure}

\begin{figure}[htb!]
\centering
\begin{subfigure}[b]{0.48\textwidth}
\centering
\begin{tikzpicture}
    \begin{axis}[
        unified bar plot,
        ylabel={MSE},
        ymin=0.33, ymax=0.49,
        ytick={0.35, 0.40, 0.45},
    ]
        \addplot[fill=gray!20,   draw=black] coordinates {(NTP,0.352) (MAE,0.364) (Diffusion,0.375) (DINO,0.360) (JEPA,0.364) (LeJEPA,0.444) (SoftCLT,0.380)};
        \addplot[fill=blue!25  , draw=black] coordinates {(NTP,0.351) (MAE,0.371) (Diffusion,0.368) (DINO,0.357) (JEPA,0.368) (LeJEPA,0.463) (SoftCLT,0.392)};
        \addplot[fill=orange!40, draw=black] coordinates {(NTP,0.350) (MAE,0.362) (Diffusion,0.390) (DINO,0.359) (JEPA,0.369) (LeJEPA,0.460) (SoftCLT,0.395)};
        \addplot[fill=teal!40,   draw=black] coordinates {(NTP,0.350) (MAE,0.369) (Diffusion,0.401) (DINO,0.353) (JEPA,0.370) (LeJEPA,0.466) (SoftCLT,0.386)};
        \addplot[fill=violet!50, draw=black] coordinates {(NTP,0.341) (MAE,0.356) (Diffusion,0.362) (DINO,0.352) (JEPA,0.370) (LeJEPA,0.377) (SoftCLT,0.365)};
    \end{axis}
\end{tikzpicture}
\subcaption{ETTm1}
\end{subfigure}
\hfill
\begin{subfigure}[b]{0.48\textwidth}
\centering
\begin{tikzpicture}
    \begin{axis}[
        unified bar plot,
        ylabel={MSE},
        ymin=0.24, ymax=0.32,
        ytick={0.24, 0.26, 0.28, 0.30, 0.32},
    ]
        \addplot[fill=gray!20,   draw=black] coordinates {(NTP,0.262) (MAE,0.264) (Diffusion,0.264) (DINO,0.254) (JEPA,0.270) (LeJEPA,0.288) (SoftCLT,0.266)};
        \addplot[fill=blue!25  , draw=black] coordinates {(NTP,0.259) (MAE,0.268) (Diffusion,0.266) (DINO,0.257) (JEPA,0.271) (LeJEPA,0.288) (SoftCLT,0.272)};
        \addplot[fill=orange!40, draw=black] coordinates {(NTP,0.260) (MAE,0.269) (Diffusion,0.275) (DINO,0.256) (JEPA,0.280) (LeJEPA,0.287) (SoftCLT,0.275)};
        \addplot[fill=teal!40,   draw=black] coordinates {(NTP,0.260) (MAE,0.265) (Diffusion,0.272) (DINO,0.265) (JEPA,0.278) (LeJEPA,0.292) (SoftCLT,0.266)};
        \addplot[fill=violet!50, draw=black] coordinates {(NTP,0.264) (MAE,0.256) (Diffusion,0.264) (DINO,0.263) (JEPA,0.275) (LeJEPA,0.279) (SoftCLT,0.260)};
    \end{axis}
\end{tikzpicture}
\subcaption{ETTm2}
\end{subfigure}

\vspace{0.05cm}
\ref{fig:forecasting_data_regime:legend}

\caption{Impact of pre-training data composition on forecasting performance (MSE, linear probing, 8-layer backbone) for the ETTm datasets. Lower is better.}
\label{fig:forecasting_data_regime_ettm}
\end{figure}

\begin{figure}[htb!]
\centering
\begin{subfigure}[b]{0.48\textwidth}
\centering
\begin{tikzpicture}
    \begin{axis}[
        unified bar plot,
        ylabel={MSE},
        ymin=0.22, ymax=0.28,
        ytick={0.22, 0.24, 0.26, 0.28},
    ]
        \addplot[fill=gray!20,   draw=black] coordinates {(NTP,0.235) (MAE,0.244) (Diffusion,0.242) (DINO,0.236) (JEPA,0.245) (LeJEPA,0.262) (SoftCLT,0.243)};
        \addplot[fill=blue!25  , draw=black] coordinates {(NTP,0.237) (MAE,0.241) (Diffusion,0.246) (DINO,0.237) (JEPA,0.241) (LeJEPA,0.267) (SoftCLT,0.244)};
        \addplot[fill=orange!40, draw=black] coordinates {(NTP,0.237) (MAE,0.242) (Diffusion,0.245) (DINO,0.236) (JEPA,0.255) (LeJEPA,0.265) (SoftCLT,0.246)};
        \addplot[fill=teal!40,   draw=black] coordinates {(NTP,0.235) (MAE,0.241) (Diffusion,0.251) (DINO,0.235) (JEPA,0.255) (LeJEPA,0.264) (SoftCLT,0.245)};
        \addplot[fill=violet!50, draw=black] coordinates {(NTP,0.235) (MAE,0.240) (Diffusion,0.237) (DINO,0.238) (JEPA,0.237) (LeJEPA,0.253) (SoftCLT,0.243)};
    \end{axis}
\end{tikzpicture}
\subcaption{Weather}
\end{subfigure}
\hfill
\begin{subfigure}[b]{0.48\textwidth}
\centering
\begin{tikzpicture}
    \begin{axis}[
        unified bar plot,
        ylabel={MSE},
        ymin=0.16, ymax=0.76,
        ytick={0.16, 0.31, 0.46, 0.61, 0.76},
    ]
        \addplot[fill=gray!20,   draw=black] coordinates {(NTP,0.180) (MAE,0.194) (Diffusion,0.227) (DINO,0.182) (JEPA,0.178) (LeJEPA,0.346) (SoftCLT,0.217)};
        \addplot[fill=blue!25  , draw=black] coordinates {(NTP,0.172) (MAE,0.192) (Diffusion,0.181) (DINO,0.179) (JEPA,0.186) (LeJEPA,0.432) (SoftCLT,0.219)};
        \addplot[fill=orange!40, draw=black] coordinates {(NTP,0.172) (MAE,0.191) (Diffusion,0.327) (DINO,0.185) (JEPA,0.195) (LeJEPA,0.433) (SoftCLT,0.223)};
        \addplot[fill=teal!40,   draw=black] coordinates {(NTP,0.173) (MAE,0.186) (Diffusion,0.307) (DINO,0.184) (JEPA,0.196) (LeJEPA,0.434) (SoftCLT,0.215)};
        \addplot[fill=violet!50, draw=black] coordinates {(NTP,0.165) (MAE,0.182) (Diffusion,0.198) (DINO,0.176) (JEPA,0.188) (LeJEPA,0.733) (SoftCLT,0.217)};
    \end{axis}
\end{tikzpicture}
\subcaption{Electricity}
\end{subfigure}

\vspace{0.05cm}
\ref{fig:forecasting_data_regime:legend}

\caption{Impact of pre-training data composition on forecasting performance (MSE, linear probing, 8-layer backbone) for Weather and Electricity. Lower is better.}
\label{fig:forecasting_data_regime_weather_electricity}
\end{figure}

\begin{figure}[htb!]
\centering
\begin{subfigure}[b]{0.48\textwidth}
\centering
\begin{tikzpicture}
    \begin{axis}[
        unified bar plot,
        ylabel={MSE},
        ymin=0.40, ymax=1.0,
        ytick={0.4, 0.55, 0.7, 0.85, 1.0},
    ]
        \addplot[fill=gray!20,   draw=black] coordinates {(NTP,0.445) (MAE,0.481) (Diffusion,0.527) (DINO,0.449) (JEPA,0.442) (LeJEPA,0.848) (SoftCLT,0.488)};
        \addplot[fill=blue!25  , draw=black] coordinates {(NTP,0.433) (MAE,0.461) (Diffusion,0.455) (DINO,0.443) (JEPA,0.447) (LeJEPA,0.861) (SoftCLT,0.512)};
        \addplot[fill=orange!40, draw=black] coordinates {(NTP,0.434) (MAE,0.465) (Diffusion,0.647) (DINO,0.452) (JEPA,0.456) (LeJEPA,0.925) (SoftCLT,0.524)};
        \addplot[fill=teal!40,   draw=black] coordinates {(NTP,0.434) (MAE,0.463) (Diffusion,0.606) (DINO,0.453) (JEPA,0.458) (LeJEPA,0.896) (SoftCLT,0.507)};
        \addplot[fill=violet!50, draw=black] coordinates {(NTP,0.410) (MAE,0.447) (Diffusion,0.458) (DINO,0.437) (JEPA,0.453) (LeJEPA,0.691) (SoftCLT,0.454)};
    \end{axis}
\end{tikzpicture}
\subcaption{Traffic}
\end{subfigure}

\vspace{0.05cm}
\ref{fig:forecasting_data_regime:legend}

\caption{Impact of pre-training data composition on forecasting performance (MSE, linear probing, 8-layer backbone) for Traffic. Lower is better.}
\label{fig:forecasting_data_regime_traffic}
\end{figure}
\FloatBarrier
\subsubsection{Extended Classification Results (Linear Probing)}\label{subsec:appendix_ext_cls}

Tables~\ref{tab:classification_results_refined}, ~\ref{tab:classification_transfer_synthetic}, ~\ref{tab:classification_transfer_mixed}, and ~\ref{tab:cls_monashsize_equal}   report classification performance under the Monash large Synthetic, Hybrid
pre-training, and Monash-sized synthetic (Monash sized) regimes,
respectively. Overall,
SSL methods present substantial gains over the non-pretrained baseline on
several datasets, and the leading objective on each dataset is largely
stable across pre-training sources.

Across all four sources, LeJEPA is the strongest frozen classifier. It
leads \textit{Handwriting}, \textit{SpokenArabicDigits}, and \textit{UWave}
under every source, often by large margins over both the baseline and the
next objective (Handwriting: $13.0$, $\Delta = +8.7$ under large Synthetic,
$16.4$, $\Delta=+12.1$ under Monash-sized synthetic and $12.7$,
$\Delta=+8.4$ under Hybrid), and it leads \textit{FaceDetection} under
Monash and Hybrid and \textit{SCP1} under Monash and large Synthetic. On
\textit{FaceDetection} and \textit{SCP1} the margin to NTP or MAE is
within about one point under every source, so those two leads are not
robust to source.

MAE and NTP lead the remaining datasets. MAE leads \textit{Heartbeat}
under every source (73.0--75.6 vs.\ 66.6 for random features) and
\textit{EthanolConcentration} under Monash and Monash-sized synthetic,
though every Ethanol result lies within the random encoder's seed
variance. NTP leads \textit{JapaneseVowels} under the three
synthetic-containing sources (95.5--96.2 vs.\ 93.0), the one dataset
without a gain under Monash pre-training.

The effect of pre-training source is present but secondary to the choice
of objective. Source moves NTP most: its \textit{UWave} accuracy rises
from 55.0 under Monash to 61.6 and 62.8 under the synthetic sources and
its \textit{FaceDetection} accuracy from 55.0 to 59.1--60.9. LeJEPA's
\textit{UWave} accuracy is 71.69 under Monash and 71.2 under matched
synthetic but 63.3 and 58.4 under large Synthetic and Hybrid, and MAE's
falls from 40.0 to 28.1--31.8 under the synthetic sources. These shifts
change absolute scores; they do not change which objective leads
\textit{Handwriting}, \textit{SpokenArabicDigits}, \textit{UWave}, or
\textit{Heartbeat}.

DINO and JEPA, the other latent-space alignment objectives, do not share
LeJEPA's advantage. DINO is second on \textit{Handwriting} under every
source (10.0--12.0) and leads \textit{SCP2} under Monash (56.0), but
falls 13--17 points below random features on \textit{JapaneseVowels}
under every source. JEPA leads \textit{SCP2} under Monash-sized synthetic
and Hybrid (55.8, 54.5) and is otherwise at or below the random encoder
on most datasets. LeJEPA's classification advantage is therefore a
property of that objective rather than of latent-space alignment.

Seed variance is low for most objectives. LeJEPA's standard deviations
range from 0.4 to 4.9 points, with the largest on \textit{UWave} and
\textit{SCP1}; Diffusion and MAE reach 5--6 points on
\textit{JapaneseVowels} and \textit{SpokenArabicDigits}; SoftCLT is the
least stable, with 9--15 points on \textit{JapaneseVowels},
\textit{SpokenArabicDigits}, and \textit{UWave} under the synthetic
sources.

Overall, classification benefits substantially from SSL, and the
objective that leads each dataset is stable across sources: LeJEPA on
the handwriting and gesture benchmarks, MAE on \textit{Heartbeat}, and
NTP on \textit{JapaneseVowels} where any gain appears. Pre-training
source refines absolute accuracy but does not alter these leaders.
\begin{table}[ht]
\centering
\setlength{\tabcolsep}{3pt}
\caption{Linear-probing classification accuracy (\%, mean~$\pm$~std over five seeds) using an 8-layer backbone pre-trained on the \textbf{large synthetic} corpus (2.5M samples). Datasets: Ethanol (EthanolConcentration), FaceDet (FaceDetection), Handwrit (Handwriting), Heartbt (Heartbeat), JapVowel (JapaneseVowels), SCP1/SCP2 (SelfRegulationSCP1/2), Digits (SpokenArabicDigits), UWave (UWaveGestureLibrary). Rows are grouped into signal-space predictive and reconstructive, latent-space alignment, and contrastive objectives; the best result per dataset is in \textbf{bold}.}
\label{tab:classification_transfer_synthetic}
\resizebox{\textwidth}{!}{%
\begin{tabular}{l*{9}{c}}
\toprule
\textbf{Objective} & \textbf{Ethanol} & \textbf{FaceDet} & \textbf{Handwrit} & \textbf{Heartbt} & \textbf{JapVowel} & \textbf{SCP1} & \textbf{SCP2} & \textbf{Digits} & \textbf{UWave} \\
\midrule
\textit{Random frozen encoder} & 27.68$\pm$1.88 & 54.49$\pm$0.54 & 4.31$\pm$0.54 & 66.63$\pm$1.33 & 93.02$\pm$0.65 & 76.72$\pm$2.86 & 50.00$\pm$2.60 & 77.61$\pm$0.96 & 36.56$\pm$2.78 \\
\midrule
\multicolumn{10}{l}{\textit{Signal-space predictive and reconstructive}} \\
NTP & 27.80$\pm$1.80 & \textbf{60.90$\pm$0.70} & 4.30$\pm$0.70 & 70.50$\pm$4.10 & \textbf{96.20$\pm$0.60} & 75.50$\pm$1.70 & 53.40$\pm$2.30 & 85.30$\pm$0.50 & 62.80$\pm$1.20 \\
MAE & 26.50$\pm$3.70 & 55.60$\pm$0.50 & 4.50$\pm$0.60 & \textbf{74.80$\pm$0.70} & 82.20$\pm$6.10 & 78.00$\pm$1.50 & 52.90$\pm$4.00 & 81.30$\pm$2.10 & 28.60$\pm$2.70 \\
Diffusion & 24.40$\pm$0.80 & 56.30$\pm$0.80 & 7.70$\pm$1.80 & 65.20$\pm$2.60 & 90.30$\pm$0.50 & 66.70$\pm$3.60 & 52.20$\pm$1.30 & 84.20$\pm$1.40 & 56.70$\pm$2.80 \\
\midrule
\multicolumn{10}{l}{\textit{Latent-space alignment}} \\
DINO & 26.80$\pm$1.10 & 53.40$\pm$0.60 & 12.00$\pm$1.30 & 71.80$\pm$0.60 & 79.80$\pm$0.80 & 71.70$\pm$1.50 & 53.60$\pm$1.90 & 82.70$\pm$1.60 & 44.00$\pm$7.70 \\
JEPA & 24.90$\pm$0.30 & 52.50$\pm$0.30 & 4.20$\pm$0.50 & 63.00$\pm$3.70 & 89.00$\pm$3.50 & 66.70$\pm$1.90 & 53.60$\pm$2.30 & 71.90$\pm$3.50 & 27.90$\pm$3.50 \\
LeJEPA & 25.10$\pm$0.50 & 60.80$\pm$0.80 & \textbf{13.00$\pm$1.10} & 66.40$\pm$1.60 & 82.50$\pm$4.70 & \textbf{83.10$\pm$1.30} & 47.10$\pm$1.80 & \textbf{90.80$\pm$1.50} & \textbf{63.30$\pm$4.80} \\
\midrule
\multicolumn{10}{l}{\textit{Contrastive}} \\
SoftCLT & \textbf{28.75$\pm$2.79} & 58.60$\pm$0.66 & 5.34$\pm$0.60 & 72.30$\pm$0.19 & 62.06$\pm$3.92 & 75.70$\pm$4.87 & \textbf{54.22$\pm$1.25} & 67.15$\pm$2.66 & 49.81$\pm$3.14 \\

\bottomrule
\end{tabular}%
}
\end{table}

\begin{table}[ht]
\centering
\setlength{\tabcolsep}{3pt}
\caption{Linear-probing classification accuracy (\%, mean~$\pm$~std over five seeds) using an 8-layer backbone pre-trained on the \textbf{hybrid} mixture of Monash and synthetic data (2.5M samples). Datasets: Ethanol (EthanolConcentration), FaceDet (FaceDetection), Handwrit (Handwriting), Heartbt (Heartbeat), JapVowel (JapaneseVowels), SCP1/SCP2 (SelfRegulationSCP1/2), Digits (SpokenArabicDigits), UWave (UWaveGestureLibrary). Rows are grouped into signal-space predictive and reconstructive, latent-space alignment, and contrastive objectives; the best result per dataset is in \textbf{bold}.}
\label{tab:classification_transfer_mixed}
\resizebox{\textwidth}{!}{%
\begin{tabular}{l*{9}{c}}
\toprule
\textbf{Objective} & \textbf{Ethanol} & \textbf{FaceDet} & \textbf{Handwrit} & \textbf{Heartbt} & \textbf{JapVowel} & \textbf{SCP1} & \textbf{SCP2} & \textbf{Digits} & \textbf{UWave} \\
\midrule
\textit{Random frozen encoder} & 27.68$\pm$1.88 & 54.49$\pm$0.54 & 4.31$\pm$0.54 & 66.63$\pm$1.33 & 93.02$\pm$0.65 & 76.72$\pm$2.86 & 50.00$\pm$2.60 & 77.61$\pm$0.96 & 36.56$\pm$2.78 \\
\midrule
\multicolumn{10}{l}{\textit{Signal-space predictive and reconstructive}} \\
NTP & \textbf{28.44$\pm$1.40} & 59.13$\pm$0.60 & 5.98$\pm$1.20 & 70.64$\pm$1.60 & \textbf{95.52$\pm$0.50} & 74.81$\pm$1.80 & 52.78$\pm$2.50 & 84.57$\pm$1.00 & 54.44$\pm$3.10 \\
MAE & 28.06$\pm$2.50 & 55.83$\pm$0.50 & 5.48$\pm$1.00 & \textbf{75.61$\pm$1.10} & 84.05$\pm$2.00 & \textbf{81.91$\pm$2.00} & 51.44$\pm$1.90 & 84.17$\pm$0.80 & 31.75$\pm$2.50 \\
Diffusion & 24.90$\pm$0.20 & 56.30$\pm$0.70 & 7.10$\pm$1.60 & 63.40$\pm$2.50 & 91.20$\pm$0.70 & 66.60$\pm$2.10 & 51.30$\pm$1.30 & 84.70$\pm$1.60 & 56.10$\pm$5.90 \\
\midrule
\multicolumn{10}{l}{\textit{Latent-space alignment}} \\
DINO & 27.80$\pm$1.10 & 54.00$\pm$0.80 & 11.60$\pm$0.90 & 71.70$\pm$0.70 & 80.30$\pm$0.80 & 68.40$\pm$1.50 & 52.70$\pm$1.70 & 83.50$\pm$1.60 & 44.80$\pm$6.00 \\
JEPA & 24.94$\pm$0.20 & 51.71$\pm$0.50 & 6.07$\pm$1.90 & 63.71$\pm$1.90 & 91.89$\pm$0.70 & 66.48$\pm$3.20 & \textbf{54.45$\pm$3.00} & 75.94$\pm$5.30 & 32.56$\pm$6.60 \\
LeJEPA & 24.87$\pm$0.90 & \textbf{60.85$\pm$0.40} & \textbf{12.66$\pm$1.90} & 64.97$\pm$1.00 & 85.08$\pm$2.70 & 81.50$\pm$4.90 & 47.00$\pm$2.50 & \textbf{92.45$\pm$0.60} & \textbf{58.44$\pm$3.10} \\
\midrule
\multicolumn{10}{l}{\textit{Contrastive}} \\
SoftCLT & 26.62$\pm$0.99 & 56.91$\pm$1.26 & 6.31$\pm$0.52 & 74.34$\pm$1.37 & 74.60$\pm$4.83 & 76.66$\pm$6.92 & 54.11$\pm$2.93 & 84.69$\pm$3.32 & 58.81$\pm$9.16 \\

\bottomrule
\end{tabular}%
}
\end{table}

\begin{table}[ht]
\centering
\setlength{\tabcolsep}{3pt}
\caption{Linear-probing classification accuracy (\%, mean~$\pm$~std over five seeds) using an 8-layer backbone pre-trained on the \textbf{Monash-sized synthetic} corpus (matched scale, per-model configuration). Datasets: Ethanol (EthanolConcentration), FaceDet (FaceDetection), Handwrit (Handwriting), Heartbt (Heartbeat), JapVowel (JapaneseVowels), SCP1/SCP2 (SelfRegulationSCP1/2), Digits (SpokenArabicDigits), UWave (UWaveGestureLibrary). Rows are grouped into signal-space predictive and reconstructive, latent-space alignment, and contrastive objectives; the best result per dataset is in \textbf{bold}.}
\label{tab:cls_monashsize_equal}
\resizebox{\textwidth}{!}{%
\begin{tabular}{l*{9}{c}}
\toprule
\textbf{Objective} & \textbf{Ethanol} & \textbf{FaceDet} & \textbf{Handwrit} & \textbf{Heartbt} & \textbf{JapVowel} & \textbf{SCP1} & \textbf{SCP2} & \textbf{Digits} & \textbf{UWave} \\
\midrule
\textit{Random frozen encoder} & 27.68$\pm$1.88 & 54.49$\pm$0.54 & 4.31$\pm$0.54 & 66.63$\pm$1.33 & 93.02$\pm$0.65 & 76.72$\pm$2.86 & 50.00$\pm$2.60 & 77.61$\pm$0.96 & 36.56$\pm$2.78 \\
\midrule
\multicolumn{10}{l}{\textit{Signal-space predictive and reconstructive}} \\
NTP & 27.84$\pm$1.37 & \textbf{60.65$\pm$0.93} & 4.80$\pm$1.20 & 70.83$\pm$2.57 & \textbf{96.11$\pm$0.47} & 75.56$\pm$1.73 & 53.55$\pm$2.85 & 86.05$\pm$0.75 & 61.56$\pm$2.38 \\
MAE & \textbf{28.14$\pm$1.83} & 53.55$\pm$1.24 & 4.30$\pm$0.51 & \textbf{73.46$\pm$2.26} & 93.89$\pm$2.03 & \textbf{83.96$\pm$3.39} & 50.33$\pm$2.97 & 76.75$\pm$2.27 & 28.06$\pm$2.30 \\
Diffusion & 25.02$\pm$0.56 & 56.90$\pm$0.88 & 9.41$\pm$1.20 & 65.85$\pm$1.54 & 91.35$\pm$0.85 & 68.13$\pm$2.67 & 52.11$\pm$2.29 & 88.55$\pm$1.22 & 55.13$\pm$2.34 \\
\midrule
\multicolumn{10}{l}{\textit{Latent-space alignment}} \\
DINO & 27.38$\pm$0.93 & 52.87$\pm$0.67 & 11.62$\pm$1.34 & 71.51$\pm$1.26 & 75.89$\pm$0.65 & 73.72$\pm$1.43 & 52.78$\pm$0.99 & 77.97$\pm$1.64 & 36.25$\pm$5.13 \\
JEPA & 25.10$\pm$0.64 & 52.06$\pm$0.33 & 6.85$\pm$1.05 & 63.51$\pm$2.53 & 91.24$\pm$0.37 & 68.94$\pm$2.79 & \textbf{55.78$\pm$2.64} & 79.92$\pm$0.41 & 42.50$\pm$1.47 \\
LeJEPA & 26.08$\pm$2.12 & 59.67$\pm$0.61 & \textbf{16.38$\pm$0.69} & 60.68$\pm$2.73 & 81.46$\pm$2.47 & 82.12$\pm$1.38 & 46.44$\pm$1.15 & \textbf{91.11$\pm$1.58} & \textbf{71.19$\pm$2.18} \\
\midrule
\multicolumn{10}{l}{\textit{Contrastive}} \\
SoftCLT & 27.91$\pm$3.82 & 52.04$\pm$1.27 & 6.16$\pm$1.04 & 73.17$\pm$0.81 & 64.90$\pm$14.72 & 61.09$\pm$3.74 & 55.00$\pm$2.14 & 64.06$\pm$12.81 & 33.56$\pm$14.72 \\
\bottomrule
\end{tabular}%
}
\end{table}

\FloatBarrier
\subsubsection{Extended Anomaly Detection Results (Linear Probing)}
\label{subsec:appendix_ext_ad}
Tables~\ref{tab:anomaly_detection_results},\ref{tab:anomaly_mixed_compact} , \ref{tab:anomaly_synthetic_compact},
and~\ref{tab:anom_monashsize} provide the full anomaly-detection
results under Monash-sized synthetic, large synthetic, and hybrid
pre-training, respectively. Overall,
anomaly detection benefits more clearly from SSL than forecasting, but the
leading objective depends on the dataset rather than on the objective
family.

Across all four sources, JEPA, LeJEPA, and Diffusion lead the
spacecraft-telemetry datasets. All three reach 56.5--57.0 $\mathrm{F1}_{\mathrm{PA}}$ on
\textit{MSL} under every source ($\Delta{=}{+}41$ over the random frozen
encoder), and JEPA and LeJEPA reach 33.8--35.5 on \textit{SMAP}
($\Delta{=}{+}26$ to ${+}28$), 6--10 points above the next objective.
On the service-monitoring datasets DINO leads instead: it attains the
highest $\mathrm{F1}_{\mathrm{PA}}$ on \textit{PSM} under every source (63.9--72.7) and on
\textit{SMD} under every source (61.9--63.5), where NTP is within
0.3--1.6 points. JEPA, the strongest objective on \textit{MSL} and
\textit{SMAP}, falls below the random frozen encoder on \textit{PSM}
under every source (19.3--25.8 vs.\ 47.6), so leadership on one dataset
does not carry to another within the same family.

The effect of pre-training source is meaningful but not uniform. NTP and
MAE improve on \textit{PSM} under the synthetic-containing sources: NTP
rises from 55.0 under Monash to 58.1, 63.6, and 67.7 under matched
synthetic, hybrid, and large synthetic, and MAE from 36.3 to 49.5, 47.4,
and 64.0, with seed standard deviations of 5--12 points. JEPA, LeJEPA,
and Diffusion move by less than one point on \textit{MSL} and
\textit{SMAP} across sources. SoftCLT is unstable under synthetic sources
(\textit{PSM} 33.8$\pm$14.4 and \textit{SMD} 40.0$\pm$11.6 at matched
scale). \textit{SWaT} is a consistent outlier: no objective exceeds the
random frozen encoder on PA or affiliation F1 under any source, and JEPA
and LeJEPA collapse to $\approx$3 $\mathrm{F1}_{\mathrm{PA}}$ throughout.

Taken together, the objective that leads each dataset is stable across
sources: JEPA and LeJEPA on \textit{MSL} and \textit{SMAP}, DINO on
\textit{PSM} and \textit{SMD} under point-adjusted F1, and JEPA under
event-wise F1 on \textit{SMD} (31.2--31.3 under every source). Source
changes absolute scores, most for NTP and MAE on \textit{PSM}, but does
not produce a universally superior pre-training regime or change which
objective leads.

Anomaly detection benefits from pre-training on four of the five benchmarks (Table~\ref{tab:anomaly_detection_results}). On the spacecraft-telemetry datasets JEPA and LeJEPA lead: on MSL they raise affiliation F1 from 19 to 49 and point-adjusted F1 from 16 to 57 ($\Delta{=}{+}41$), matched by Diffusion, and on SMAP from 8 to 27 and 8 to 35, 6--8 points above MAE, the strongest signal-space objective; seed standard deviations are below one point. On the service-monitoring datasets NTP and DINO lead instead: DINO raises PSM affiliation F1 from 46 to 55 and point-adjusted F1 from 48 to 71, NTP and DINO gain 6 on SMD under both, while JEPA falls below the baseline on PSM and SoftCLT on PSM and SMD. Event-wise F1 confirms the benefit but changes the leader: it rises from 7 to 13 on MSL (MAE, NTP), 16 to 42 on PSM (LeJEPA, DINO), and 18 to 31 on SMD (JEPA, LeJEPA). SWaT is the exception: under point-adjusted and affiliation F1 no objective improves on the random frozen encoder, and the event-wise gain (NTP 2.9 vs.\ 1.1) is within two points.

The transfer gains persist under full fine-tuning (Tables~\ref{tab:anomaly_finetuning_results}
and~\ref{tab:classification_finetuning_results_with_baseline}), but the ranking does not. JEPA and LeJEPA lead
\textit{MSL} at $\mathrm{F1}_{\mathrm{PA}}$ 57.3 vs.\ 49.4 from scratch ($\Delta=+7.9$), Diffusion
and LeJEPA lead \textit{PSM} at 48.3 vs.\ 28.3 ($\Delta=+20.0$), and LeJEPA
reaches 85.4 on \textit{UWave} vs.\ 48.3 ($\Delta=+37$) and 25.1 on
\textit{Handwriting} vs.\ 4.5 ($\Delta=+20$), the two yielding the largest transfer gain in
the table. Yet DINO, below the random frozen encoder on \textit{FaceDetection}
under linear probing (52.0 vs.\ 54.5), leads it after fine-tuning (65.7
vs.\ 50.4), and Diffusion, below it on \textit{PSM} (43.8 vs.\ 47.6), leads
there too (48.3); conversely NTP gains 7 $\mathrm{F1}_{\mathrm{PA}}$ on \textit{PSM} frozen but
falls to 24.1 vs.\ 28.3 adapted, and DINO becomes unstable under
anomaly-detection fine-tuning, with seed variance an order of magnitude above
the other objectives (\textit{MSL} PA 31.0$\pm$6.5) and results below the
baseline on every dataset except \textit{SWaT}.

\paragraph{Sensitivity to the detection metric}
Which objective leads depends on the F1 criterion
(Table~\ref{tab:anomaly_detection_results}). On \textit{SMD}, DINO attains
the highest point-adjusted F1 (62.2) while JEPA attains the highest
event-wise F1 (31.2 vs.\ 17.5 for DINO); on \textit{PSM}, DINO leads
point-adjusted and affiliation F1 (71.2, 54.8) while LeJEPA matches it
event-wise (41.9 vs.\ 41.9); on \textit{MSL}, the point-adjusted leaders
JEPA and LeJEPA ($\approx$57) trail MAE and NTP event-wise (12.2 vs.\ 13.3
and 13.1). The same reversals hold under synthetic and hybrid pre-training
(Appendix~\ref{subsec:appendix_ext_ad}) and under full fine-tuning, where NTP
leads \textit{SMD} point-adjusted (54.2) and SoftCLT affiliation (58.1) but
JEPA, LeJEPA, and Diffusion lead event-wise (31.2, 30.6, 30.2 vs.\ 20.8 from
scratch). Event-wise and affiliation scoring preserve the benefit of
pre-training on \textit{MSL}, \textit{PSM}, \textit{SMAP}, and \textit{SMD}
but change the leading objective on each of them.

\begin{table}[ht]
\centering
\setlength{\tabcolsep}{3pt}
\caption{Linear-probing anomaly-detection F1 (\%, mean~$\pm$~std over five seeds) using an 8-layer backbone pre-trained on the \textbf{large synthetic} corpus (2.5M samples). For each dataset we report point-adjusted (PA), event-wise (Ev), and affiliation-based (Af) F1. Rows are grouped into signal-space predictive and reconstructive, latent-space alignment, and contrastive objectives; the best result per dataset and metric is in \textbf{bold}.}
\label{tab:anomaly_synthetic_compact}
\resizebox{\textwidth}{!}{%
\begin{tabular}{l*{15}{c}}
\toprule
\textbf{Objective}
& \multicolumn{3}{c}{\textbf{MSL}} & \multicolumn{3}{c}{\textbf{PSM}} & \multicolumn{3}{c}{\textbf{SMAP}} & \multicolumn{3}{c}{\textbf{SMD}} & \multicolumn{3}{c}{\textbf{SWaT}} \\
\cmidrule(lr){2-4}\cmidrule(lr){5-7}\cmidrule(lr){8-10}\cmidrule(lr){11-13}\cmidrule(lr){14-16}
& PA & Ev & Af & PA & Ev & Af & PA & Ev & Af & PA & Ev & Af & PA & Ev & Af \\
\midrule
\textit{Random frozen encoder} & 15.57$\pm$0.27 & 7.32$\pm$0.06 & 18.66$\pm$0.14 & 47.55$\pm$2.10 & 16.40$\pm$1.23 & 46.12$\pm$1.74 & 7.98$\pm$0.14 & 2.47$\pm$0.04 & 8.43$\pm$0.09 & 56.08$\pm$1.38 & 17.55$\pm$0.36 & 58.25$\pm$0.79 & \textbf{11.84$\pm$0.01} & 1.09$\pm$0.06 & \textbf{14.23$\pm$0.16} \\
\midrule
\multicolumn{16}{l}{\textit{Signal-space predictive and reconstructive}} \\
NTP & 24.66$\pm$0.51 & 11.54$\pm$0.17 & 27.10$\pm$0.33 & 67.66$\pm$1.17 & 41.39$\pm$3.77 & 51.03$\pm$2.36 & 17.42$\pm$0.77 & 4.73$\pm$0.14 & 14.53$\pm$0.53 & 61.84$\pm$0.66 & 17.25$\pm$0.68 & 64.17$\pm$0.60 & 11.70$\pm$0.10 & 1.59$\pm$0.52 & 12.14$\pm$2.64 \\
MAE & 29.32$\pm$0.27 & \textbf{12.95$\pm$0.04} & 29.95$\pm$0.20 & 63.95$\pm$4.78 & 32.87$\pm$4.45 & 54.44$\pm$1.98 & 25.90$\pm$0.71 & 6.44$\pm$0.15 & 20.80$\pm$0.47 & 57.96$\pm$1.84 & 17.02$\pm$0.79 & 59.90$\pm$1.31 & 11.70$\pm$0.10 & 1.66$\pm$0.30 & 12.89$\pm$1.83 \\
Diffusion & 56.66$\pm$0.01 & 12.27$\pm$0.01 & 48.50$\pm$0.02 & 43.78$\pm$2.74 & 40.73$\pm$1.27 & 40.32$\pm$2.66 & \textbf{36.37$\pm$0.43} & 6.76$\pm$0.05 & \textbf{28.19$\pm$0.35} & 52.42$\pm$0.27 & 30.24$\pm$0.31 & 55.93$\pm$0.14 & 7.19$\pm$1.10 & 0.86$\pm$0.16 & 8.32$\pm$1.06 \\
\midrule
\multicolumn{16}{l}{\textit{Latent-space alignment}} \\
DINO & 25.51$\pm$1.61 & 12.03$\pm$0.54 & 27.15$\pm$1.01 & \textbf{72.66$\pm$2.25} & 40.52$\pm$4.92 & \textbf{58.06$\pm$2.18} & 18.72$\pm$2.27 & 5.50$\pm$0.73 & 15.34$\pm$1.54 & \textbf{63.48$\pm$1.32} & 16.71$\pm$0.40 & \textbf{65.38$\pm$0.68} & 11.74$\pm$0.06 & 1.45$\pm$0.11 & 12.97$\pm$1.46 \\
JEPA & 56.95$\pm$0.09 & 12.20$\pm$0.04 & 49.25$\pm$0.10 & 19.33$\pm$0.84 & 25.96$\pm$0.61 & 22.76$\pm$0.27 & 33.76$\pm$0.42 & 6.29$\pm$0.09 & 25.74$\pm$0.50 & 52.65$\pm$0.13 & \textbf{31.25$\pm$0.05} & 56.07$\pm$0.09 & 3.21$\pm$0.00 & \textbf{1.90$\pm$0.10} & 4.12$\pm$0.02 \\
LeJEPA & \textbf{57.03$\pm$0.07} & 12.21$\pm$0.03 & \textbf{49.26$\pm$0.05} & 47.15$\pm$0.23 & \textbf{41.61$\pm$0.42} & 42.69$\pm$0.13 & 34.10$\pm$0.53 & 6.35$\pm$0.12 & 27.06$\pm$0.29 & 52.55$\pm$0.16 & 30.95$\pm$0.04 & 55.98$\pm$0.09 & 3.10$\pm$0.07 & 0.94$\pm$0.06 & 4.11$\pm$0.07 \\
\midrule
\multicolumn{16}{l}{\textit{Contrastive}} \\
SoftCLT & 25.99$\pm$4.82 & 12.08$\pm$1.66 & 26.91$\pm$4.26 & 37.12$\pm$13.89 & 31.65$\pm$7.94 & 36.76$\pm$10.46 & 28.16$\pm$1.21 & \textbf{7.06$\pm$0.10} & 22.37$\pm$1.00 & 45.81$\pm$5.15 & 22.18$\pm$6.04 & 48.98$\pm$4.59 & 10.01$\pm$1.66 & 1.39$\pm$0.38 & 12.02$\pm$1.87 \\
\bottomrule
\end{tabular}%
}
\end{table}

\begin{table}[ht]
\centering
\setlength{\tabcolsep}{3pt}
\caption{Linear-probing anomaly-detection F1 (\%, mean~$\pm$~std over five seeds) using an 8-layer backbone pre-trained on the \textbf{hybrid} mixture of Monash and synthetic data (2.5M samples). For each dataset we report point-adjusted (PA), event-wise (Ev), and affiliation-based (Af) F1. Rows are grouped into signal-space predictive and reconstructive, latent-space alignment, and contrastive objectives; the best result per dataset and metric is in \textbf{bold}}
\label{tab:anomaly_mixed_compact}
\resizebox{\textwidth}{!}{%
\begin{tabular}{l*{15}{c}}
\toprule
\textbf{Objective}
& \multicolumn{3}{c}{\textbf{MSL}} & \multicolumn{3}{c}{\textbf{PSM}} & \multicolumn{3}{c}{\textbf{SMAP}} & \multicolumn{3}{c}{\textbf{SMD}} & \multicolumn{3}{c}{\textbf{SWaT}} \\
\cmidrule(lr){2-4}\cmidrule(lr){5-7}\cmidrule(lr){8-10}\cmidrule(lr){11-13}\cmidrule(lr){14-16}
& PA & Ev & Af & PA & Ev & Af & PA & Ev & Af & PA & Ev & Af & PA & Ev & Af \\
\midrule
\textit{Random frozen encoder} & 15.57$\pm$0.27 & 7.32$\pm$0.06 & 18.66$\pm$0.14 & 47.55$\pm$2.10 & 16.40$\pm$1.23 & 46.12$\pm$1.74 & 7.98$\pm$0.14 & 2.47$\pm$0.04 & 8.43$\pm$0.09 & 56.08$\pm$1.38 & 17.55$\pm$0.36 & 58.25$\pm$0.79 & \textbf{11.84$\pm$0.01} & 1.09$\pm$0.06 & \textbf{14.23$\pm$0.16} \\
\midrule
\multicolumn{16}{l}{\textit{Signal-space predictive and reconstructive}} \\
NTP & 26.34$\pm$0.43 & 12.30$\pm$0.10 & 28.03$\pm$0.31 & 63.57$\pm$2.92 & 31.96$\pm$2.27 & 57.45$\pm$1.90 & 19.29$\pm$0.42 & 5.26$\pm$0.11 & 15.87$\pm$0.32 & 61.33$\pm$0.91 & 18.82$\pm$0.86 & 63.71$\pm$0.32 & 11.67$\pm$0.12 & 1.72$\pm$0.35 & 12.01$\pm$2.64 \\
MAE & 29.16$\pm$0.35 & 12.81$\pm$0.25 & 29.76$\pm$0.29 & 47.44$\pm$11.65 & 24.78$\pm$4.10 & 47.09$\pm$6.98 & 27.32$\pm$2.63 & 6.83$\pm$0.49 & 21.53$\pm$1.83 & 61.62$\pm$1.25 & 17.71$\pm$0.62 & 63.93$\pm$1.01 & 11.51$\pm$0.15 & 1.72$\pm$0.31 & 10.55$\pm$2.19 \\
Diffusion & 56.66$\pm$0.00 & 12.26$\pm$0.01 & 48.55$\pm$0.08 & 44.64$\pm$2.25 & 26.41$\pm$3.09 & 43.03$\pm$2.74 & 27.48$\pm$2.17 & 5.85$\pm$0.19 & 22.64$\pm$1.58 & 52.27$\pm$0.27 & 29.85$\pm$0.28 & 55.94$\pm$0.20 & 9.70$\pm$2.40 & 1.20$\pm$0.36 & 11.55$\pm$2.67 \\
\midrule
\multicolumn{16}{l}{\textit{Latent-space alignment}} \\
DINO & 23.52$\pm$0.74 & 12.41$\pm$0.56 & 25.83$\pm$0.52 & \textbf{72.56$\pm$1.06} & \textbf{42.51$\pm$4.53} & \textbf{58.88$\pm$2.57} & 18.94$\pm$1.23 & 5.61$\pm$0.51 & 15.52$\pm$0.83 & \textbf{61.94$\pm$0.51} & 16.53$\pm$0.34 & \textbf{64.07$\pm$0.39} & 11.68$\pm$0.13 & 1.67$\pm$0.10 & 11.56$\pm$1.81 \\
JEPA & \textbf{56.99$\pm$0.05} & 12.19$\pm$0.04 & \textbf{49.24$\pm$0.04} & 19.96$\pm$2.25 & 26.42$\pm$1.56 & 23.07$\pm$0.99 & 33.96$\pm$0.36 & 6.33$\pm$0.09 & 25.92$\pm$0.60 & 52.56$\pm$0.16 & \textbf{31.16$\pm$0.12} & 55.98$\pm$0.11 & 3.20$\pm$0.04 & \textbf{1.80$\pm$0.14} & 4.13$\pm$0.03 \\
LeJEPA & 56.97$\pm$0.05 & 12.17$\pm$0.02 & 49.21$\pm$0.07 & 46.52$\pm$0.96 & 42.42$\pm$0.38 & 41.88$\pm$1.11 & \textbf{34.50$\pm$0.46} & 6.42$\pm$0.10 & \textbf{27.11$\pm$0.38} & 52.69$\pm$0.03 & 31.07$\pm$0.09 & 56.02$\pm$0.07 & 3.06$\pm$0.07 & 1.16$\pm$0.18 & 4.06$\pm$0.06 \\
\midrule
\multicolumn{16}{l}{\textit{Contrastive}} \\
SoftCLT & 25.56$\pm$6.11 & \textbf{13.00$\pm$0.18} & 26.54$\pm$5.27 & 50.34$\pm$1.48 & 38.26$\pm$2.09 & 46.18$\pm$2.33 & 30.56$\pm$0.34 & \textbf{7.45$\pm$0.19} & 24.03$\pm$0.26 & 49.17$\pm$3.40 & 27.48$\pm$0.73 & 52.46$\pm$4.04 & 11.13$\pm$0.52 & 1.06$\pm$0.14 & 13.11$\pm$0.89 \\
\bottomrule
\end{tabular}%
}
\end{table}

\begin{table}[ht]
\centering
\setlength{\tabcolsep}{3pt}
\caption{Linear-probing anomaly-detection F1 (\%, mean~$\pm$~std over five seeds) using an 8-layer backbone pre-trained on the \textbf{Monash-sized synthetic} corpus (matched scale, per-model configuration). For each dataset we report point-adjusted (PA), event-wise (Ev), and affiliation-based (Af) F1. Rows are grouped into signal-space predictive and reconstructive, latent-space alignment, and contrastive objectives; the best result per dataset and metric is in \textbf{bold}.}
\label{tab:anom_monashsize}
\resizebox{\textwidth}{!}{%
\begin{tabular}{l*{15}{c}}
\toprule
\textbf{Objective}
& \multicolumn{3}{c}{\textbf{MSL}} & \multicolumn{3}{c}{\textbf{PSM}} & \multicolumn{3}{c}{\textbf{SMAP}} & \multicolumn{3}{c}{\textbf{SMD}} & \multicolumn{3}{c}{\textbf{SWaT}} \\
\cmidrule(lr){2-4}\cmidrule(lr){5-7}\cmidrule(lr){8-10}\cmidrule(lr){11-13}\cmidrule(lr){14-16}
& PA & Ev & Af & PA & Ev & Af & PA & Ev & Af & PA & Ev & Af & PA & Ev & Af \\
\midrule
\textit{Random frozen encoder} & 15.57$\pm$0.27 & 7.32$\pm$0.06 & 18.66$\pm$0.14 & 47.55$\pm$2.10 & 16.40$\pm$1.23 & 46.12$\pm$1.74 & 7.98$\pm$0.14 & 2.47$\pm$0.04 & 8.43$\pm$0.09 & 56.08$\pm$1.38 & 17.55$\pm$0.36 & 58.25$\pm$0.79 & \textbf{11.84$\pm$0.01} & 1.09$\pm$0.06 & \textbf{14.23$\pm$0.16} \\
\midrule
\multicolumn{16}{l}{\textit{Signal-space predictive and reconstructive}} \\
NTP & 25.01$\pm$0.61 & 11.72$\pm$0.28 & 27.26$\pm$0.37 & 58.10$\pm$8.56 & 29.27$\pm$7.40 & 51.90$\pm$4.65 & 17.07$\pm$0.82 & 4.68$\pm$0.24 & 14.30$\pm$0.62 & 61.53$\pm$1.59 & 17.33$\pm$0.82 & 63.70$\pm$0.93 & 11.81$\pm$0.02 & \textbf{1.71$\pm$0.23} & 13.32$\pm$1.25 \\
MAE & 28.98$\pm$0.72 & 12.83$\pm$0.24 & 29.72$\pm$0.42 & 49.52$\pm$11.79 & 23.64$\pm$2.97 & 47.03$\pm$3.98 & 24.45$\pm$1.76 & 6.24$\pm$0.34 & 19.63$\pm$1.27 & 59.73$\pm$1.40 & 17.53$\pm$0.60 & 61.46$\pm$1.38 & 9.70$\pm$3.86 & 1.36$\pm$0.48 & 11.18$\pm$2.15 \\
Diffusion & 56.66$\pm$0.00 & 12.23$\pm$0.03 & 48.59$\pm$0.06 & 43.41$\pm$2.83 & 34.18$\pm$4.56 & 41.17$\pm$2.88 & 30.84$\pm$0.70 & 6.30$\pm$0.07 & 24.94$\pm$0.56 & 52.33$\pm$0.52 & 29.81$\pm$0.35 & 55.91$\pm$0.31 & 7.35$\pm$0.25 & 0.93$\pm$0.15 & 8.93$\pm$0.45 \\
\midrule
\multicolumn{16}{l}{\textit{Latent-space alignment}} \\
DINO & 25.56$\pm$0.80 & 11.96$\pm$0.34 & 27.39$\pm$0.39 & \textbf{63.91$\pm$7.12} & 33.67$\pm$2.49 & \textbf{55.15$\pm$5.22} & 18.60$\pm$1.55 & 5.39$\pm$0.51 & 15.31$\pm$1.10 & \textbf{63.06$\pm$0.51} & 16.24$\pm$0.17 & \textbf{65.39$\pm$0.45} & 11.75$\pm$0.09 & 1.36$\pm$0.20 & 13.20$\pm$1.64 \\
JEPA & 56.94$\pm$0.07 & 12.16$\pm$0.03 & 49.13$\pm$0.06 & 24.60$\pm$3.98 & 29.40$\pm$2.78 & 25.68$\pm$2.26 & \textbf{35.50$\pm$0.63} & 6.70$\pm$0.14 & \textbf{27.87$\pm$0.50} & 52.63$\pm$0.12 & \textbf{31.19$\pm$0.05} & 56.03$\pm$0.03 & 3.21$\pm$0.01 & 1.61$\pm$0.16 & 4.16$\pm$0.02 \\
LeJEPA & \textbf{56.97$\pm$0.04} & 12.20$\pm$0.02 & \textbf{49.22$\pm$0.07} & 46.87$\pm$0.23 & \textbf{41.52$\pm$0.59} & 42.46$\pm$0.30 & 34.47$\pm$0.56 & 6.43$\pm$0.12 & 27.15$\pm$0.31 & 52.49$\pm$0.14 & 30.93$\pm$0.09 & 55.94$\pm$0.03 & 3.07$\pm$0.09 & 0.91$\pm$0.06 & 4.09$\pm$0.07 \\
\midrule
\multicolumn{16}{l}{\textit{Contrastive}} \\
SoftCLT & 30.37$\pm$0.07 & \textbf{13.20$\pm$0.04} & 30.80$\pm$0.04 & 40.20$\pm$1.29 & 35.59$\pm$0.84 & 37.81$\pm$0.76 & 29.66$\pm$0.41 & \textbf{7.29$\pm$0.06} & 23.50$\pm$0.33 & 52.47$\pm$0.34 & 30.49$\pm$0.20 & 56.39$\pm$0.12 & 11.22$\pm$0.37 & 1.42$\pm$0.44 & 13.24$\pm$0.46 \\
\bottomrule
\end{tabular}%
}
\end{table}

\FloatBarrier

\begin{figure}[t]
\centering
\pgfplotsset{
    unified bar plot/.style={
        ybar,
        bar width=2.4pt,
        width=1.05\linewidth,
        height=3.4cm,
        xtick=data,
        xticklabel style={font=\fontsize{5}{6}\selectfont},
        grid=major,
        grid style={dashed, gray!30},
        enlarge x limits=0.14,
        ylabel style={font=\fontsize{6}{7}\selectfont, xshift=-0.8em},
        yticklabel style={font=\tiny},
        symbolic x coords={NTP, MAE, Diffusion, DINO, JEPA, LeJEPA, SoftCLT},
    },
    baseline line/.style={dashed, black, thick},
}
\newcommand{\baselineline}[1]{%
    \draw[baseline line] ({rel axis cs:0,0}|-{axis cs:NTP,#1}) -- ({rel axis cs:1,0}|-{axis cs:NTP,#1});%
}

\begin{subfigure}[b]{0.48\textwidth}
\centering
\begin{tikzpicture}
    \begin{axis}[
        unified bar plot,
        ylabel={Mean $\mathrm{F1}_{\mathrm{PA}}$},
        ymin=0, ymax=0.5,
        ytick={0, 0.1, 0.2, 0.3, 0.4, 0.5},
        legend to name=fig:data_composition_all:legend,
        legend columns=-1,
        legend style={font=\fontsize{6}{7}\selectfont, draw=none, /tikz/every even column/.append style={column sep=0.5em}},
    ]
        \addplot[fill=gray!20,   draw=black] coordinates {(NTP,0.359) (MAE,0.323) (Diffusion,0.372) (DINO,0.381) (JEPA,0.347) (LeJEPA,0.389) (SoftCLT,0.336)};
        \addplot[fill=blue!25,   draw=black] coordinates {(NTP,0.347) (MAE,0.345) (Diffusion,0.381) (DINO,0.366) (JEPA,0.346) (LeJEPA,0.388) (SoftCLT,0.328)};
        \addplot[fill=orange!40, draw=black] coordinates {(NTP,0.367) (MAE,0.378) (Diffusion,0.393) (DINO,0.384) (JEPA,0.332) (LeJEPA,0.388) (SoftCLT,0.294)};
        \addplot[fill=teal!40,   draw=black] coordinates {(NTP,0.364) (MAE,0.354) (Diffusion,0.382) (DINO,0.377) (JEPA,0.333) (LeJEPA,0.387) (SoftCLT,0.334)};
        \legend{Monash, Small Synthetic, Large Synthetic, Hybrid}
    \end{axis}
\end{tikzpicture}
\subcaption{Anomaly detection}
\end{subfigure}
\hfill
\begin{subfigure}[b]{0.48\textwidth}
\centering
\begin{tikzpicture}
    \begin{axis}[
        unified bar plot,
        ylabel={Mean accuracy},
        ymin=0, ymax=0.8,
        ytick={0, 0.2, 0.4, 0.6, 0.8},
    ]
        \addplot[fill=gray!20,   draw=black] coordinates {(NTP,0.574) (MAE,0.528) (Diffusion,0.553) (DINO,0.518) (JEPA,0.523) (LeJEPA,0.609) (SoftCLT,0.528)};
        \addplot[fill=blue!25,   draw=black] coordinates {(NTP,0.597) (MAE,0.547) (Diffusion,0.569) (DINO,0.533) (JEPA,0.540) (LeJEPA,0.595) (SoftCLT,0.487)};
        \addplot[fill=orange!40, draw=black] coordinates {(NTP,0.596) (MAE,0.538) (Diffusion,0.560) (DINO,0.551) (JEPA,0.504) (LeJEPA,0.591) (SoftCLT,0.527)};
        \addplot[fill=teal!40,   draw=black] coordinates {(NTP,0.585) (MAE,0.554) (Diffusion,0.557) (DINO,0.550) (JEPA,0.520) (LeJEPA,0.586) (SoftCLT,0.570)};
    \end{axis}
\end{tikzpicture}
\subcaption{Classification}
\end{subfigure}

\vspace{0.05cm}
\ref{fig:data_composition_all:legend}

\caption{Impact of pre-training corpus under linear probing (8-layer backbone). Bars: Monash (680K), Small Synthetic (680K, matched scale), Large Synthetic (2.5M), and Hybrid (2.5M). (a)~$\mathrm{F1}_{\mathrm{PA}}$ averaged over the five anomaly-detection datasets; higher is better. (b)~Accuracy averaged over the nine UEA datasets; higher is better. The corresponding forecasting results are shown per dataset in Figures~\ref{fig:forecasting_data_regime_etth}--\ref{fig:forecasting_data_regime_traffic}.}
\label{fig:data_composition_all}
\end{figure}

\FloatBarrier

\section{Training Duration, Depth, and Convergence}
\label{sec:DDc}
In this section we provide results on training duration, transformer depth, and convergence.

\subsection{Extended Results for 40 Epoch Evaluation}
\label{sec:40_ep}
To test whether the per-model results are limited by pre-training budget, we double the schedule from 20 to 40 epochs while holding every other hyper-parameter at its native per-model value: the same 8-layer encoder, 336-step context, patch length, optimiser, and learning rate, so the only change is twice as many parameter updates. All backbones are pre-trained on Monash and evaluated by linear probing over 5 seeds.
We run the same experiment with a 24-layered encoder on single seed. Tables~\ref{tab:forecasting_l8_monash_ep40}, ~\ref{tab:cls_monash_ep40}, \ref{tab:anom_monash_ep40}, and ~\ref{tab:classification_depth24} report classification, anomaly detection, forecasting, and 24 layer classification.
We also report results of 24 layered transformer with 40 epochs

\paragraph{Results.}
Doubling the pre-training budget leaves the 20-epoch conclusions intact. Rankings are unchanged across all three task families: LeJEPA still leads the gesture and morphology benchmarks, signal-space objectives the periodic ones, JEPA and LeJEPA the spacecraft-telemetry anomaly datasets, and the Random frozen encoder baseline still attains the lowest forecasting MSE on ETTh1, Electricity, and Traffic. Most cells move by less than one accuracy point or $0.005$ MSE. SoftCLT is the exception. SoftCLT improves forecasting on every dataset (ETTh1 $0.456 \rightarrow 0.442$, Electricity $0.217 \rightarrow 0.195$, Traffic $0.488 \rightarrow 0.431$), but remains above the random encoder on six of seven, matching it only on Traffic ($0.431$ vs.\ $0.435$). The additional updates otherwise amplify the failure modes already visible at 20 epochs: JEPA collapses on JapaneseVowels, SpokenArabicDigits, and UWave with seed standard deviations of $8$--$22$ points, and LeJEPA's forecasting degradation deepens on every dataset (Traffic $0.848 \to 0.940$), consistent with the isotropy induced by SIGReg accumulating with training. The forecasting transfer gain does not emerge with more training, so the gap to the random encoder is not a symptom of under-training, in line with the convergence analysis in Appendix~\ref{subsec:appendix_depth_scaling}.
\begin{table}[ht]
\centering
\setlength{\tabcolsep}{3pt}
\caption{Linear-probing forecasting error (MSE and MAE, mean~$\pm$~std over five seeds, averaged over horizons $\{96,192,336,720\}$) using an 8-layer backbone pre-trained on the \textbf{Monash} dataset for \textbf{40 epochs}. Lower is better; the best result per dataset and metric is in \textbf{bold}.}
\label{tab:forecasting_l8_monash_ep40}
\resizebox{\textwidth}{!}{%
\begin{tabular}{l*{14}{c}}
\toprule
\textbf{Objective} & \multicolumn{2}{c}{\textbf{ETTh1}} & \multicolumn{2}{c}{\textbf{ETTh2}} & \multicolumn{2}{c}{\textbf{ETTm1}} & \multicolumn{2}{c}{\textbf{ETTm2}} & \multicolumn{2}{c}{\textbf{Weather}} & \multicolumn{2}{c}{\textbf{Electricity}} & \multicolumn{2}{c}{\textbf{Traffic}} \\
\cmidrule(lr){2-3}\cmidrule(lr){4-5}\cmidrule(lr){6-7}\cmidrule(lr){8-9}\cmidrule(lr){10-11}\cmidrule(lr){12-13}\cmidrule(lr){14-15}
 & MSE & MAE & MSE & MAE & MSE & MAE & MSE & MAE & MSE & MAE & MSE & MAE & MSE & MAE \\
\midrule
\textit{Random frozen encoder} & \textbf{0.414$\pm$0.002} & \textbf{0.426$\pm$0.002} & 0.350$\pm$0.002 & 0.393$\pm$0.001 & 0.356$\pm$0.001 & \textbf{0.379$\pm$0.001} & \textbf{0.261$\pm$0.001} & 0.318$\pm$0.000 & 0.248$\pm$0.000 & 0.281$\pm$0.001 & \textbf{0.172$\pm$0.000} & \textbf{0.267$\pm$0.000} & 0.435$\pm$0.000 & 0.298$\pm$0.000 \\
\midrule
\multicolumn{15}{l}{\textit{Signal-space predictive and reconstructive}} \\
NTP & 0.436$\pm$0.002 & 0.441$\pm$0.001 & 0.362$\pm$0.001 & 0.400$\pm$0.001 & \textbf{0.351$\pm$0.000} & 0.381$\pm$0.000 & 0.262$\pm$0.001 & 0.319$\pm$0.001 & \textbf{0.235$\pm$0.000} & \textbf{0.270$\pm$0.001} & 0.179$\pm$0.000 & 0.277$\pm$0.001 & 0.444$\pm$0.001 & 0.311$\pm$0.001 \\
MAE & 0.451$\pm$0.004 & 0.443$\pm$0.002 & 0.362$\pm$0.002 & 0.397$\pm$0.003 & 0.366$\pm$0.001 & 0.392$\pm$0.000 & 0.263$\pm$0.002 & 0.324$\pm$0.002 & 0.244$\pm$0.002 & 0.280$\pm$0.002 & 0.192$\pm$0.001 & 0.294$\pm$0.002 & 0.474$\pm$0.004 & 0.342$\pm$0.006 \\
Diffusion & 0.436$\pm$0.004 & 0.448$\pm$0.004 & 0.334$\pm$0.001 & 0.385$\pm$0.002 & 0.379$\pm$0.004 & 0.397$\pm$0.003 & 0.264$\pm$0.001 & 0.321$\pm$0.001 & 0.243$\pm$0.002 & 0.280$\pm$0.002 & 0.235$\pm$0.015 & 0.350$\pm$0.016 & 0.550$\pm$0.036 & 0.404$\pm$0.024 \\
\midrule
\multicolumn{15}{l}{\textit{Latent-space alignment}} \\
DINO & 0.445$\pm$0.005 & 0.448$\pm$0.004 & 0.367$\pm$0.001 & 0.401$\pm$0.001 & 0.356$\pm$0.001 & 0.386$\pm$0.001 & 0.263$\pm$0.004 & \textbf{0.318$\pm$0.003} & 0.235$\pm$0.000 & 0.271$\pm$0.000 & 0.183$\pm$0.001 & 0.283$\pm$0.001 & 0.451$\pm$0.002 & 0.326$\pm$0.002 \\
JEPA & 0.437$\pm$0.005 & 0.442$\pm$0.003 & 0.361$\pm$0.004 & 0.404$\pm$0.005 & 0.368$\pm$0.005 & 0.389$\pm$0.004 & 0.273$\pm$0.006 & 0.328$\pm$0.005 & 0.253$\pm$0.009 & 0.286$\pm$0.009 & 0.189$\pm$0.009 & 0.287$\pm$0.010 & 0.457$\pm$0.018 & 0.323$\pm$0.013 \\
LeJEPA & 0.654$\pm$0.006 & 0.555$\pm$0.003 & 0.397$\pm$0.002 & 0.428$\pm$0.002 & 0.480$\pm$0.007 & 0.458$\pm$0.004 & 0.295$\pm$0.001 & 0.353$\pm$0.001 & 0.267$\pm$0.001 & 0.302$\pm$0.001 & 0.402$\pm$0.010 & 0.473$\pm$0.008 & 0.957$\pm$0.030 & 0.601$\pm$0.014 \\
\midrule
\multicolumn{15}{l}{\textit{Contrastive}} \\
SoftCLT & 0.442$\pm$0.014 & 0.448$\pm$0.012 & 0.372$\pm$0.013 & 0.408$\pm$0.009 & 0.377$\pm$0.012 & 0.400$\pm$0.012 & 0.263$\pm$0.007 & 0.322$\pm$0.008 & 0.244$\pm$0.001 & 0.279$\pm$0.001 & 0.195$\pm$0.025 & 0.292$\pm$0.028 & \textbf{0.431$\pm$0.000} & \textbf{0.293$\pm$0.000} \\
\bottomrule
\end{tabular}%
}
\end{table}

\begin{table}[ht]
\centering
\setlength{\tabcolsep}{3pt}
\caption{Linear-probing classification accuracy (\%, mean~$\pm$~std over five seeds) using an 8-layer backbone pre-trained on \textbf{Monash} for \textbf{40 epochs}. Datasets: Ethanol (EthanolConcentration), FaceDet (FaceDetection), Handwrit (Handwriting), Heartbt (Heartbeat), JapVowel (JapaneseVowels), SCP1/SCP2 (SelfRegulationSCP1/2), Digits (SpokenArabicDigits), UWave (UWaveGestureLibrary). Rows are grouped into signal-space predictive and reconstructive, latent-space alignment, and contrastive objectives; the best result per dataset is in \textbf{bold}.}
\label{tab:cls_monash_ep40}
\resizebox{\textwidth}{!}{%
\begin{tabular}{l*{9}{c}}
\toprule
\textbf{Objective} & \textbf{Ethanol} & \textbf{FaceDet} & \textbf{Handwrit} & \textbf{Heartbt} & \textbf{JapVowel} & \textbf{SCP1} & \textbf{SCP2} & \textbf{Digits} & \textbf{UWave} \\
\midrule
\textit{Random frozen encoder} & 27.68$\pm$1.88 & 54.49$\pm$0.54 & 4.31$\pm$0.54 & 66.63$\pm$1.33 & 93.02$\pm$0.65 & 76.72$\pm$2.86 & 50.00$\pm$2.60 & 77.61$\pm$0.96 & 36.56$\pm$2.78 \\
\midrule
\multicolumn{10}{l}{\textit{Signal-space predictive and reconstructive}} \\
NTP & 27.20$\pm$1.50 & 54.90$\pm$0.30 & 6.20$\pm$1.40 & 70.80$\pm$1.80 & \textbf{93.50$\pm$0.90} & 76.10$\pm$1.30 & 53.70$\pm$2.10 & 86.50$\pm$1.20 & 57.60$\pm$2.60 \\
MAE & 25.70$\pm$1.59 & 52.82$\pm$0.93 & 5.98$\pm$1.12 & \textbf{74.44$\pm$0.79} & 92.00$\pm$1.06 & \textbf{83.89$\pm$2.89} & 51.56$\pm$2.80 & 73.26$\pm$2.27 & 26.00$\pm$6.48 \\
Diffusion & 25.20$\pm$1.30 & 57.70$\pm$0.50 & 9.50$\pm$0.60 & 64.10$\pm$2.00 & 90.60$\pm$1.40 & 67.50$\pm$1.70 & 51.30$\pm$1.70 & 88.50$\pm$0.30 & 56.70$\pm$1.60 \\
\midrule
\multicolumn{10}{l}{\textit{Latent-space alignment}} \\
DINO & 26.60$\pm$1.40 & 53.40$\pm$1.50 & 11.60$\pm$1.20 & 72.40$\pm$0.40 & 76.70$\pm$1.40 & 68.70$\pm$1.10 & 53.80$\pm$1.50 & 75.70$\pm$2.00 & 41.20$\pm$7.50 \\
JEPA & 24.90$\pm$0.20 & 50.10$\pm$0.50 & 3.80$\pm$0.00 & 66.80$\pm$1.70 & 38.90$\pm$22.00 & 61.20$\pm$4.10 & \textbf{54.60$\pm$2.50} & 52.90$\pm$13.60 & 20.10$\pm$7.90 \\
LeJEPA & 25.02$\pm$1.51 & \textbf{60.60$\pm$0.87} & \textbf{15.83$\pm$1.72} & 62.05$\pm$1.76 & 83.73$\pm$1.59 & 81.77$\pm$2.10 & 48.00$\pm$2.07 & \textbf{91.87$\pm$0.98} & \textbf{76.00$\pm$2.56} \\
\midrule
\multicolumn{10}{l}{\textit{Contrastive}} \\
SoftCLT & \textbf{28.20$\pm$1.30} & 52.70$\pm$1.00 & 7.50$\pm$1.70 & 72.70$\pm$0.80 & 69.50$\pm$7.00 & 64.40$\pm$3.80 & 52.60$\pm$1.90 & 58.20$\pm$12.30 & 41.90$\pm$14.80 \\
\bottomrule
\end{tabular}%
}
\end{table}

\begin{table}[ht]
\centering
\setlength{\tabcolsep}{3pt}
\caption{Linear-probing anomaly-detection F1 (\%, mean~$\pm$~std over five seeds) using an 8-layer backbone pre-trained on \textbf{Monash} for \textbf{40 epochs}. For each dataset we report point-adjusted (PA), event-wise (Ev), and affiliation-based (Af) F1. Rows are grouped into signal-space predictive and reconstructive, latent-space alignment, and contrastive objectives; the best result per dataset and metric is in \textbf{bold}.}
\label{tab:anom_monash_ep40}
\resizebox{\textwidth}{!}{%
\begin{tabular}{l*{15}{c}}
\toprule
\textbf{Objective}
& \multicolumn{3}{c}{\textbf{MSL}} & \multicolumn{3}{c}{\textbf{PSM}} & \multicolumn{3}{c}{\textbf{SMAP}} & \multicolumn{3}{c}{\textbf{SMD}} & \multicolumn{3}{c}{\textbf{SWaT}} \\
\cmidrule(lr){2-4}\cmidrule(lr){5-7}\cmidrule(lr){8-10}\cmidrule(lr){11-13}\cmidrule(lr){14-16}
& PA & Ev & Af & PA & Ev & Af & PA & Ev & Af & PA & Ev & Af & PA & Ev & Af \\
\midrule
\textit{Random frozen encoder} & 15.57$\pm$0.27 & 7.32$\pm$0.06 & 18.66$\pm$0.14 & 47.55$\pm$2.10 & 16.40$\pm$1.23 & 46.12$\pm$1.74 & 7.98$\pm$0.14 & 2.47$\pm$0.04 & 8.43$\pm$0.09 & 56.08$\pm$1.38 & 17.55$\pm$0.36 & 58.25$\pm$0.79 & \textbf{11.84$\pm$0.01} & 1.09$\pm$0.06 & \textbf{14.23$\pm$0.16} \\
\midrule
\multicolumn{16}{l}{\textit{Signal-space predictive and reconstructive}} \\
NTP & 28.97$\pm$0.53 & 13.08$\pm$0.16 & 29.71$\pm$0.30 & 52.94$\pm$7.71 & 31.48$\pm$9.15 & 48.71$\pm$1.80 & 22.78$\pm$0.48 & 6.20$\pm$0.09 & 18.48$\pm$0.32 & \textbf{61.97$\pm$0.34} & 20.02$\pm$0.22 & \textbf{64.58$\pm$0.41} & 11.55$\pm$0.11 & \textbf{2.57$\pm$0.36} & 10.40$\pm$1.72 \\
MAE & 29.36$\pm$0.10 & \textbf{13.32$\pm$0.09} & 29.93$\pm$0.05 & 30.11$\pm$3.76 & 15.92$\pm$2.40 & 38.77$\pm$1.88 & 27.38$\pm$0.94 & \textbf{7.25$\pm$0.08} & 21.55$\pm$0.70 & 59.85$\pm$1.71 & 19.55$\pm$0.63 & 62.75$\pm$1.51 & 3.85$\pm$3.94 & 0.82$\pm$0.27 & 9.69$\pm$2.30 \\
Diffusion & 56.67$\pm$0.01 & 12.24$\pm$0.02 & 48.62$\pm$0.06 & 45.51$\pm$1.16 & 23.52$\pm$2.35 & 43.82$\pm$1.59 & 25.39$\pm$1.54 & 6.06$\pm$0.41 & 20.67$\pm$0.84 & 52.33$\pm$0.23 & 29.83$\pm$0.19 & 56.19$\pm$0.32 & 11.18$\pm$1.08 & 1.71$\pm$0.36 & 13.15$\pm$1.11 \\
\midrule
\multicolumn{16}{l}{\textit{Latent-space alignment}} \\
DINO & 25.10$\pm$0.99 & 12.31$\pm$0.50 & 27.24$\pm$0.65 & \textbf{66.04$\pm$8.28} & 35.16$\pm$7.38 & \textbf{54.59$\pm$4.01} & 21.04$\pm$2.74 & 5.82$\pm$0.82 & 17.11$\pm$1.98 & 60.65$\pm$4.61 & 19.02$\pm$3.81 & 62.49$\pm$4.04 & 11.71$\pm$0.09 & 1.70$\pm$0.46 & 12.30$\pm$2.14 \\
JEPA & \textbf{57.01$\pm$0.07} & 12.18$\pm$0.05 & \textbf{49.20$\pm$0.03} & 21.98$\pm$2.64 & 27.71$\pm$1.73 & 24.24$\pm$1.38 & 33.55$\pm$0.28 & 6.26$\pm$0.05 & 25.67$\pm$0.19 & 52.63$\pm$0.13 & \textbf{31.22$\pm$0.04} & 56.03$\pm$0.05 & 3.22$\pm$0.00 & 1.99$\pm$0.16 & 4.13$\pm$0.02 \\
LeJEPA & 56.96$\pm$0.08 & 12.18$\pm$0.03 & 49.20$\pm$0.06 & 47.06$\pm$1.11 & \textbf{42.15$\pm$0.53} & 42.02$\pm$0.96 & \textbf{34.19$\pm$1.36} & 6.35$\pm$0.28 & \textbf{26.97$\pm$0.94} & 52.67$\pm$0.01 & 31.00$\pm$0.09 & 55.98$\pm$0.01 & 3.11$\pm$0.09 & 1.12$\pm$0.20 & 4.10$\pm$0.07 \\
\midrule
\multicolumn{16}{l}{\textit{Contrastive}} \\
SoftCLT & 30.28$\pm$0.18 & 13.17$\pm$0.08 & 30.73$\pm$0.09 & 45.52$\pm$3.55 & 37.97$\pm$3.55 & 40.32$\pm$2.43 & 29.70$\pm$0.39 & 7.18$\pm$0.09 & 23.53$\pm$0.33 & 52.21$\pm$0.21 & 30.60$\pm$0.14 & 55.98$\pm$0.16 & 11.30$\pm$0.31 & 1.93$\pm$0.59 & 13.06$\pm$0.42 \\
\bottomrule
\end{tabular}%
}
\end{table}

\begin{table}[ht]
\centering
\caption{Linear-probing classification accuracy (\%, single seed) using a \textbf{24-layer} backbone pre-trained on \textbf{Monash} for \textbf{40 epochs}. Higher is better; best per dataset (among completed objectives) in \textbf{bold}. Datasets: Ethanol (EthanolConcentration), FaceDet (FaceDetection), Handwrit (Handwriting), Heartbt (Heartbeat), JapVowel (JapaneseVowels), SCP1/SCP2 (SelfRegulationSCP1/2), Digits (SpokenArabicDigits), UWave (UWaveGestureLibrary).}
\label{tab:classification_depth24}
\resizebox{\textwidth}{!}{%
\begin{tabular}{l*{9}{c}}
\toprule
\textbf{Objective} & \textbf{Ethanol} & \textbf{FaceDet} & \textbf{Handwrit} & \textbf{Heartbt} & \textbf{JapVowel} & \textbf{SCP1} & \textbf{SCP2} & \textbf{Digits} & \textbf{UWave} \\
\midrule
\textit{Random frozen encoder} & 25.10 & 53.21 & 3.76 & 67.32 & 82.70 & 75.43 & 51.67 & 54.21 & 31.87 \\
\midrule
\multicolumn{10}{l}{\textit{Signal-space predictive and reconstructive}} \\
NTP & 24.71 & 59.48 & 7.88 & 67.80 & 89.73 & 74.74 & 52.22 & 88.40 & 53.12 \\
MAE & 25.10 & 51.50 & 3.76 & 71.22 & 73.24 & 56.31 & 49.44 & 55.66 & 26.25 \\
Diffusion & 25.10 & 56.58 & 10.71 & 65.85 & \textbf{91.62} & 73.04 & 53.89 & 89.72 & 55.63 \\
\midrule
\multicolumn{10}{l}{\textit{Latent-space alignment}} \\
DINO & 27.38 & 51.08 & 11.76 & 70.73 & 72.43 & 73.04 & \textbf{54.44} & 65.58 & 30.00 \\
JEPA & 25.10 & 52.16 & 3.76 & 61.46 & 90.27 & 68.26 & 53.33 & 74.17 & 42.19 \\
LeJEPA & 26.62 & \textbf{61.83} & \textbf{15.53} & 65.37 & 73.78 & \textbf{78.50} & 47.22 & \textbf{90.72} & \textbf{76.25} \\
\midrule
\multicolumn{10}{l}{\textit{Contrastive}} \\
SoftCLT & \textbf{28.52} & 50.26 & 4.47 & \textbf{73.66} & 52.16 & 63.14 & 53.89 & 34.79 & 25.31 \\
\bottomrule
\end{tabular}%
}
\end{table}

\FloatBarrier
\subsection{Depth Scaling}
\label{subsec:appendix_depth_scaling}

\paragraph{Downstream performance across depths}
\begin{figure}[t]
\centering
\begin{tikzpicture}
\begin{groupplot}[
    group style={
        group size=3 by 1,
        horizontal sep=0.7cm
    },
    width=0.3\textwidth,
    height=3.8cm,
    xlabel={Layers},
    xtick={2,4,8,12,24},
    tick label style={font=\scriptsize},
    label style={font=\scriptsize},
    title style={font=\small},
    grid=both,
    xmin=2, xmax=24
]
\nextgroupplot[
    title={ETTh1},
    ylabel={MSE},
    ymin=0.40, ymax=0.75,
    legend style={
        at={(0.5,-0.45)},
        anchor=north,
        legend columns=4,
        font=\scriptsize,
        draw=none
    }
]
\addplot[color=blue, dashed, mark=*] coordinates {(2,0.4108)(4,0.4108)(8,0.4130)(12,0.4128)(24,0.4075)};
\addlegendentry{Random frozen encoder}
\addplot[color=green!60!black, mark=square*] coordinates {(2,0.4355)(4,0.4365)(8,0.4403)(12,0.4335)(24,0.4368)};
\addlegendentry{NTP}
\addplot[color=orange, mark=triangle*] coordinates {(2,0.4155)(4,0.43625)(8,0.43725)(12,0.4410)(24,0.44625)};
\addlegendentry{DINO}
\addplot[color=red, mark=diamond*] coordinates {(2,0.4290)(4,0.4345)(8,0.43775)(12,0.4420)(24,0.45475)};
\addlegendentry{JEPA}
\addplot[color=purple, mark=*] coordinates {(2,0.4460)(4,0.4400)(8,0.4383)(12,0.4515)(24,0.5538)};
\addlegendentry{MAE}
\addplot[color=brown, mark=*] coordinates {(2,0.5600)(4,0.59275)(8,0.55918)(12,0.55825)(24,0.5970)};
\addlegendentry{LeJEPA}
\addplot[color=cyan!55!black, mark=pentagon*] coordinates {(2,0.4522)(4,0.4448)(8,0.4686)(12,0.7082)(24,0.7199)};
\addlegendentry{SoftCLT}
\addplot[color=magenta, mark=star] coordinates {(2,0.4160)(4,0.41425)(8,0.4175)(12,0.4145)(24,0.42275)};
\addlegendentry{Diffusion}
\nextgroupplot[
    title={ETTh2},
    ymin=0.30, ymax=0.46,
    yticklabels={}
]
\addplot[color=blue, dashed, mark=*] coordinates {(2,0.34625)(4,0.34400)(8,0.34425)(12,0.34400)(24,0.34425)};
\addplot[color=green!60!black, mark=square*] coordinates {(2,0.35600)(4,0.35825)(8,0.36325)(12,0.35850)(24,0.36125)};
\addplot[color=orange, mark=triangle*] coordinates {(2,0.34475)(4,0.35223)(8,0.36400)(12,0.36550)(24,0.36775)};
\addplot[color=red, mark=diamond*] coordinates {(2,0.35250)(4,0.37475)(8,0.37850)(12,0.38475)(24,0.39025)};
\addplot[color=purple, mark=*] coordinates {(2,0.35000)(4,0.35425)(8,0.36500)(12,0.36550)(24,0.37075)};
\addplot[color=brown, mark=*] coordinates {(2,0.38300)(4,0.39600)(8,0.39925)(12,0.40425)(24,0.40675)};
\addplot[color=cyan!55!black, mark=pentagon*] coordinates {(2,0.3801)(4,0.3767)(8,0.3853)(12,0.4272)(24,0.4310)};
\addplot[color=magenta, mark=star] coordinates {(2,0.32025)(4,0.31700)(8,0.31750)(12,0.32175)(24,0.32975)};
\nextgroupplot[
    title={ETTm1},
    ymin=0.33, ymax=0.73,
    yticklabels={}
]
\addplot[color=blue, dashed, mark=*] coordinates {(2,0.35775)(4,0.35675)(8,0.35600)(12,0.35625)(24,0.35925)};
\addplot[color=green!60!black, mark=square*] coordinates {(2,0.35075)(4,0.34975)(8,0.35125)(12,0.35150)(24,0.35725)};
\addplot[color=orange, mark=triangle*] coordinates {(2,0.35825)(4,0.35450)(8,0.35650)(12,0.35725)(24,0.35950)};
\addplot[color=red, mark=diamond*] coordinates {(2,0.36525)(4,0.36650)(8,0.35775)(12,0.35575)(24,0.35875)};
\addplot[color=purple, mark=*] coordinates {(2,0.36600)(4,0.35625)(8,0.35850)(12,0.36375)(24,0.38875)};
\addplot[color=brown, mark=*] coordinates {(2,0.38250)(4,0.40700)(8,0.39650)(12,0.39925)(24,0.43275)};
\addplot[color=cyan!55!black, mark=pentagon*] coordinates {(2,0.3728)(4,0.3825)(8,0.3747)(12,0.6923)(24,0.6952)};
\addplot[color=magenta, mark=star] coordinates {(2,0.36550)(4,0.36100)(8,0.36275)(12,0.35825)(24,0.35525)};
\end{groupplot}
\end{tikzpicture}
\caption{Forecasting performance (MSE) as a function of backbone depth.}
\vspace{-1.5em}
\label{fig:depth_scaling_combined}
\end{figure}

Figure~\ref{fig:depth_scaling_combined} shows forecasting performance as a function of
Transformer depth. Across objectives, MSE is approximately flat between 2 and
8 layers and often increases at greater depths; the shallowest configuration
is frequently within approximately 0.01 MSE of the best result. The
rise beyond 8 layers is particularly visible for MAE on \textit{ETTh1}
(0.438 at $N{=}8$ vs.\ 0.554 at $N{=}24$) and SoftCLT on \textit{ETTm1}
(0.375 at $N{=}8$ vs.\ 0.692 at $N{=}12$); these are single-seed sweep values
and differ from the five-seed the main tables.
The broad conclusion that SSL provides limited forecasting benefit remains
stable across depths. Although the deeper models train stably under the fixed
budget (Appendix~\ref{subsec:appendix_depth_scaling}), this experiment is not
compute-normalized and does not isolate whether degradation at greater depth
arises from overfitting, optimization efficiency, or objective--capacity
interactions.

\paragraph{Implications for Depth Scaling.}
These results support our main finding that the absence of forecasting gains beyond eight layers is not caused by insufficient training. Even at 24 layers, models converge within the allocated budget, suggesting that the lack of improvement at greater depth reflects the learned representations rather than optimization constraints. While we do not perform compute-normalized scaling (e.g., matched optimization steps), the consistent convergence behavior across depths indicates that additional training would likely yield only marginal improvements.

Finally, deeper models incur substantially higher computational cost. For example, 24-layer models trained on Synthetic or Hybrid data require up to 28 hours per run (Appendix~\ref{subset:appendix_compute}), reinforcing the practical relevance of the depth result.
\subsection{Convergence Analysis}
\label{subsec:conv_anl}
\paragraph{Convergence Under Fixed Pre-training Budget.}
All models are pre-trained for a fixed budget of 20 epochs across all backbone depths, including the deepest 24-layer configurations. Empirically, we observe stable convergence behavior across all evaluated SSL paradigms. Training loss decreases consistently in early epochs and plateaus thereafter, with no evidence of instability or divergence even for deep models.

Model selection is based on the best validation checkpoint rather than the final epoch. Across methods, peak validation performance typically occurs before the full budget is exhausted, most commonly around epochs 15--16. This indicates that the fixed 20-epoch budget is sufficient for convergence and does not disadvantage deeper models.

\paragraph{Training Dynamics for Deep Models.}
Figure~\ref{fig:pretrain_loss_all_hybrid} presents representative training curves for 8-layer models pre-trained on the Synthetic dataset for all seven paradigms. All methods exhibit rapid initial improvement followed by saturation. DINO converges quickly within the first few epochs, MAE shows a gradual but consistent decrease, and NTP stabilizes early with minor fluctuations. Improvements beyond epoch 15 are marginal across all methods.

\begin{figure}[ht]
\centering
\begin{tikzpicture}
\begin{axis}[
    width=0.8\linewidth,
    height=6.5cm,
    xlabel={Epoch},
    ylabel={Pre-training loss},
    xmin=1, xmax=20,
    ymin=0, ymax=2.5,
    xtick={1,5,10,15,20},
    grid=major,
    grid style={dashed, gray!30},
    legend style={
        at={(0.5,1.02)},
        anchor=south,
        legend columns=4,
        font=\small
    },
    tick label style={font=\small},
    label style={font=\small}
]

\addplot[thick, blue] coordinates {
(1,2.097) (2,1.518) (3,1.516) (4,1.502) (5,1.501) (6,1.573) (7,1.499) (8,1.470) (9,1.463) (10,1.448)
(11,1.447) (12,1.437) (13,1.424) (14,1.412) (15,1.399) (16,1.386) (17,1.368) (18,1.352) (19,1.337) (20,1.334)
};

\addplot[thick, red, dashed] coordinates {
(1,0.850) (2,0.766) (3,0.550) (4,0.462) (5,0.433) (6,0.419) (7,0.411) (8,0.405) (9,0.401) (10,0.398)
(11,0.395) (12,0.393) (13,0.390) (14,0.389) (15,0.388) (16,0.387) (17,0.386) (18,0.385) (19,0.385) (20,0.385)
};

\addplot[thick, orange, dotted] coordinates {
(1,0.558) (2,0.462) (3,0.445) (4,0.436) (5,0.430) (6,0.425) (7,0.422) (8,0.418) (9,0.416) (10,0.414)
(11,0.412) (12,0.410) (13,0.408) (14,0.407) (15,0.406) (16,0.405) (17,0.404) (18,0.403) (19,0.403) (20,0.402)
};

\addplot[thick, green!60!black, dashdotted] coordinates {
(1,0.886) (2,0.733) (3,0.696) (4,0.692) (5,0.685) (6,0.680) (7,0.676) (8,0.679) (9,0.676) (10,0.678)
(11,0.680) (12,0.672) (13,0.662) (14,0.660) (15,0.656) (16,0.653) (17,0.652) (18,0.649) (19,0.648) (20,0.648)
};

\addplot[thick, violet, densely dashed] coordinates {
(1,0.129) (2,0.056) (3,0.050) (4,0.048) (5,0.049) (6,0.048) (7,0.045) (8,0.046) (9,0.043) (10,0.042)
(11,0.040) (12,0.042) (13,0.044) (14,0.043) (15,0.041) (16,0.042) (17,0.042) (18,0.040) (19,0.042) (20,0.042)
};

\addplot[thick, brown, densely dotted] coordinates {
(1,0.097) (2,0.080) (3,0.077) (4,0.077) (5,0.076) (6,0.076) (7,0.076) (8,0.076) (9,0.076) (10,0.076)
(11,0.076) (12,0.076) (13,0.076) (14,0.076) (15,0.076) (16,0.076) (17,0.076) (18,0.076) (19,0.076) (20,0.076)
};

\addplot[thick, cyan!60!black] coordinates {
(1,1.074) (2,0.577) (3,0.570) (4,0.542) (5,0.543) (6,0.530) (7,0.703) (8,0.539) (9,0.524) (10,0.529)
(11,0.547) (12,0.551) (13,0.513) (14,0.510) (15,0.527) (16,0.508) (17,0.529) (18,0.509) (19,0.507) (20,0.507)
};

\legend{DINO, MAE, NTP, JEPA, LeJEPA, Diffusion, SoftCLT}

\end{axis}
\end{tikzpicture}
\caption{Pre-training loss vs.\ epoch for all seven objectives, pre-trained on \textbf{hybrid} data (Monash $+$ synthetic, 2.5M samples, single seed, 20 epochs). Each curve is the model's own raw per-epoch training loss; because the objectives differ (DINO/JEPA/LeJEPA latent-space, MAE/Diffusion reconstruction, NTP prediction, SoftCLT contrastive), absolute values are not directly comparable across methods.}
\label{fig:pretrain_loss_all_hybrid}
\end{figure}
\FloatBarrier
\section{Controlled Ablations}
\label{app:ablations}

\subsection{Augmentations Ablation Study LeJEPA}
\label{subsec:appendix_augmentations}

\paragraph{Setup.}
To isolate the impact of augmentation design on LeJEPA, we evaluate several families of transformations beyond the default DWT-based pipeline. Each augmentation is applied symmetrically to construct dual views and is parameterized by a single magnitude controlling the strength of the transformation. We consider the following families:

\begin{itemize}
    \item \textbf{Galilean \& Special-Relativistic:} global temporal rescaling and local Lorentz-style transformations of the form $\gamma(x - v \cdot t)$.
    \item \textbf{Coordinate Transformations:} global polar warps and local rotations in the value--time plane.
    \item \textbf{Hyperbolic Transformations:} global hyperbolic warps and local Möbius shifts on the Poincaré disk.
    \item \textbf{DWT (baseline):} wavelet-based augmentations with structured perturbations in the frequency domain.
\end{itemize}


\paragraph{Results.}
Across the forecasting benchmarks, DWT-based augmentations consistently provide the most stable and competitive performance.

\paragraph{LeJEPA}
For \textbf{forecasting}, DWT achieves on par with hyperbolic, (lowest MSE on ETTM1, lowest MAE on ETTM2, and ETTH1), on par with hyperbolic, and passing the others.
while on \textbf{classification}, LeJEPA wins 5 out of 9, winning them by a large margin, while on the other 4, DWT is highly competitive.

\paragraph{Discussion.}
These results indicate that the effectiveness of augmentations depends strongly on the alignment between the induced invariances and the underlying signal structure. DWT-based augmentations preserve temporal locality while introducing structured perturbations in the frequency domain, which appears well-suited for forecasting.

In contrast, the alternative transformations introduce more global or nonlinear distortions that are less aligned with the statistical properties of real-world time series. While these transformations are theoretically motivated, they do not translate into effective inductive biases in practice.

\paragraph{Conclusion.}
Overall, the ablation confirms that the choice of augmentation is critical for LeJEPA. DWT-based augmentations provide a robust and well-aligned inductive bias, while alternative transformation families fail to consistently improve performance. This justifies our design choice of using wavelet-based augmentations as the default across all experiments.
\begin{table}[ht]
\centering
\caption{LeJEPA Forecasting performance (MSE and MAE) across augmentation strategies on Monash. Lower is better; best results per dataset and metric are highlighted in \textbf{bold}.}
\label{tab:appendix_aug_forecasting_lejepa}
\small
\setlength{\tabcolsep}{6pt}
\begin{tabular}{lcccccc}
\toprule
\textbf{Augmentation} & \textbf{Loss} & \textbf{ETTH1} & \textbf{ETTH2} & \textbf{ETTM1} & \textbf{ETTM2} & \textbf{Weather} \\
\midrule

Galilean \& Relativistic
& MSE & 0.682 & \textbf{0.391} & 0.465 & 0.295 & 0.262 \\
& MAE & 0.572 & \textbf{0.426} & 0.448 & 0.353 & 0.298 \\

Coordinate / Polar
& MSE & 0.642 & 0.398 & 0.455 & 0.294 & \textbf{0.261} \\
& MAE & 0.552 & 0.430 & 0.446 & 0.351 & \textbf{0.296} \\

DWT
& MSE & 0.611 & 0.399 & \textbf{0.444} & 0.296 & 0.263 \\
& MAE & \textbf{0.537} & 0.427 & 0.439 & \textbf{0.349} & 0.298 \\

Hyperbolic
& MSE & \textbf{0.606} & 0.406 & 0.448 & \textbf{0.293} & 0.263 \\
& MAE & 0.538 & 0.441 & \textbf{0.437} & 0.350 & 0.299 \\

\bottomrule
\end{tabular}
\end{table}
\begin{table}[ht]
\centering
\caption{Linear-probing classification accuracy (\%, single seed, 123) across augmentation strategies for a \textbf{LeJEPA} backbone pre-trained on Monash. Higher is better; best per dataset in \textbf{bold}. Datasets: Ethanol (EthanolConcentration), FaceDet (FaceDetection), Handwrit (Handwriting), Heartbt (Heartbeat), JapVowel (JapaneseVowels), SCP1/SCP2 (SelfRegulationSCP1/2), Digits (SpokenArabicDigits), UWave (UWaveGestureLibrary).}
\label{tab:appendix_aug_classification_lejepa}
\resizebox{\textwidth}{!}{%
\begin{tabular}{lccccccccc}
\toprule
\textbf{Augmentation} & \textbf{Ethanol} & \textbf{FaceDet} & \textbf{Handwrit} & \textbf{Heartbt} & \textbf{JapVowel} & \textbf{SCP1} & \textbf{SCP2} & \textbf{Digits} & \textbf{UWave} \\
\midrule
Galilean \& Relativistic & \textbf{28.14} & 62.54 & 16.12 & 60.98 & 73.78 & 81.57 & 48.89 & 91.86 & 71.56 \\
Coordinate / Polar       & 24.33 & 62.54 & \textbf{16.47} & 60.00 & 84.05 & 80.89 & 45.00 & 92.09 & 74.06 \\
DWT                      & 25.10 & 60.64 & 13.41 & \textbf{64.88} & \textbf{87.57} & \textbf{84.98} & 47.22 & \textbf{92.68} & \textbf{76.56} \\
Hyperbolic               & 22.05 & \textbf{65.07} & 11.76 & 61.46 & 67.03 & 73.38 & \textbf{50.56} & 77.81 & 70.63 \\
\bottomrule
\end{tabular}%
}
\end{table}



\FloatBarrier
\subsection{Results for Same Parameter Evaluation}
\label{subsec:fair_eval}

We isolate the effect of the pre-training \emph{objective} from confounds of capacity and optimisation by re-training every model under an identical budget: an 8-layer encoder with a 4-layer predictor (when needed), embedding dimension 128, patch size 16, learning rate $1{\times}10^{-4}$, batch size 64, and 20 pre-training epochs, with a fixed context window. All models are pre-trained on Monash and evaluated by linear probing with these settings held constant, so any difference in downstream performance reflects the SSL objective alone. NTP and MAE already use this configuration in the main experiments, so their rows are unchanged from the Monash tables. Tables~\ref{tab:fc_monash_equal}, \ref{tab:cls_monash_equal}, and \ref{tab:anom_monash_equal} give the per-dataset results.

\paragraph{Overall Trends.}
Under a matched budget the paradigm ranking is strongly task-dependent. Predictive \textit{Latent-space alignment} (JEPA, LeJEPA) dominates anomaly detection and gesture/telemetry classification, \textit{Signal-space predictive and reconstructive} objectives are strongest on periodic and speech signals, and the \textit{random frozen encoder} baseline is essentially unbeaten on forecasting. Invariance-based DINO is competitive on point-adjusted anomaly metrics but collapses on shift-sensitive datasets, and LeJEPA -- the strongest classifier -- is the weakest forecaster, consistent with the precision--invariance hypothesis of Section~\ref{sec:discussion}.

\paragraph{Forecasting.}
Table~\ref{tab:fc_monash_equal} shows that, under a matched budget, pre-training provides no consistent forecasting benefit. The Random frozen encoder baseline attains the lowest MAE on ETTh1, ETTm1, Electricity, and Traffic and the lowest MSE on ETTh1 and Electricity; the remaining wins are marginal and split between Diffusion (ETTh2, MSE 0.329), DINO (ETTm2, MSE 0.257), and NTP/JEPA (Weather MSE 0.235, Traffic MSE 0.433). Most pre-trained models land within 0.01--0.02 MSE of the baseline. LeJEPA, the strongest classifier, is by far the weakest forecaster
(ETTh1 MSE 0.618, Traffic 0.822), consistent with the near-isotropic
embedding geometry induced by SIGReg (Appendix~\ref{subsec:appendix_lejepa}).
\begin{table}[ht]
\centering
\setlength{\tabcolsep}{3pt}
\caption{Linear-probing forecasting error (MSE and MAE, mean~$\pm$~std over five seeds, averaged over horizons $\{96,192,336,720\}$) using an 8-layer backbone pre-trained on the \textbf{Monash} dataset under an \textbf{equal pre-training budget} (identical LR, batch size, epochs, and context across all objectives). Rows are grouped into signal-space predictive and reconstructive, latent-space alignment, and contrastive objectives; lower is better and the best result per dataset and metric is in \textbf{bold}.}
\label{tab:fc_monash_equal}
\resizebox{\textwidth}{!}{%
\begin{tabular}{l*{14}{c}}
\toprule
\textbf{Objective} & \multicolumn{2}{c}{\textbf{ETTh1}} & \multicolumn{2}{c}{\textbf{ETTh2}} & \multicolumn{2}{c}{\textbf{ETTm1}} & \multicolumn{2}{c}{\textbf{ETTm2}} & \multicolumn{2}{c}{\textbf{Weather}} & \multicolumn{2}{c}{\textbf{Electricity}} & \multicolumn{2}{c}{\textbf{Traffic}} \\
\cmidrule(lr){2-3}\cmidrule(lr){4-5}\cmidrule(lr){6-7}\cmidrule(lr){8-9}\cmidrule(lr){10-11}\cmidrule(lr){12-13}\cmidrule(lr){14-15}
 & MSE & MAE & MSE & MAE & MSE & MAE & MSE & MAE & MSE & MAE & MSE & MAE & MSE & MAE \\
\midrule
\textit{Random frozen encoder} & \textbf{0.414$\pm$0.002} & \textbf{0.426$\pm$0.002} & 0.350$\pm$0.002 & 0.393$\pm$0.001 & 0.356$\pm$0.001 & \textbf{0.379$\pm$0.001} & 0.261$\pm$0.001 & 0.318$\pm$0.000 & 0.248$\pm$0.000 & 0.281$\pm$0.001 & \textbf{0.172$\pm$0.000} & \textbf{0.267$\pm$0.000} & 0.435$\pm$0.000 & \textbf{0.298$\pm$0.000} \\
\midrule
\multicolumn{15}{l}{\textit{Signal-space predictive and reconstructive}} \\
NTP & 0.435$\pm$0.002 & 0.441$\pm$0.001 & 0.362$\pm$0.003 & 0.400$\pm$0.002 & \textbf{0.352$\pm$0.001} & 0.381$\pm$0.000 & 0.262$\pm$0.001 & 0.319$\pm$0.001 & 0.235$\pm$0.000 & 0.270$\pm$0.001 & 0.180$\pm$0.001 & 0.277$\pm$0.001 & 0.445$\pm$0.001 & 0.311$\pm$0.001 \\
MAE & 0.438$\pm$0.002 & 0.438$\pm$0.002 & 0.368$\pm$0.004 & 0.402$\pm$0.003 & 0.364$\pm$0.002 & 0.390$\pm$0.001 & 0.264$\pm$0.004 & 0.324$\pm$0.003 & 0.244$\pm$0.001 & 0.279$\pm$0.001 & 0.194$\pm$0.001 & 0.295$\pm$0.001 & 0.481$\pm$0.003 & 0.345$\pm$0.003 \\
Diffusion & 0.433$\pm$0.002 & 0.446$\pm$0.001 & \textbf{0.333$\pm$0.002} & \textbf{0.384$\pm$0.002} & 0.375$\pm$0.003 & 0.395$\pm$0.002 & 0.263$\pm$0.001 & 0.320$\pm$0.001 & 0.242$\pm$0.001 & 0.280$\pm$0.001 & 0.226$\pm$0.002 & 0.342$\pm$0.003 & 0.523$\pm$0.007 & 0.383$\pm$0.004 \\
\midrule
\multicolumn{15}{l}{\textit{Latent-space alignment}} \\
DINO & 0.437$\pm$0.006 & 0.443$\pm$0.003 & 0.363$\pm$0.003 & 0.396$\pm$0.003 & 0.356$\pm$0.002 & 0.385$\pm$0.001 & \textbf{0.257$\pm$0.005} & \textbf{0.313$\pm$0.003} & 0.236$\pm$0.001 & 0.270$\pm$0.001 & 0.179$\pm$0.003 & 0.278$\pm$0.004 & 0.443$\pm$0.005 & 0.316$\pm$0.008 \\
JEPA & 0.440$\pm$0.001 & 0.441$\pm$0.001 & 0.365$\pm$0.003 & 0.406$\pm$0.002 & 0.365$\pm$0.001 & 0.386$\pm$0.001 & 0.275$\pm$0.002 & 0.331$\pm$0.001 & \textbf{0.235$\pm$0.000} & \textbf{0.269$\pm$0.000} & 0.174$\pm$0.000 & 0.268$\pm$0.000 & \textbf{0.433$\pm$0.001} & 0.305$\pm$0.001 \\
LeJEPA & 0.618$\pm$0.016 & 0.541$\pm$0.008 & 0.388$\pm$0.003 & 0.421$\pm$0.002 & 0.439$\pm$0.010 & 0.436$\pm$0.006 & 0.282$\pm$0.002 & 0.340$\pm$0.001 & 0.258$\pm$0.001 & 0.294$\pm$0.001 & 0.342$\pm$0.008 & 0.430$\pm$0.007 & 0.822$\pm$0.021 & 0.540$\pm$0.010 \\
\midrule
\multicolumn{15}{l}{\textit{Contrastive}} \\
SoftCLT & 0.458$\pm$0.005 & 0.457$\pm$0.002 & 0.359$\pm$0.013 & 0.404$\pm$0.006 & 0.380$\pm$0.001 & 0.404$\pm$0.002 & 0.265$\pm$0.001 & 0.323$\pm$0.002 & 0.244$\pm$0.002 & 0.279$\pm$0.001 & 0.216$\pm$0.003 & 0.318$\pm$0.002 & 0.489$\pm$0.002 & 0.359$\pm$0.002 \\
\bottomrule
\end{tabular}%
}
\end{table}

\paragraph{Classification.}
Table~\ref{tab:cls_monash_equal} shows that with matched capacity LeJEPA is the strongest encoder, taking the top accuracy on five of nine datasets -- FaceDetection (61.8), Handwriting (15.2), SCP1 (80.0), SpokenArabicDigits (93.1), and UWaveGestureLibrary (78.8) -- with the largest margins on the gesture benchmarks (UWave $+42$ over the baseline). The remaining datasets are led by SoftCLT on Heartbeat (73.2) and JEPA on SCP2 (57.1), while the random frozen encoder itself attains the highest accuracy on EthanolConcentration (27.7) and JapaneseVowels (93.0). Under a matched budget, no signal-space objective leads any dataset.

\begin{table}[ht]
\centering
\setlength{\tabcolsep}{3pt}
\caption{Linear-probing classification accuracy (\%, mean~$\pm$~std over five seeds) using an 8-layer backbone pre-trained on the \textbf{Monash} dataset under an \textbf{equal pre-training budget} (identical LR, batch size, epochs, and context across all objectives). Datasets: Ethanol (EthanolConcentration), FaceDet (FaceDetection), Handwrit (Handwriting), Heartbt (Heartbeat), JapVowel (JapaneseVowels), SCP1/SCP2 (SelfRegulationSCP1/2), Digits (SpokenArabicDigits), UWave (UWaveGestureLibrary). Rows are grouped into signal-space predictive and reconstructive, latent-space alignment, and contrastive objectives; the best result per dataset is in \textbf{bold}.}
\label{tab:cls_monash_equal}
\resizebox{\textwidth}{!}{%
\begin{tabular}{l*{9}{c}}
\toprule
\textbf{Objective} & \textbf{Ethanol} & \textbf{FaceDet} & \textbf{Handwrit} & \textbf{Heartbt} & \textbf{JapVowel} & \textbf{SCP1} & \textbf{SCP2} & \textbf{Digits} & \textbf{UWave} \\
\midrule
\textit{Random frozen encoder} & \textbf{27.68$\pm$1.88} & 54.49$\pm$0.54 & 4.31$\pm$0.54 & 66.63$\pm$1.33 & 93.02$\pm$0.65 & 76.72$\pm$2.86 & 50.00$\pm$2.60 & 77.61$\pm$0.96 & 36.56$\pm$2.78 \\
\midrule
\multicolumn{10}{l}{\textit{Signal-space predictive and reconstructive}} \\
NTP & 24.94$\pm$0.19 & 57.76$\pm$0.60 & 6.26$\pm$2.08 & 66.83$\pm$3.16 & 89.62$\pm$0.99 & 74.75$\pm$1.22 & 52.89$\pm$3.57 & 88.29$\pm$0.40 & 55.06$\pm$1.83 \\
MAE & 24.56$\pm$0.46 & 54.20$\pm$0.91 & 5.06$\pm$1.30 & 62.05$\pm$1.39 & 86.27$\pm$2.59 & 75.70$\pm$1.97 & 51.66$\pm$2.65 & 81.51$\pm$2.38 & 34.56$\pm$2.93 \\
Diffusion & 24.94$\pm$0.79 & 56.11$\pm$0.24 & 7.50$\pm$0.45 & 65.17$\pm$2.48 & 91.13$\pm$0.55 & 67.44$\pm$4.07 & 52.55$\pm$1.09 & 85.80$\pm$1.15 & 52.00$\pm$2.79 \\
\midrule
\multicolumn{10}{l}{\textit{Latent-space alignment}} \\
DINO & 25.94$\pm$0.37 & 51.44$\pm$0.18 & 8.80$\pm$0.73 & 72.68$\pm$0.53 & 78.00$\pm$0.78 & 69.42$\pm$2.61 & 54.22$\pm$0.75 & 75.79$\pm$2.63 & 32.44$\pm$3.41 \\
JEPA & 24.87$\pm$0.46 & 52.90$\pm$0.37 & 8.97$\pm$0.36 & 60.00$\pm$2.86 & 91.40$\pm$0.52 & 70.85$\pm$2.28 & \textbf{57.11$\pm$2.39} & 83.45$\pm$0.36 & 46.63$\pm$1.93 \\
LeJEPA & 25.70$\pm$1.89 & \textbf{61.78$\pm$0.30} & \textbf{15.20$\pm$0.55} & 61.85$\pm$1.52 & 88.38$\pm$1.04 & \textbf{80.00$\pm$1.47} & 49.55$\pm$2.29 & \textbf{93.08$\pm$0.29} & \textbf{78.81$\pm$2.07} \\
\midrule
\multicolumn{10}{l}{\textit{Contrastive}} \\
SoftCLT & 26.54$\pm$1.19 & 52.51$\pm$0.68 & 7.18$\pm$0.45 & \textbf{73.17$\pm$0.62} & 70.43$\pm$1.17 & 64.99$\pm$1.75 & 53.11$\pm$1.60 & 87.42$\pm$1.23 & 55.00$\pm$5.81 \\
\bottomrule
\end{tabular}%
}
\end{table}

\paragraph{Anomaly Detection.}
Table~\ref{tab:anom_monash_equal} reports point-adjusted (PA), event-wise (Ev), and affiliation (Af) F1 (Sec~\ref{app:anom_metric}). Predictive Latent-space alignment leads the spacecraft-telemetry benchmarks: LeJEPA tops MSL (PA 57.0, Af 49.3) and JEPA tops SMAP (PA 37.1, Af 29.3), both far above the Random frozen encoder baseline (MSL PA 15.6, SMAP PA 8.0). On PSM and SMD the Signal-space predictive and reconstructive and invariance objectives take over the point metrics -- DINO leads PSM (PA 55.6, Af 49.9) and NTP leads SMD (PA 61.8, Af 64.3) -- while the event-wise metric consistently favours the latent-alignment and contrastive models (LeJEPA PSM-Ev 42.4; JEPA/LeJEPA/SoftCLT SMD-Ev $\approx$31), which the point-adjusted score under-credits. Two consistent failure modes appear: SWaT is degenerate for every method (all PA within a point of the baseline, and JEPA/LeJEPA collapse to $\approx$3), and MAE is unstable on PSM (PA 32.6$\pm$4.8) and SWaT (5.8$\pm$4.8).
\begin{table}[ht]
\centering
\setlength{\tabcolsep}{3pt}
\caption{Linear-probing anomaly-detection F1 (\%, mean~$\pm$~std over five seeds) using an 8-layer backbone pre-trained on the \textbf{Monash} dataset under an \textbf{equal pre-training budget} (identical LR, batch size, epochs, and context across all objectives). For each dataset we report point-adjusted (PA), event-wise (Ev), and affiliation-based (Af) F1. Rows are grouped into signal-space predictive and reconstructive, latent-space alignment, and contrastive objectives; the best result per dataset and metric is in \textbf{bold}.}
\label{tab:anom_monash_equal}
\resizebox{\textwidth}{!}{%
\begin{tabular}{l*{15}{c}}
\toprule
\textbf{Objective}
& \multicolumn{3}{c}{\textbf{MSL}} & \multicolumn{3}{c}{\textbf{PSM}} & \multicolumn{3}{c}{\textbf{SMAP}} & \multicolumn{3}{c}{\textbf{SMD}} & \multicolumn{3}{c}{\textbf{SWaT}} \\
\cmidrule(lr){2-4}\cmidrule(lr){5-7}\cmidrule(lr){8-10}\cmidrule(lr){11-13}\cmidrule(lr){14-16}
& PA & Ev & Af & PA & Ev & Af & PA & Ev & Af & PA & Ev & Af & PA & Ev & Af \\
\midrule
\textit{Random frozen encoder} & 15.57$\pm$0.27 & 7.32$\pm$0.06 & 18.66$\pm$0.14 & 47.55$\pm$2.10 & 16.40$\pm$1.23 & 46.12$\pm$1.74 & 7.98$\pm$0.14 & 2.47$\pm$0.04 & 8.43$\pm$0.09 & 56.08$\pm$1.38 & 17.55$\pm$0.36 & 58.25$\pm$0.79 & \textbf{11.84$\pm$0.01} & 1.09$\pm$0.06 & \textbf{14.23$\pm$0.16} \\
\midrule
\multicolumn{16}{l}{\textit{Signal-space predictive and reconstructive}} \\
NTP & 28.58$\pm$0.34 & 13.11$\pm$0.09 & 29.45$\pm$0.19 & 55.02$\pm$9.27 & 29.64$\pm$4.71 & 49.55$\pm$2.94 & 22.68$\pm$0.61 & 6.19$\pm$0.10 & 18.39$\pm$0.42 & \textbf{61.75$\pm$0.34} & 20.00$\pm$0.43 & \textbf{64.29$\pm$0.37} & 11.50$\pm$0.01 & \textbf{2.86$\pm$0.36} & 9.54$\pm$0.16 \\
MAE & 29.36$\pm$0.16 & \textbf{13.29$\pm$0.07} & 29.96$\pm$0.20 & 36.31$\pm$5.19 & 19.13$\pm$3.30 & 41.34$\pm$2.72 & 27.05$\pm$0.85 & 7.17$\pm$0.11 & 21.45$\pm$0.64 & 61.09$\pm$1.27 & 19.05$\pm$0.76 & 63.34$\pm$1.25 & 7.72$\pm$4.80 & 1.28$\pm$0.53 & 10.87$\pm$2.39 \\
Diffusion & 56.66$\pm$0.01 & 12.26$\pm$0.02 & 48.56$\pm$0.08 & 46.35$\pm$0.67 & 21.57$\pm$1.95 & 45.02$\pm$1.78 & 24.38$\pm$0.59 & 5.94$\pm$0.10 & 20.06$\pm$0.48 & 52.35$\pm$0.21 & 29.81$\pm$0.27 & 56.08$\pm$0.37 & 9.25$\pm$1.95 & 1.52$\pm$0.61 & 10.59$\pm$1.44 \\
\midrule
\multicolumn{16}{l}{\textit{Latent-space alignment}} \\
DINO & 18.75$\pm$0.64 & 9.29$\pm$0.51 & 21.26$\pm$0.60 & \textbf{55.56$\pm$7.97} & 31.26$\pm$6.64 & \textbf{49.88$\pm$5.75} & 13.38$\pm$0.66 & 3.75$\pm$0.17 & 11.89$\pm$0.50 & 59.58$\pm$0.66 & 16.51$\pm$0.72 & 61.49$\pm$0.86 & 11.83$\pm$0.01 & 1.22$\pm$0.13 & 14.16$\pm$0.21 \\
JEPA & 56.84$\pm$0.04 & 12.10$\pm$0.03 & 49.10$\pm$0.09 & 43.31$\pm$0.38 & 37.48$\pm$0.27 & 39.23$\pm$0.32 & \textbf{37.13$\pm$0.13} & \textbf{7.09$\pm$0.03} & \textbf{29.28$\pm$0.08} & 52.64$\pm$0.06 & 30.95$\pm$0.12 & 55.99$\pm$0.12 & 3.02$\pm$0.03 & 1.10$\pm$0.06 & 4.04$\pm$0.02 \\
LeJEPA & \textbf{57.03$\pm$0.05} & 12.22$\pm$0.04 & \textbf{49.26$\pm$0.06} & 47.38$\pm$0.58 & \textbf{42.41$\pm$0.81} & 42.81$\pm$0.32 & 34.24$\pm$0.29 & 6.38$\pm$0.04 & 26.99$\pm$0.21 & 52.74$\pm$0.14 & \textbf{30.99$\pm$0.06} & 56.06$\pm$0.14 & 3.18$\pm$0.07 & 1.01$\pm$0.06 & 4.13$\pm$0.05 \\
\midrule
\multicolumn{16}{l}{\textit{Contrastive}} \\
SoftCLT & 30.32$\pm$0.09 & 13.18$\pm$0.03 & 30.73$\pm$0.07 & 48.17$\pm$0.85 & 34.50$\pm$2.19 & 42.01$\pm$1.73 & 28.98$\pm$0.28 & 7.06$\pm$0.05 & 23.00$\pm$0.21 & 52.58$\pm$0.35 & 30.70$\pm$0.19 & 56.25$\pm$0.16 & 11.37$\pm$0.08 & 2.33$\pm$1.70 & 12.19$\pm$2.20 \\
\bottomrule
\end{tabular}%
}
\end{table}

\paragraph{Summary.}
With capacity, optimisation, and context held identical across all models, the objective alone produces a clean task-dependent split: predictive \textit{Latent-space alignment} gives the best initialisation for anomaly detection and classification (LeJEPA in particular), \textit{Signal-space predictive and reconstructive} objectives excel on periodic/speech signals, while random init wins on forecasting.

\subsection{LeJEPA ablations}
\label{subsubsec:LeJEPA_ABB}
\paragraph{Sensitivity to SIGReg Strength.}
Removing SIGReg produces representation collapse in the evaluated
setting: at $\lambda=0$, the reported per-dimension embedding
standard deviation is $0.000$, and mean classification accuracy
falls to $31.26\%$, compared with $54.11\%$ for the random frozen
encoder.

All evaluated positive weights avoid this collapse.
Across $\lambda\in\{0.002,0.008,0.02,0.05\}$, effective rank ranges
from $107.7$ to $114.5$, and per-dimension embedding standard
deviation ranges from $0.964$ to $0.983$.
Mean classification accuracy is $60.43\%$, $61.00\%$, $60.24\%$,
and $59.84\%$, respectively. Each configuration exceeds the random
encoder's mean accuracy, although individual datasets remain
sensitive to the regularization weight.
Increasing $\lambda$ does not produce a monotonic improvement in
classification performance.

The forecasting outcome differs: every evaluated positive weight
yields higher MSE than the random encoder on all five forecasting
datasets in the table. Thus, preventing collapse and obtaining
classification gains do not by themselves establish a positive
forecasting transfer gain.

\subsection{Representation Analyses}
\label{app:representations}

\subsubsection{Embedding Analysis}\label{subsec:appendix_tsne}

\begin{figure}[ht] \centering \includegraphics[scale=0.1]{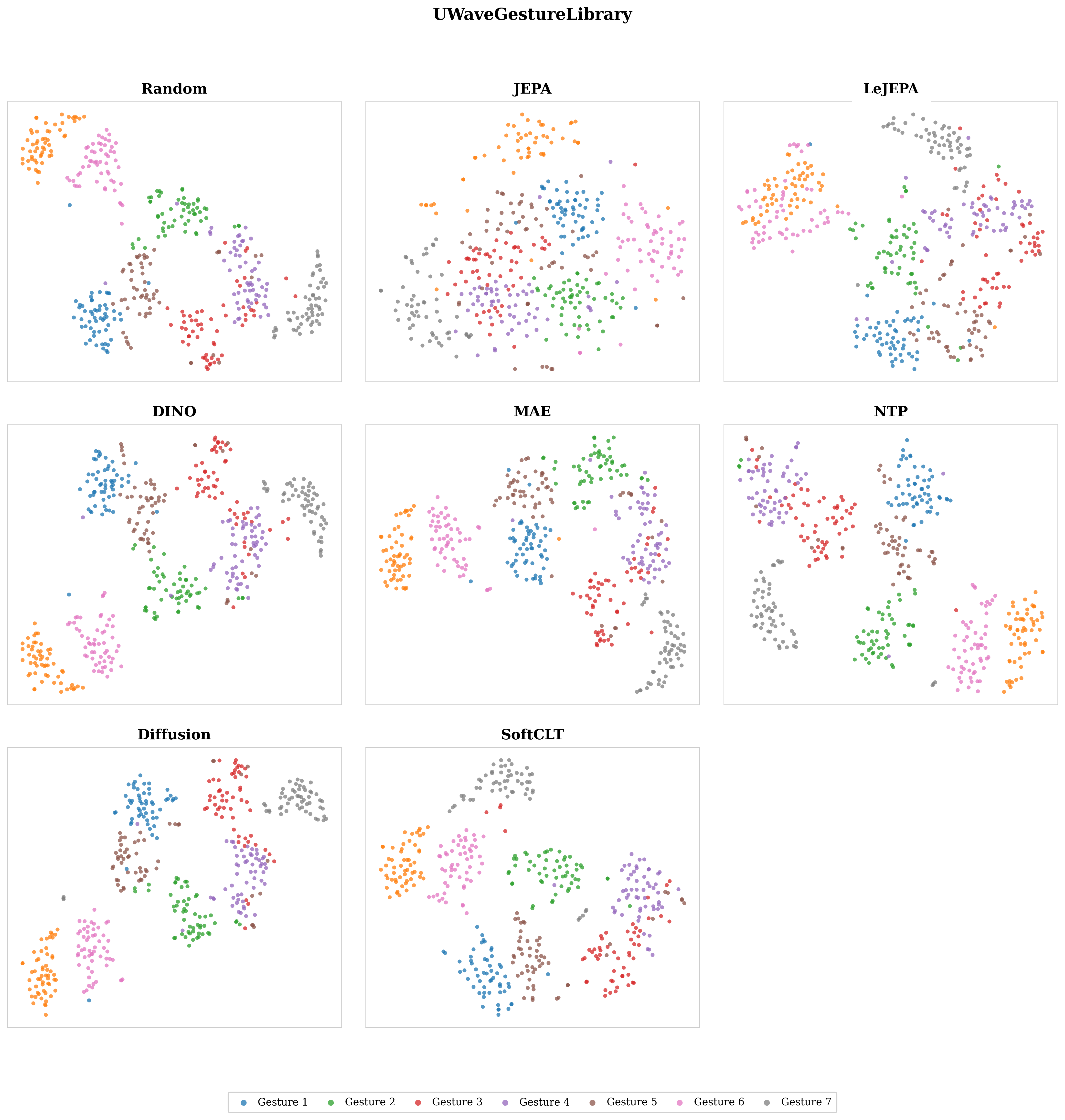} \caption{Embedding Analysis} \label{fig:tsne} \end{figure}

We provide a qualitative analysis of the learned representations using a t-SNE visualization on the \textit{UWaveGestureLibrary} dataset (Figure~\ref{fig:tsne}). Each point corresponds to an embedded time series instance, colored by its class label.

Since this dataset is used exclusively for classification, the visualization reflects the degree of \emph{class separability} induced by each SSL paradigm. Clear differences emerge between methods. {DINO} produces well-separated and compact clusters, indicating strong discriminative structure in the embedding space. {LeJEPA} also yields highly separable clusters, though with slightly more dispersion, suggesting that its invariance constraints preserve class-level structure while allowing some intra-class variability. {JEPA} forms coherent but less sharply separated clusters, reflecting a softer organization of the latent space.

In contrast, \textit{Signal-space predictive and reconstructive} methods exhibit weaker class separation. {MAE} produces partially clustered representations with noticeable overlap between classes, while {NTP} shows more entangled embeddings with limited global organization. {Diffusion} lies between these extremes, forming moderately structured but less compact clusters compared to alignment-based approaches.

Overall, the visualization is consistent with the classification results. LeJEPA's separable clusters translate directly into top linear-probing accuracy. DINO's compact clusters indicate that class structure is present in the embedding, yet its linear-probing accuracy on this dataset is low; since t-SNE is non-linear, this structure is not linearly accessible from the frozen representation, which is consistent with DINO's recovery under full fine-tuning (Section~\ref{sec:results}).

\FloatBarrier
\subsubsection{Temporal Latent Drift}\label{subsec:appendix_drift}
To probe how the learned representations respond to temporal perturbations, we measure the displacement of frozen encoder outputs under two families of input perturbation. For each objective we load the Monash-pre-trained backbone (seed 1337, context 336) and embed 300 non-overlapping windows from each of Weather and ETTh1 (z-scored per channel, stride 48). Each window is perturbed either by a circular time-shift of $k \in \{1,2,4,8,16,32\}$ steps or by zeroing a random $p \in \{10\%, 20\%\}$ of its timesteps, and re-embedded. Drift is the cosine distance $D = 1 - \cos(\mathbf{z}, \mathbf{z}_\varepsilon)$ between the clean and perturbed embeddings, averaged over windows and datasets. We mean-pool the patch tokens before computing $D$; concatenating tokens instead would register a whole-patch shift as a large change purely because tokens move to different slots, which conflates re-ordering with a change in content. Results are from a single pre-training seed and carry no downstream label; they describe representation geometry only.
\begin{table}[h]
\centering
\small
\caption{Temporal latent drift (cosine distance, lower means the representation moved less) of frozen Monash-pre-trained encoders under time-shift and time-masking, averaged over Weather and ETTh1. Shifts of one full patch (16 steps) map every patch onto a neighbouring slot; sub-patch shifts change patch boundaries.}
\label{tab:appendix_drift}
\begin{tabular}{llccc}
\toprule
Family & Model & Shift 4 & Shift 16 & Mask 20\% \\
\midrule
\multirow{3}{*}{Signal-space predictive and reconstructive} & NTP & 0.023 & 0.030 & 0.018 \\
 & MAE & 0.030 & 0.013 & 0.017 \\
 & Diffusion & 0.008 & 0.001 & 0.053 \\
\midrule
\multirow{3}{*}{Latent-space alignment} & DINO & 0.188 & 0.383 & 0.078 \\
 & JEPA & 0.673 & 0.005 & 0.479 \\
 & LeJEPA & 0.062 & 0.237 & 0.157 \\
\midrule
Contrastive & SoftCLT & 0.048 & 0.005 & 0.064 \\
\bottomrule
\end{tabular}
\end{table}

\begin{figure}[h]
\centering
\includegraphics[width=0.6\linewidth]{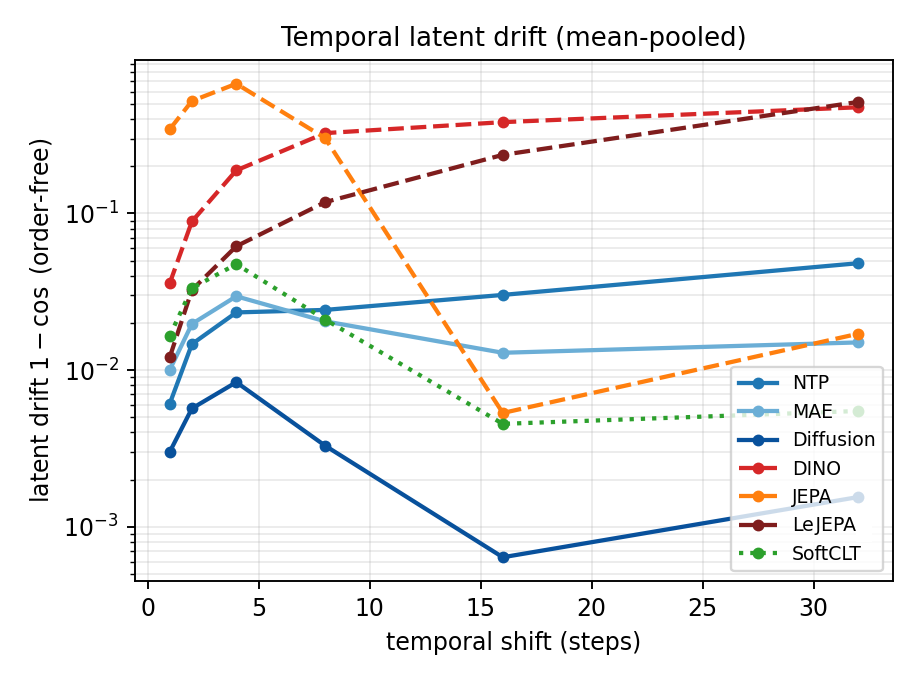}
\caption{Temporal latent drift as a function of shift (log scale, mean-pooled representation). Curves that dip at multiples of the patch length (16, 32) indicate representations determined by patch content irrespective of position.}
\label{fig:appendix_drift}
\end{figure}

\paragraph{Findings.} Two patterns emerge (Table~\ref{tab:appendix_drift}, Figure~\ref{fig:appendix_drift}). First, whole-patch shifts separate the objectives into two groups. MAE, Diffusion, JEPA, and SoftCLT return to near-zero drift at $k{=}16$: after pooling, their representation is largely a function of which patch contents are present, not where they occur. NTP does not ($0.023 \to 0.030$), consistent with its causal mask making token representations position-dependent by construction. DINO and LeJEPA show no such dip and drift monotonically with shift, so their tokens depend on absolute position and the pooled embedding describes the window as a whole. This is a direct geometric counterpart of the distinction drawn in Section~\ref{sec:discussion}: the reconstructive and denoising objectives yield local, content-based representations, while the two view-based objectives yield holistic ones. Second, latent-space alignment encoders are \emph{more} displaced by local masking than signal-space predictive and reconstructive ones (LeJEPA 0.16, JEPA 0.48 vs.\ NTP and MAE $\approx$0.02). We caution against reading this as a lack of robustness: cosine drift in a near-isotropic embedding (Appendix~\ref{subsec:appendix_lejepa}) registers any change, whereas in an embedding dominated by a few high-variance directions most input changes are invisible to cosine distance even when a linear head can exploit them. The probe therefore characterises what the representations encode, not how much fine-grained information they retain; it does not by itself establish the precision side of the precision--invariance hypothesis.

\FloatBarrier
\subsubsection{LeJEPA Analysis}
\label{app:lejepa_analysis}
\begin{figure}[ht]
\centering
\includegraphics[width=0.95\textwidth]{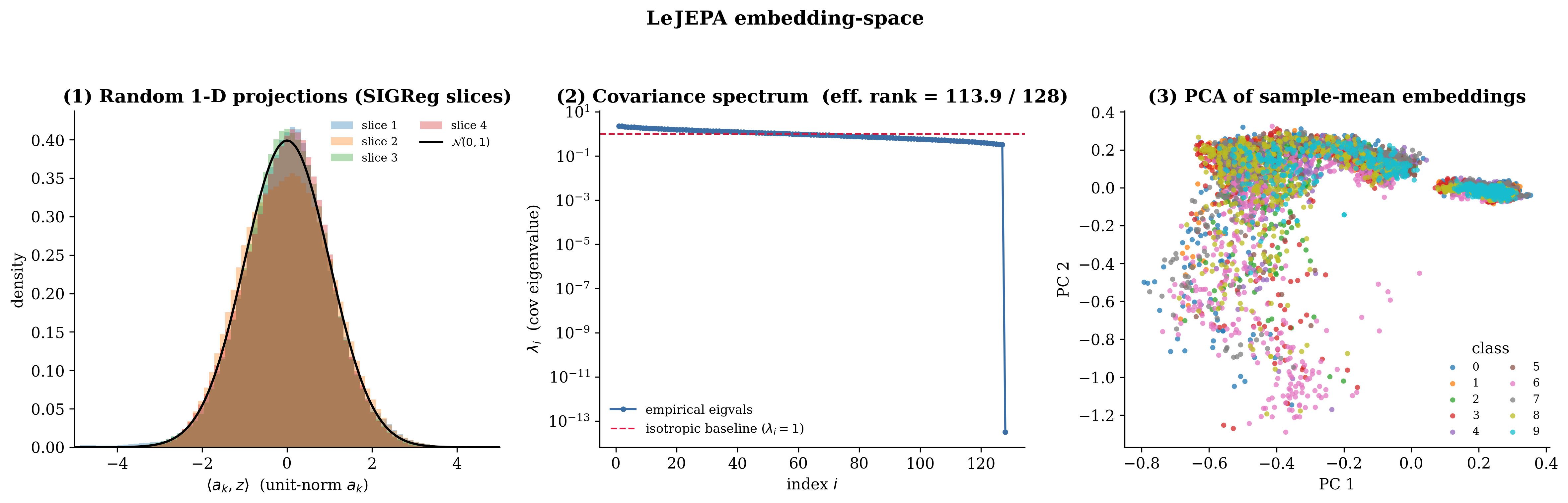}

\caption{
Geometry of LeJEPA's mean-pooled embeddings on
SpokenArabicDigits.
Left: distributions along sampled random directions compared
with a standard normal distribution.
Middle: covariance spectrum, with effective rank approximately
114 out of 128 dimensions.
Right: PCA visualization colored by class, showing class-related
structure within the approximately isotropic representation.
}

\label{fig:sigreg_diagnostics}
\end{figure}

\paragraph{Embedding Geometry.}
Figure~\ref{fig:sigreg_diagnostics} examines mean-pooled embeddings
in $\mathbb{R}^{D}$, whereas SIGReg operates on flattened patch
embeddings in $\mathbb{R}^{LD}$.
The diagnostics therefore characterize geometry after pooling.
The displayed random projections approximately follow a standard
normal distribution, and the covariance spectrum indicates high
effective rank and approximately isotropic structure.
The PCA visualization also suggests that class-related structure
remains present. Downstream classification results provide the
quantitative evidence that these representations retain useful
discriminative information.

\begin{figure}[ht]
\centering
\includegraphics[width=0.7\textwidth]{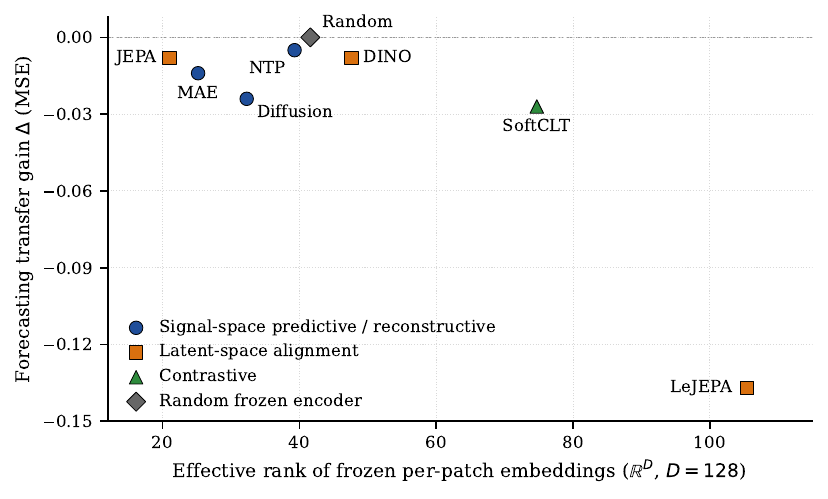}
\caption{
Forecasting transfer gain $\Delta$ (Eq.~\ref{eq:transfer_gain}; MSE,
linear probing, Monash pre-training), averaged over the seven
datasets in Table~\ref{tab:forecasting_l8_monash_clean}, against
the effective rank of frozen per-patch embeddings in
$\mathbb{R}^{D}$.
Negative transfers indicate degradation.
These ranks are computed from per-patch embeddings, whereas
Figure~\ref{fig:sigreg_diagnostics} examines mean-pooled embeddings.
LeJEPA exhibits high effective rank and the largest negative
forecasting transfer among the evaluated objectives.
}
\label{fig:sigreg_MSE}
\end{figure}

\begin{table}[ht]
\centering
\setlength{\tabcolsep}{3pt}
\caption{
Sensitivity of LeJEPA to the SIGReg weight $\lambda$.
Left: effective rank of the frozen per-patch embedding covariance
and per-dimension embedding standard deviation ($D=128$).
Middle: linear-probe forecasting MSE averaged over horizons
$\{96,192,336,720\}$.
Right: linear-probe classification accuracy (\%), with the
unweighted mean across nine datasets.
The $\lambda=0$ configuration removes SIGReg and exhibits
representation collapse.
A dash denotes an unreported result.
Datasets: Ethanol (EthanolConcentration), FaceDet (FaceDetection),
Handwrit (Handwriting), Heartbt (Heartbeat),
JapVowel (JapaneseVowels), SCP1/SCP2 (SelfRegulationSCP1/2),
Digits (SpokenArabicDigits), and UWave (UWaveGestureLibrary).
}
\label{tab:lejepa_lambda_sweep}
\resizebox{\textwidth}{!}{%
\begin{tabular}{c cc ccccc ccccccccc c}
\toprule
& \multicolumn{2}{c}{\textbf{Isotropy}} & \multicolumn{5}{c}{\textbf{Forecasting MSE}} & \multicolumn{10}{c}{\textbf{Classification accuracy (\%)}} \\
\cmidrule(lr){2-3} \cmidrule(lr){4-8} \cmidrule(lr){9-18}
$\lambda$ & Rank & Std & ETTh1 & ETTh2 & ETTm1 & ETTm2 & Weather & Ethanol & FaceDet & Handwrit & Heartbt & JapVowel & SCP1 & SCP2 & Digits & UWave & Mean \\
\midrule
0      & 20.5  & 0.000 & 0.718 & 0.410 & 0.699 & 0.341 & --    & 24.71 & 50.00 & 3.76  & 72.20 & 8.38  & 49.83 & 50.00 & 10.00 & 12.50 & 31.26 \\
\midrule
0.002  & 107.7 & 0.972 & 0.665 & 0.390 & 0.486 & 0.306 & 0.266 & 24.33 & 59.96 & 13.88 & 61.95 & 88.11 & 83.28 & 46.11 & 91.91 & 74.38 & 60.43 \\
0.008 & 109.2 & 0.964 & 0.616 & 0.394 & 0.444 & 0.288 & 0.262 & 26.09
& 61.65 & 13.86 & 62.54 & 86.98 & 83.07 & 48.89 & 93.27 & 71.69 & 60.89 \\
0.02   & 114.2 & 0.981 & 0.619 & 0.390 & 0.451 & 0.291 & 0.264 & 25.86 & 62.68 & 16.24 & 61.95 & 85.68 & 80.55 & 45.56 & 92.68 & 70.94 & 60.24 \\
0.05   & 114.5 & 0.983 & 0.636 & 0.395 & 0.456 & 0.294 & 0.261 & 23.95 & 61.27 & 17.06 & 62.44 & 80.81 & 80.89 & 43.89 & 91.04 & 77.19 & 59.84 \\
\midrule
Random frozen encoder & -- & -- & 0.414 & 0.350 & 0.356 & 0.261 & 0.248 & 27.68 & 54.49 & 4.31 & 66.63 & 93.02 & 76.72 & 50.00 & 77.61 & 36.56 & 54.11 \\
\bottomrule
\end{tabular}%
}
\end{table}

\paragraph{Association with Forecasting Performance.}
\label{app:impact_for}

Figure~\ref{fig:sigreg_MSE} associates LeJEPA's high effective rank
with a negative forecasting transfer. The regularization sweep
reinforces the coexistence of useful classification representations
and poor forecasting performance: all evaluated positive weights
avoid collapse and improve mean classification accuracy over random
features, while their forecasting MSE remains above the control.

These observations motivate the hypothesis that the geometry
encouraged by SIGReg interacts differently with classification and
forecasting. They do not establish isotropy as the cause of the
forecasting degradation. Removing SIGReg produces a collapsed
representation, so that ablation does not provide a usable
non-isotropic control. Comparisons with DINO also change the
alignment loss, teacher dynamics, and other objective components.
Isolating the mechanism requires interventions that vary
representation geometry while preserving non-collapsed solutions
and otherwise comparable training conditions.

\FloatBarrier


\end{document}